\pdfoutput=1
\documentclass[11pt]{article}

\usepackage{acl}

\usepackage{times}
\usepackage{latexsym}

\usepackage[T1]{fontenc}
\usepackage[utf8]{inputenc}

\usepackage{microtype}

\usepackage{inconsolata}

\usepackage{graphicx}

\usepackage{subcaption}

\usepackage{float}

\usepackage{amsmath}

\usepackage{footnote}
\usepackage{hyperref}
\let\oldfootnote\footnote
\renewcommand{\footnote}[1]{%
  \oldfootnote{\begin{minipage}{\linewidth}
    #1
  \end{minipage}}%
}

\usepackage{rotating}
\usepackage{multirow}
\usepackage{booktabs}
\usepackage{array}
\usepackage{graphicx}
\usepackage{xcolor}
\usepackage{colortbl}

\title{Less Is More? Examining Fairness in Pruned Large Language Models for Summarising Opinions}
\author{Nannan Huang \\
  RMIT University, Australia \\
  \texttt{amber.huang@student.rmit.edu.au} 
  \And
  Haytham M. Fayek \\
  RMIT University, Australia \\
  \texttt{haytham.fayek@ieee.org} 
  \AND
  Xiuzhen Zhang \\
  RMIT University, Australia \\
  \texttt{xiuzhen.zhang@rmit.edu.au}}

\begin{document}
\maketitle
\begin{abstract}
% Large language models (LLMs) have shown impressive capabilities in generating abstractive summaries. However, there are two key challenges: their substantial computational requirements and ensuring fair output generation. 
Model compression through post-training pruning offers a way to reduce model size and computational requirements without significantly impacting model performance. 
However, the effect of pruning on the fairness of LLM-generated summaries remains unexplored, particularly for opinion summarisation where biased outputs could influence public views.
% However, the effect that pruning procedures have on the fairness of LLM generated summaries remains an unexplored area of research. Fairness in summarising text with opinions is crucial because generating biased summaries has the risk of potentially influencing public opinions.  
In this paper, we present a comprehensive empirical analysis of opinion summarisation, examining three state-of-the-art pruning methods and various calibration sets across three open-source LLMs using four fairness metrics.
Our systematic analysis reveals that pruning methods have a greater impact on fairness than calibration sets. Building on these insights, we propose High Gradient Low Activation (HGLA) pruning, which identifies and removes parameters that are redundant for input processing but influential in output generation. 
Our experiments demonstrate that HGLA can better maintain or even improve fairness compared to existing methods, showing promise across models and tasks where traditional methods have limitations. 
Our human evaluation shows HGLA-generated outputs are fairer than existing state-of-the-art pruning methods.
Code is available at:~\url{https://github.com/amberhuang01/HGLA}.
% We propose an enhanced pruning method to improve fairness---HGLA. This procedure identifies parameters that are redundant for processing inputs (indicated by low activations) but removing could change the final output (indicated by high gradients). By selectively pruning these parameters, we aim to potentially enhance the model's fairness. Our systematic examination reveals that pruning methods impact model fairness more than calibration sets. Activation-based methods are generally preferred, but HGLA pruning improves fairness where these methods struggle.
\end{abstract}

\section{Introduction}
% Scaling language models has led to emergent capabilities \cite{brown2020language, radford2019language, chowdhery2023palm, touvron2023llama, le2023bloom}, however, it has also led to increased computational resources \cite{wu2022sustainable, zhu2023survey, menghani2023efficient}. Therefore, reducing the size of LLMs while preserving model performance through post-training pruning has become popular and attracted substantial research interest \cite{frantar2023sparsegpt, sun2023simple, das2023beyond}. 
% While state-of-the-art post-training pruning methods are effective at maintaining model performance, their impact beyond performance metrics remains largely understudied. To the best of our knowledge, the work by ~\cite{chrysostomou2024investigating}, which examines model hallucination after pruning, is one of the only studies that investigates pruning from a non-performance angle.
Scaling language models has led to emergent capabilities \cite{brown2020language, radford2019language, chowdhery2023palm, touvron2023llama, le2023bloom}, however, it has also led to increased computational resources \cite{wu2022sustainable, zhu2023survey, menghani2023efficient}. While reducing the size of LLMs through post-training pruning has become popular and attracted substantial research interest \cite{frantar2023sparsegpt, sun2023simple, das2023beyond}, their impact beyond performance metrics remains largely understudied, with the work by~\citet{chrysostomou2024investigating} on model hallucination being one of the only studies that investigates pruning from a non-performance angle.

Another critical but unexplored aspect of pruning is its impact on model fairness.
% The effect of pruning on model fairness remains largely unexplored. 
The use of AI language systems with inherent biases could shape the way audiences interpret and process information \cite{jakesch2023co, durmus2023towards, epstein2023art}, especially in tasks such as opinion summarisation where biased outputs could significantly influence public opinion. It is well-acknowledged that LLMs were exposed to uncurated data that may contain societal bias, which inevitably perpetuates social stereotypes in models \cite{vig2020causal, sheng2019woman, liang2021towards, gallegos2024bias, li2023survey} and propagates to downstream tasks \cite{feng-etal-2023-pretraining}.

% Pruning procedures are comprised of two key elements: the pruning method and the calibration set. 
Post-training pruning typically uses a small calibration set of samples to identify parameters for removal.
Given the fairness concerns, it is crucial to examine all aspects of post-training pruning procedures that could impact model fairness, including both pruning methods and the calibration set.
While recent work has shown that the choice of calibration set significantly impacts model performance \cite{williams2024impact}, its effect on model fairness remains unexplored. This gap is particularly concerning given the established relationship between training data and model bias.

To the best of our knowledge, our work presents the first systematic investigation of how pruning affects fairness in LLM-generated summaries. 
Through a comprehensive analysis using three state-of-the-art pruning methods, various calibration sets across three large language models, and four fairness metrics, our results demonstrate that pruning methods have a greater influence on model fairness than calibration sets.
We find that pruning more model parameters can lead to decreased fairness while maintaining performance, suggesting that performance-focused pruning methods may inadvertently amplify biases. This complex relationship varies across tasks, pruning methods, and calibration sets, highlighting the critical importance of carefully selecting pruning methods and continuously monitoring fairness metrics during the pruning process.

In response to these limitations, we introduce High Gradient Low Activation (HGLA) pruning using calibration sets---a novel procedure that identifies and removes parameters that are redundant in input processing (i.e., low activation) but influential in output generation (i.e., high gradient), offering a promising method to maintain or even improve model fairness during the post-training pruning process.
Our human evaluation study demonstrates that this pruning procedure produces fairer summaries compared to other state-of-the-art post-pruning methods.

\section{Related Work}
\subsection{Post-training Pruning for LLMs}
% Model pruning aims to reduce model size while maintaining performance. 
% Magnitude pruning examines the absolute value of model weights, providing a simple baseline method. The activation magnitudes provide information on how specific parameters contribute to processing input information. We can determine which parameters or connections are less crucial by analysing activation strengths using a calibration dataset. For example, parameters that consistently show weak activations may be unnecessary and redundant in processing input. 
Model pruning aims to reduce model size while maintaining performance. Magnitude pruning examines the absolute value of model weights as a simple baseline. The activation magnitudes indicate how parameters contribute to processing inputs. By analysing activation strengths using calibration data, we can identify less crucial parameters, such as those showing consistently weak activations that may be unnecessary and redundant in processing input.
Recent post-training pruning methods such as SparseGPT~\cite{frantar2023sparsegpt} and Wanda~\cite{sun2023simple} are mainly guided using activation information to prune such redundant parameters. 
A model's gradients provide information on how sensitive its output is with respect to its parameters. Parameters with larger gradients are more sensitive on the model's output, as small changes to these parameters would significantly change the output. GBLM-Pruner \cite{das2023beyond} combines the information in both activation and gradient with a scaling factor to decide how much contribution is made by the gradient. Parameters that produce the lowest gradient and activation should be removed since removing these parameters has the minimum impact on both the output of the model and also how the model processes input information.
These methods require calibration data, typically sampled from C4~\cite{raffel2020exploring}. While research shows calibration set selection significantly impacts model performance~\cite{williams2024impact}, its impact on model fairness has not been studied.

While LLMs can maintain performance with up to 50\% pruning \cite{jaiswal2023compressing}, pruning's effect on fairness remains unexplored. Prior work has shown pruning can reduce hallucination by increasing source document reliance \cite{chrysostomou2024investigating} but the impact on model fairness has not been studied. The key contributions of this work are: (1) evaluating fairness across multiple pruning methods, (2) examining effects of different calibration sets, and (3) based on the limitations identified in existing methods, we propose a novel pruning procedure that addresses these challenges.

\subsection{Fairness in Summarising Opinions}
Prior work has explored social bias in language models across gender, race, and other attributes \cite{vig2020causal, sheng2019woman, liang2021towards, ladhak2023pre, santurkar2023whose}, showing how these biases propagate from training data to downstream tasks \cite{feng-etal-2023-pretraining}. In opinion summarisation, fair models should represent diverse opinions proportionally to their source documents~\cite{shandilya2018fairness}, and biased outputs risk misrepresenting input distributions. Prior research has examined fairness across demographic attributes such as gender, race, political orientation \cite{dash2019summarizing}, dialect \cite{blodgett-etal-2016-demographic}, and opinion diversity \cite{huang-etal-2023-examining}. 

Recent studies have investigated fairness in fine-tuning \cite{huang-etal-2024-bias} and prompt-based models \cite{zhang2023fair}. 
% While fair models should represent diverse opinions proportionally to their population distribution, and biased outputs risk misrepresenting true opinion distributions, the impact of post-training pruning on fairness in abstractive summarisation remains unexplored.
However, the impact of post-training pruning on fairness in abstractive summarisation remains unexplored.

\section{High Gradient Low Activation Pruning (HGLA)}
Formally, pruning methods assign a significance score $\mathbf{S}_{i,j}$ to each element of a layer's weight matrix $\mathbf{W}_{i,j}$. State-of-the-art post-training pruning methods can also factor in additional information, such as the layer's input activations $\mathbf{X}$ or gradients $\mathbf{G}$, derived from a calibration dataset.

% Unlike pruning that solely focuses on model performance by considering only activation or gradient information, from a fairness perspective in LLM generation, we should aim to identify and remove parameters that are redundant in processing input (i.e., showing low activation) but would significantly affect the generated output (i.e., showing high gradient) if removed.
% The goal is to change the output to produce summaries that are less biased and dissimilar to those generated by the vanilla model.

While traditional pruning methods remove parameters based on either low activation or low gradient values to maintain model performance, our fairness-aware approach uses a calibration set to identify a specific type of parameters: those showing low activation but high gradients. These parameters, while not actively processing input features (i.e., low activation), significantly influence the model's output (i.e., high gradients). By targeting such parameters using examples from the calibration set, we aim to modify the model's behaviour to generate less biased outputs that diverge from the vanilla model's outputs.

% Unlike the generation performance perspective considering only the activation or gradient information, when considering fairness in LLM generation, we should aim to identify and remove parameters that are redundant in processing input (i.e., low activation) but would change the generated output (i.e., high gradient) if removed.
% This approach aims to differentiate the pruned model's output from the biased original model. 
% This procedure uses only biased input, which represents opinions from one side, since we aim to remove parameters that are less active and redundant in processing single-sided information. For example, we target parameters that are biased toward or against the intrinsically biased side and sensitive in generation.

The procedure focuses on using a calibration set to isolate and identify parameters that are less activated when processing information from a specific perspective (i.e., redundant in processing information representing a specific side), while removing these weights would produce different outputs (i.e., generate less biased outputs that diverge from the vanilla model's outputs).
Our hypothesis is that these parameters may encode patterns that affect output generation while being non-essential for input processing. By identifying and removing such parameters, we aim to maintain or potentially enhance model fairness while preserving the model's ability to effectively process input information.

Given a weight matrix, $\mathbf{W}$, a gradient matrix, $\mathbf{G}$, and an input feature activation, $\mathbf{X}$, the goal is to generate the masking matrix $\mathbf{W}_m$. We followed the work by \citet{sun2023simple} in using the $\ell_2$ norm in measuring activation magnitudes and \citet{das2023beyond} in using the $\ell_2$ normalisation across samples’ gradients as follows:
\begin{equation}
\mathbf{W}_{\mathrm{m}} = \left|\frac{|\mathbf{W}[i, j]| \cdot \|\mathbf{X}[:, j]\|_2}{|\mathbf{G}[:, i, j]|_p}\right|
\label{eqn:low_act_high_gradient_original}
\end{equation}

In our preliminary analysis, we found that Equation \ref{eqn:low_act_high_gradient_original} is dominated by the inverse of gradients with high magnitude, which resulted in removing parameters with these gradients, that led in some instances to model collapse. We therefore normalised the inverse of the gradients so that the magnitude of the activations and gradients will be on the same scale as follows:
\begin{align}
\label{eqn:scaled_gradient}
|\mathbf{G'}[:, i, j]|_p &= |\mathbf{G}[:, i, j]|_p \cdot \nonumber \\
&\quad\frac{\frac{1}{mn}\sum_{i, j}(|\mathbf{W}[i, j]| \cdot \|\mathbf{X}[:, j]\|_2)} 
{\frac{1}{mn}\sum_{i,j}|\mathbf{G}[:, i, j]|_p}
\end{align}

\section{Experiments}
\subsection{Datasets}
\label{sec:datasets}
We include the following datasets in our study: (1) MOS \cite{bilal2022template} (2) FewSum \cite{bravzinskas2022efficient} (3) AmaSum \cite{bravzinskas2021learning} (4) FairSumm \cite{dash2019summarizing} (5) Amazon reviews 2023 \cite{hou2024bridging}\footnote{\url{https://huggingface.co/datasets/McAuley-Lab/Amazon-Reviews-2023}}. Our experimental framework utilises three functionally distinct dataset categories: performance evaluation, fairness evaluation and calibration datasets.

Performance evaluation datasets quantitatively assess summarisation quality preservation using established automatic performance evaluation metrics mentioned in Section~\ref{sec:evaluation} against gold-standard reference summaries. For political tweet summarisation, we employ the manually validated test set from the political partition of MOS~\cite{bilal2022template}, which contains coherent opinion collections for evaluating summarisation performance on political discourse. For product review summarisation, we utilise the established test sets from FewSum~\cite{bravzinskas2022efficient} and AmaSum~\cite{bravzinskas2021learning}, which provide human-annotated reference summaries for comprehensive quality assessment.
 
Fairness evaluation datasets measure opinion representation balance through fairness metrics mentioned in Section~\ref{sec:evaluation}. We construct separate test sets specifically designed to measure bias, built from FairSumm~\cite{dash2019summarizing} for political tweet fairness evaluation and Amazon Reviews 2023~\cite{hou2024bridging} for product review fairness evaluation. These fairness evaluation datasets are structurally independent from the performance evaluation test sets described above, as fairness assessment requires balanced opinion distributions rather than reference summaries. Complete fairness dataset construction methodology is documented in Appendix~\ref{sec:fairness_evaluation_test_set}.
 
Calibration datasets inform parameter selection during the pruning process and are constructed from the aforementioned source datasets using methodology detailed in Section~\ref{sec:diff_set_construct}. These serve exclusively as algorithmic guidance for pruning rather than evaluation benchmarks. 

% We include the following datasets: (1) MOS \cite{bilal2022template} (2) FewSum \cite{bravzinskas2022efficient} (3) AmaSum \cite{bravzinskas2021learning} (4) FairSumm \cite{dash2019summarizing} (5) Amazon reviews 2023 \cite{hou2024bridging}\footnote{\url{https://huggingface.co/datasets/McAuley-Lab/Amazon-Reviews-2023}}. 

% For model performance evaluation, we use the test set that has been manually confirmed as coherent and contains opinions on the political partition of MOS to test model performance in summarising political tweets. For review summarisation, we use the test sets in FewSum and AmaSum. 

% For fairness evaluation, we create separate test sets specifically designed to measure bias. These fairness test sets are built from FairSumm (for political tweets) and Amazon Reviews 2023 (for product reviews). Note that these are different from the performance evaluation test sets described above. Details about the generation process are provided in Appendix~\ref{sec:fairness_evaluation_test_set}.

\subsection{Models}
We use three popular open-source LLMs: (1) TinyLlama \cite{zhang2024tinyllama} (2) Gemma \cite{team2024gemma} (3) Llama 3 \cite{dubey2024llama}. We use both base models (Llama 3 and TinyLlama) as well as instruction-tuned models (Gemma). We select these models to cover a diverse range of model sizes ranging from 1.1B, 2B and 8B. The prompt template we use is in Section~\ref{sec:promt}.

\subsection{Baseline Pruning Methods}
\label{sec:pruning_methods}
We compare different SOTA post-training pruning methods and their effects on fairness, including: (1) Magnitude \cite{han2015learning}, (2) SparseGPT \cite{jakesch2023co}, (3) Wanda \cite{sun2023simple}, and (4) GBLM-Pruner \cite{das2023beyond}.

We perform unstructured pruning for all pruning methods since it offers finer control over weight retention, allowing us to preserve crucial features that may be important for fair representation of minority groups.
For magnitude, SparseGPT and Wanda we use the implementation provided by \citet{sun2023simple}\footnote{\url{https://github.com/locuslab/wanda}}. For GBLM-Pruner, we use the implementation provided by \citet{das2023beyond}\footnote{\url{https://github.com/VILA-Lab/GBLM-Pruner}}, we use both the gradient only version denoted as GBLM-Gradient and gradient with activation version denoted as GBLM-Pruner for the remaining of the paper. We use the $\ell_2$ norm of the gradients and the scaling factor $\alpha$, we use a value of 100 as suggested in GBLM-Pruner \cite{das2023beyond} to account for the small magnitude of gradients when combining activations and gradients. 

\subsection{Calibration Sets}
\label{sec:diff_set_construct}
Each calibration set consists of 128 input collections, with each collection containing either 30 political tweets from FairSumm~\cite{dash2019summarizing} or 8 reviews from Amazon Reviews 2023~\cite{hou2024bridging}.
From the \textit{input perspective}, we directly utilise labels to construct calibration sets.
From the \textit{output perspective}, we randomly construct 50,000 input collections and use the vanilla model to generate summaries. We then construct the calibration set by sampling based on the output SPD according to specific conditions. The descriptions of each calibration set we construct are as follows:
% From the \textit{input perspective}, we construct sets that contain only single-sided or biased information by selecting input from the same side only until the target length is reached. The second type of calibration set represents fair input where information from both sides is included in an equal proportion, i.e., selecting 8 reviews for the same product, 4 with positive opinions and 4 with negative opinions. The third type includes a mixture of fair and biased input, where half of the calibration set i.e., out of 128 input clusters 64 input clusters are biased and the remaining 64 input clusters are fair, including equal information from both sides. From the \textit{output perspective}, we randomly construct 50,000 input clusters and then use the vanilla model to generate summaries. Based on the output SPD, we construct the calibration set by sampling according to specific conditions. Similar to the input perspective, we generated biased, fair, and a mixture of biased and fair calibration sets from the output perspective.
    
\begin{itemize}
    \item \textbf{Single-sided input} contains only single-sided or biased information by selecting input from the one only until the target length is reached.
    \item \textbf{Fair input} contains information from both sides in equal proportion (i.e., selecting 8 reviews for the same product: 4 with positive opinions and 4 with negative opinions).
    \item \textbf{Mixed input} includes a balanced mix of fair and biased inputs, with 128 total input collections. Half of these contain single-sided information, while the other half contain balanced information from both perspectives.
    \item \textbf{Biased output} based on the output SPD, we construct the calibration set by selecting inputs that produce extreme SPD values (i.e., positive or negative 1) using the vanilla model.
    \item \textbf{Fair output} based on the output SPD, we construct the calibration set by selecting inputs that produce fair SPD values (i.e., SPD is 0) using the vanilla model.
    \item \textbf{Mixed output} based on the output SPD, we create a mixed calibration set that includes inputs producing both fair and biased outputs. Half of the calibration set produces biased output, while the remaining half produces fair output, as determined using the vanilla model.
\end{itemize}

\subsection{Evaluation Metrics}
\label{sec:evaluation}
\noindent\textbf{Performance:} the quality of the generated summaries is evaluated through comparison with the corresponding reference summaries using two automatic evaluation metrics, ROUGE \cite{lin2004rouge}, an n-gram overlapping metric, and, BERTScore \cite{zhang2019bertscore}, an embedding-based evaluation metric. For BERTScore, we use the F1 measure. For ROUGE, we use ROUGE-1 and ROUGE-2 for unigrams and bigrams, alongside ROUGE-L, which measures the longest common overlapped sequence between generated and reference summaries. 

\noindent\textbf{Fairness:} A fair model should represent diverse groups equally or proportionally to their population. We evaluate model fairness using multiple metrics: Second-Order SPD (SPD)~\cite{huang-etal-2024-bias}, Binary Unfair Rate (BUR), Unfair Error Rate (UER), and Second-Order Fairness (SOF) \cite{zhang2023fair}.

\begin{itemize}
    \item \textbf{Second-Order SPD (SPD)}~\cite{huang-etal-2024-bias}: Evaluates sentence-level social attributes using a fine-tuned model, comparing distributions between summaries and source documents.
    \item \textbf{Binary Unfair Rate (BUR)}~\cite{zhang2023fair}: Measures overall fairness by calculating the ratio of fair summaries to the total number of generated summaries. 
    \item \textbf{Unfair Error Rate (UER)}: Evaluates underrepresentation in the generated summaries by calculating the average difference between target and generated social value distributions. 
    \item \textbf{Second-Order Fairness (SOF)}: Assesses the spread of unfairness across different values by calculating the variance of UER across all values, which highlights which values are subjected to more unfairness in each sample. 
\end{itemize}

To calculate SPD, we classify sentences in the generated summaries using steps and models described in Section \ref{sec:fairness_evaluation_classification}, then compare proportions of different opinions.
For BUR, UER, and SOF, we follow the methodology proposed by~\citet{zhang2023fair}, using the average of n-gram, BERTScore~\cite{zhang2019bertscore}, and BARTScore~\cite{yuan2021bartscore} matching. We use the publicly available implementation\footnote{\url{https://github.com/psunlpgroup/FairSumm}}.

\begin{figure*}[tbp]
    \centering
    \begin{subfigure}[t]{0.48\textwidth}
        \centering
        \includegraphics[width=\linewidth]{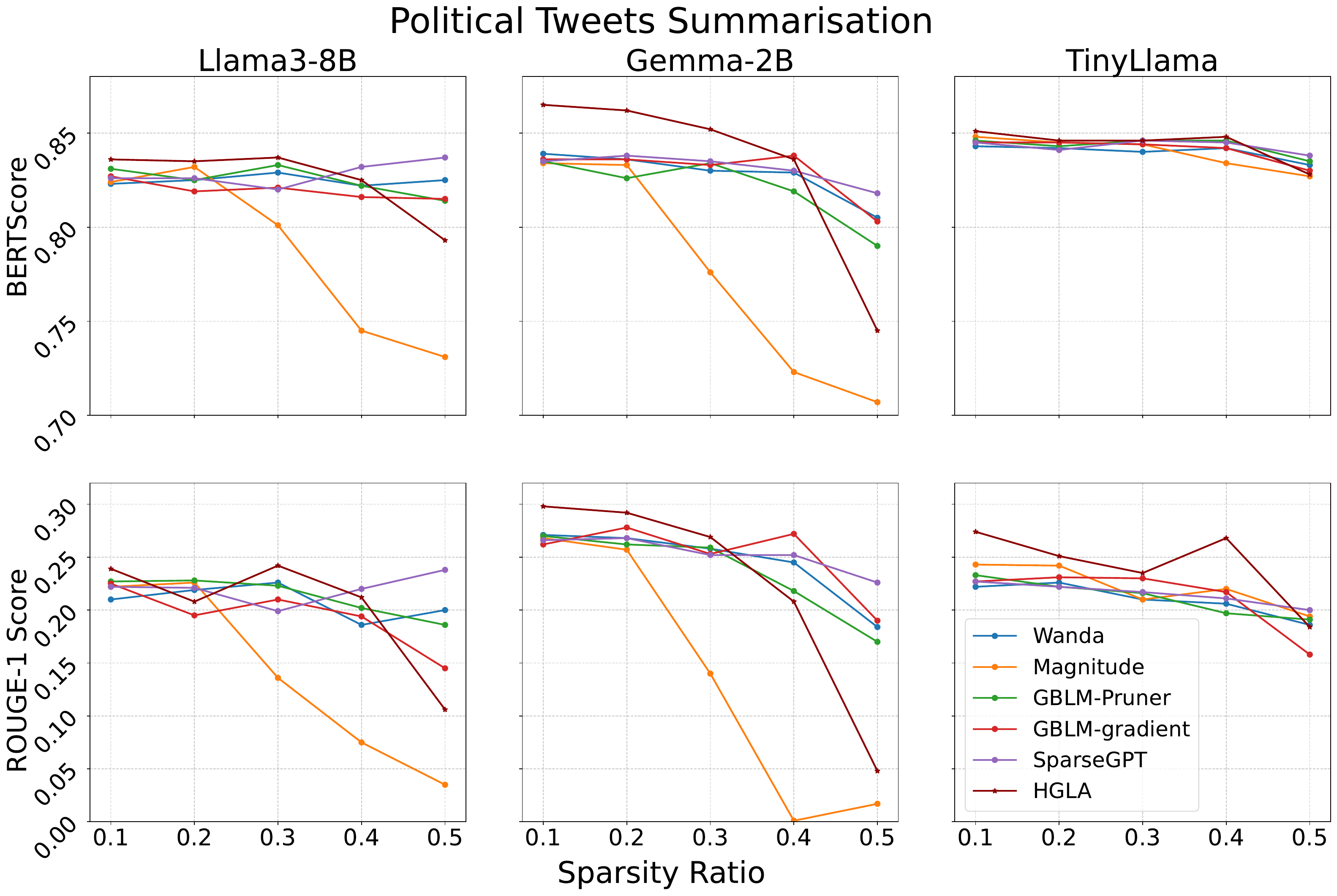}
        \caption{Political tweet summarisation}
        \label{fig:random_political_fairness}
    \end{subfigure}
    \hfill
    \begin{subfigure}[t]{0.48\textwidth}
        \centering
        \includegraphics[width=\linewidth]{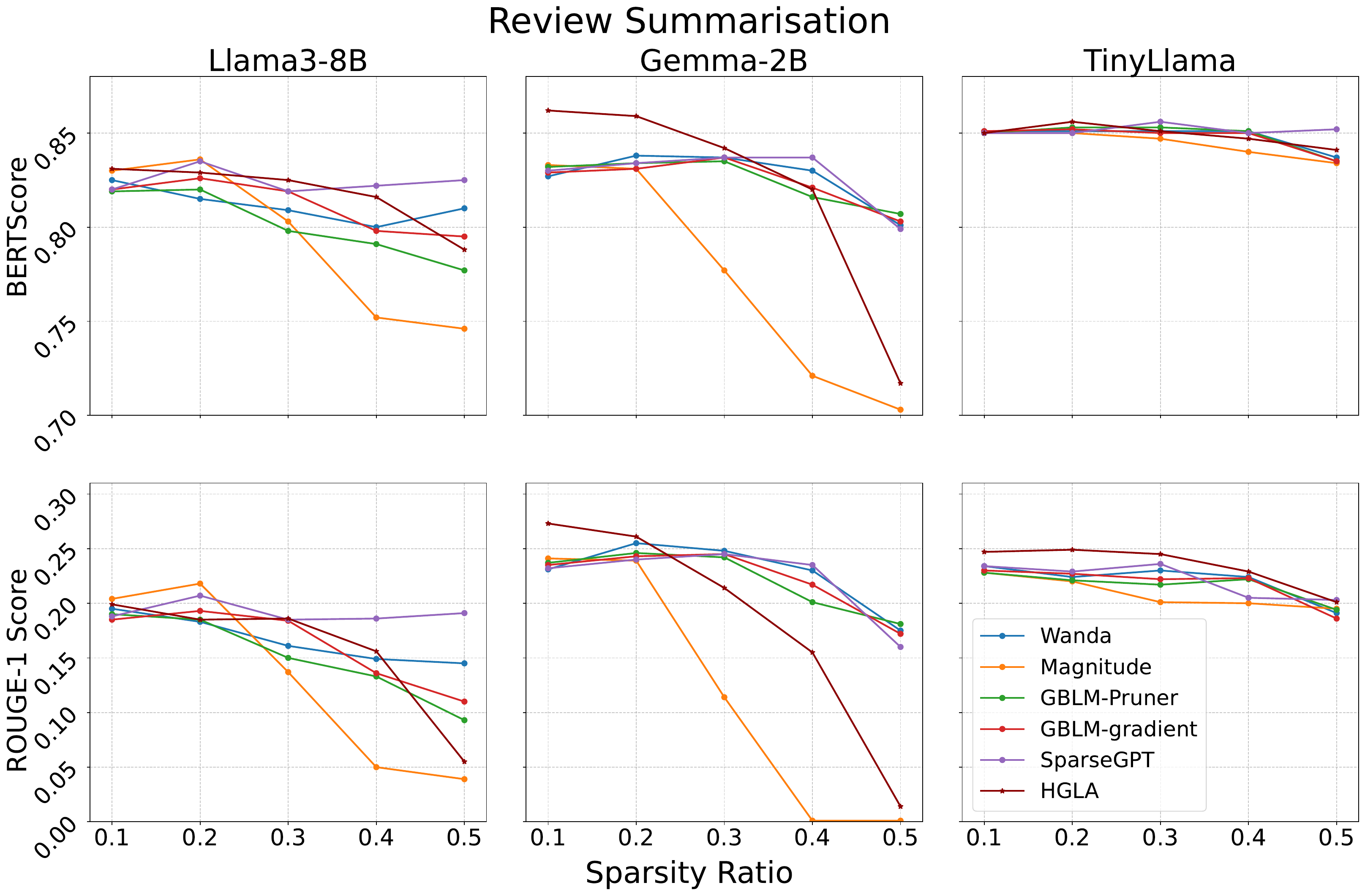}
        \caption{Review summarisation}
        \label{fig:random_review_fairness}
    \end{subfigure}
    \caption{BERTScore and ROUGE-1 for different pruning methods across political tweet summarisation and review summarisation. Apart from TinyLlama, models using magnitude pruning show significant degradation after a 20\% sparsity ratio. SparseGPT generally offers the most robust model performance.
}
    \label{fig:combined_random_fairness}
\end{figure*}

\subsection{Implementation Details}
We use the model implementation and weights available from Hugging Face \cite{wolf-etal-2020-transformers}. We perform experiments using either one or two NVIDIA A100 (40GB) GPUs. For the pruning methods, we use the hyperparameters provided by \citet{frantar2023sparsegpt}, \citet{sun2023simple} and \citet{das2023beyond}.
% For all pruning procedures, we perform unstructured pruning.
For calibration sets, we make a slight modification, instead of using an input length of 2048 tokens, we use 512 since we observed that most of the input lengths in summarising political tweets and reviews are around 512 rather than 2048.

\section{Results and Discussion}

\subsection{Pruning and Summarisation Performance}
\label{sec:pruning_summarisation_performance}
We evaluate model performance using ROUGE 1 \cite{lin2004rouge} and BERTScore \cite{zhang2019bertscore} with random calibration sets (construction details in Section~\ref{sec:random_set_construct}). Results are visualised in Figure~\ref{fig:combined_random_fairness}, with full details in Appendix~\ref{sec:model_performance_ramdom_calibration} and performance-fairness tradeoffs in Figure~\ref{fig:performance_fairness_tradeoff}.
Magnitude pruning \cite{han2015learning} shows significant degradation compared to other methods, while SparseGPT \cite{jakesch2023co} maintains performance best at 50\% sparsity, except for Gemma-2B \cite{team2024gemma} on review summarisation. Given performance degradation at 50\% sparsity across most methods, we eliminate magnitude pruning and cap sparsity at 40\% for fairness comparisons.

\subsection{Pruning, Calibration Set, and Fairness}
\label{sec:pruning_calibration_fairness}
Previous studies found that calibration sets used in pruning methods play an important role in model performance \cite{williams2024impact}. Therefore, to examine model fairness, we use the curated calibration sets mentioned in Section~\ref{sec:diff_set_construct}. 

\begin{figure*}[htbp]
    \centering
    \begin{subfigure}[t]{0.45\textwidth}
        \centering
        \includegraphics[width=0.95\linewidth]{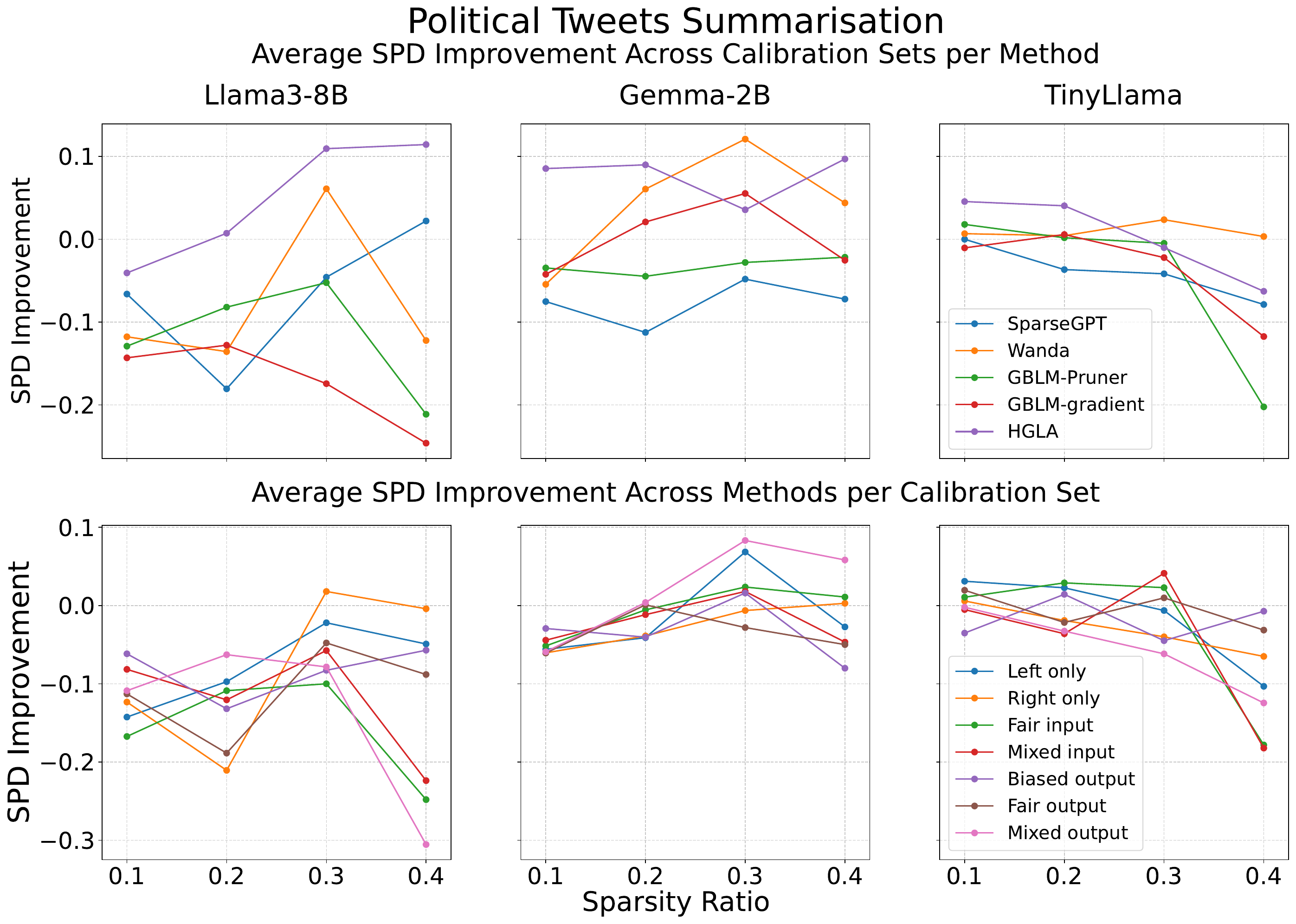}
        \caption{Political tweet summarisation - SPD}
        \label{fig:political_by_method_calibration}
    \end{subfigure}
    \hfill
    \begin{subfigure}[t]{0.45\textwidth}
        \centering
        \includegraphics[width=0.95\linewidth]{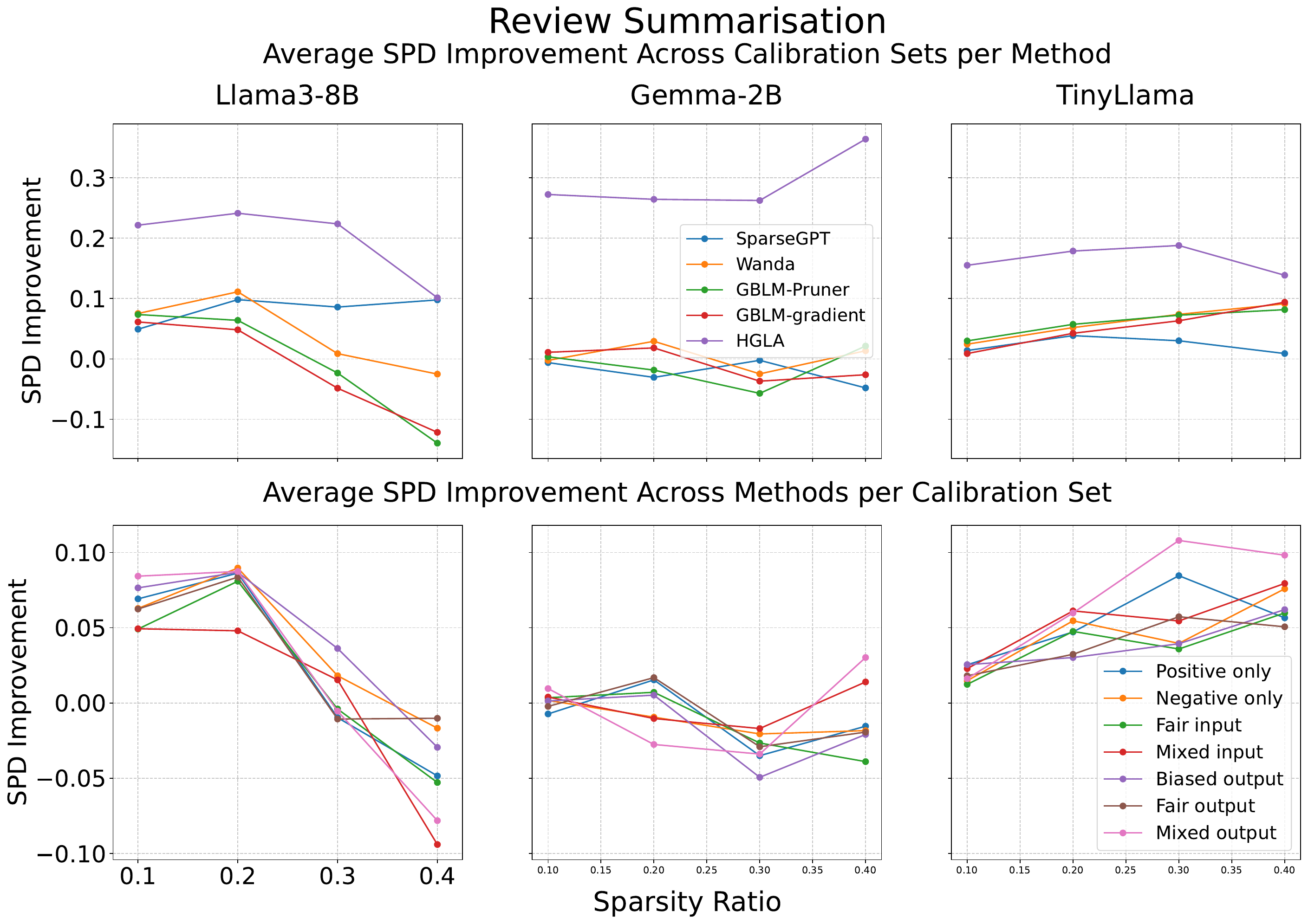}
        \caption{Review summarisation - SPD}
        \label{fig:review_by_method_calibration}
    \end{subfigure}
    
    \centering
    
    \begin{subfigure}[t]{0.45\textwidth}
        \centering
        \includegraphics[width=0.95\linewidth]{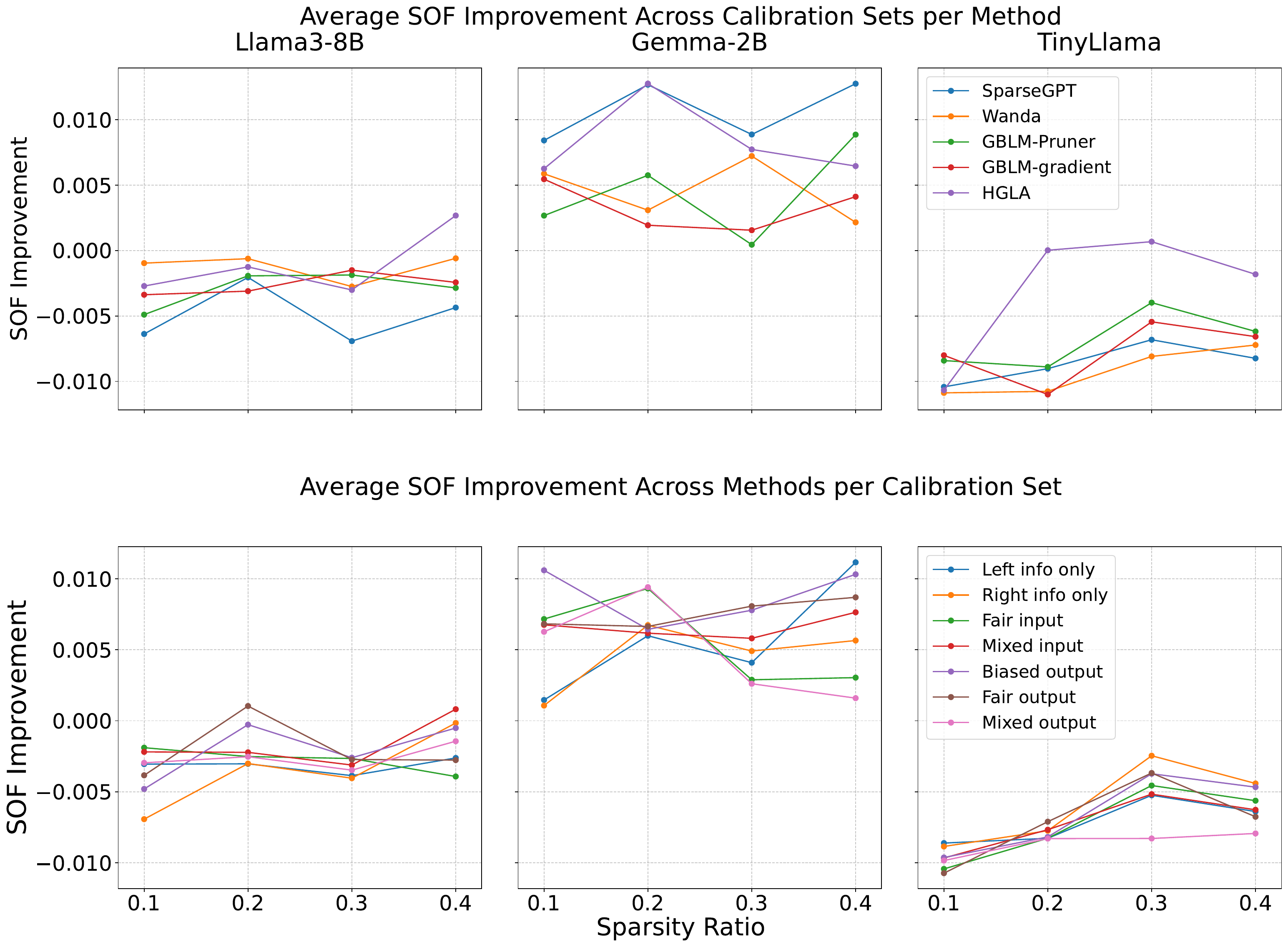}
        \caption{Political tweet summarisation - SOF}
        \label{fig:political_by_method_calibration_sof}
    \end{subfigure}
    \hfill
    \begin{subfigure}[t]{0.45\textwidth}
        \centering
        \includegraphics[width=0.95\linewidth]{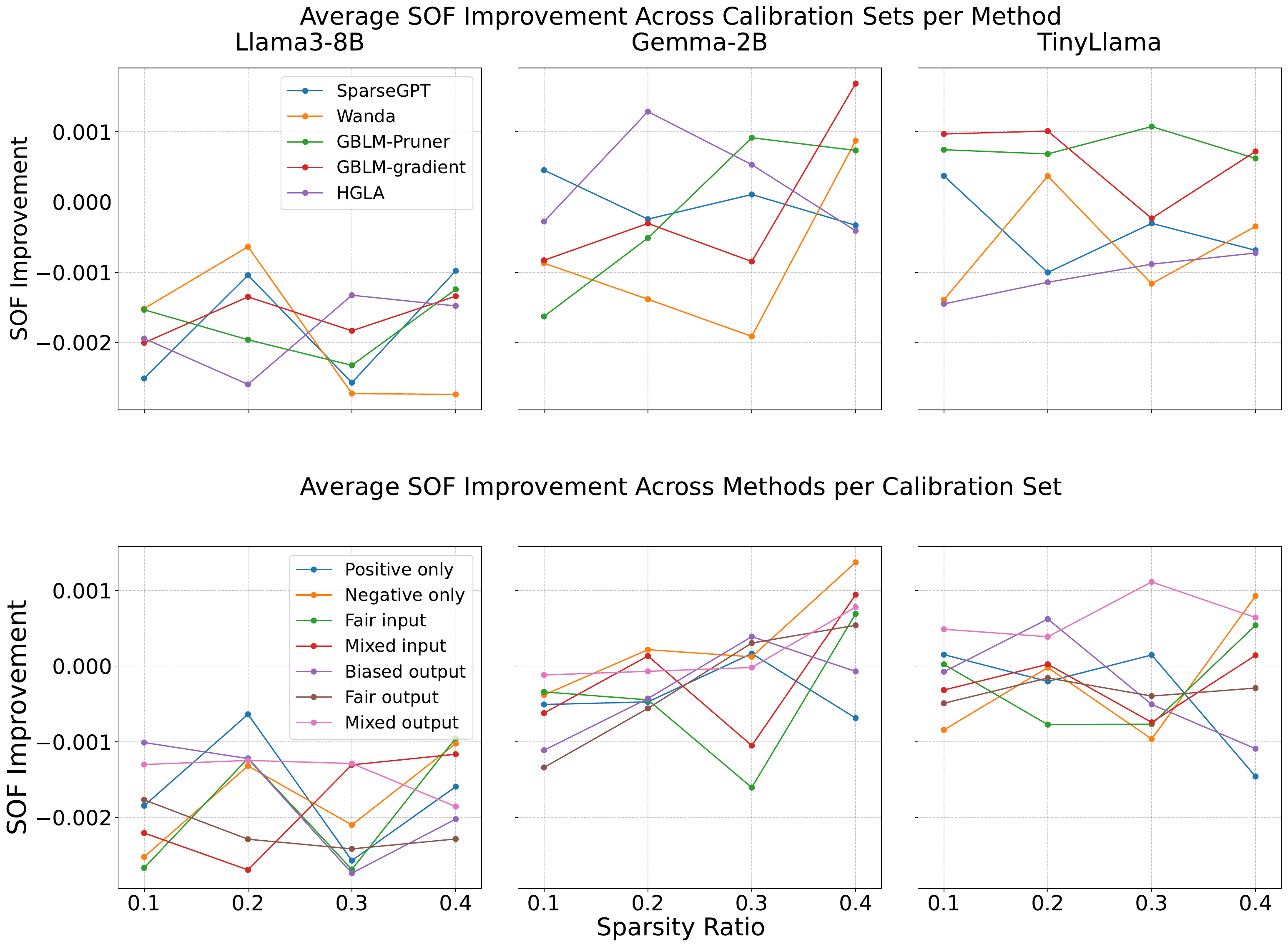}
        \caption{Review summarisation - SOF}
        \label{fig:review_by_method_calibration_sof}
    \end{subfigure}
    
    \begin{subfigure}[t]{0.45\textwidth}
        \centering
        \includegraphics[width=0.95\linewidth]{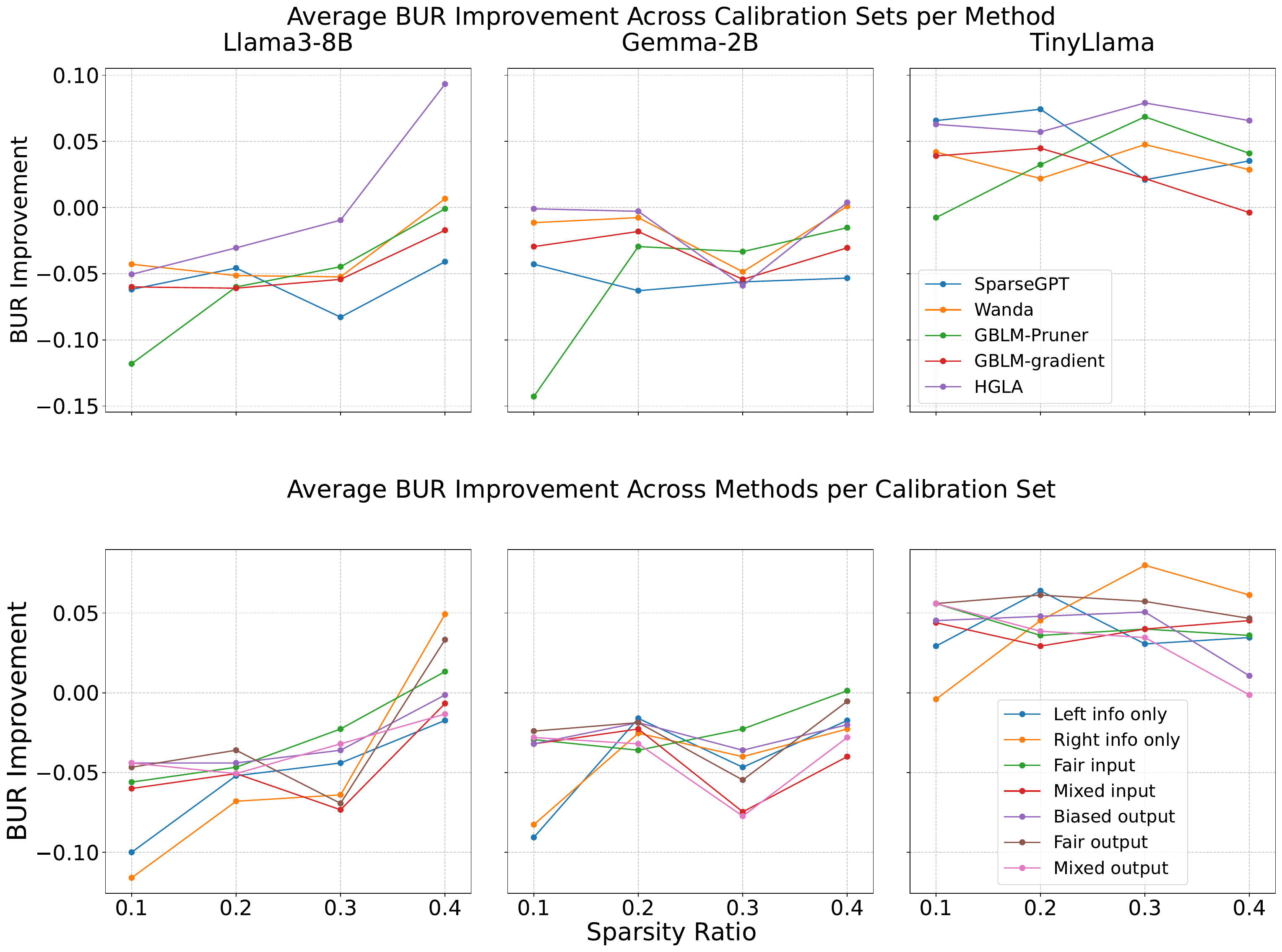}
        \caption{Political tweet summarisation - BUR}
        \label{fig:political_by_method_calibration_bur}
    \end{subfigure}
    \hfill
    \begin{subfigure}[t]{0.45\textwidth}
        \centering
        \includegraphics[width=0.95\linewidth]{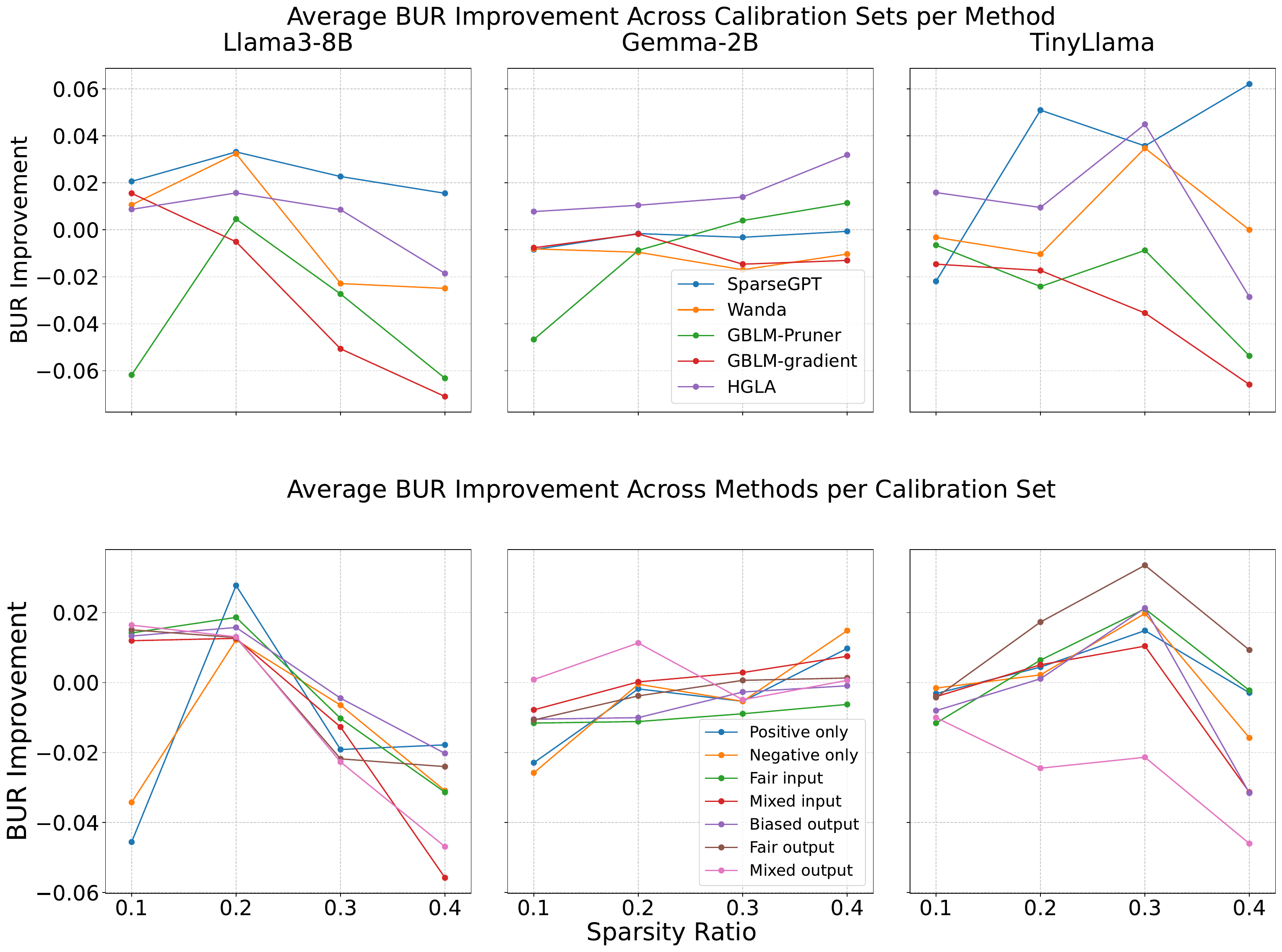}
        \caption{Review summarisation - BUR}
        \label{fig:review_by_method_calibration_bur}
    \end{subfigure}
    
    \begin{subfigure}[t]{0.45\textwidth}
        \centering
        \includegraphics[width=0.95\linewidth]{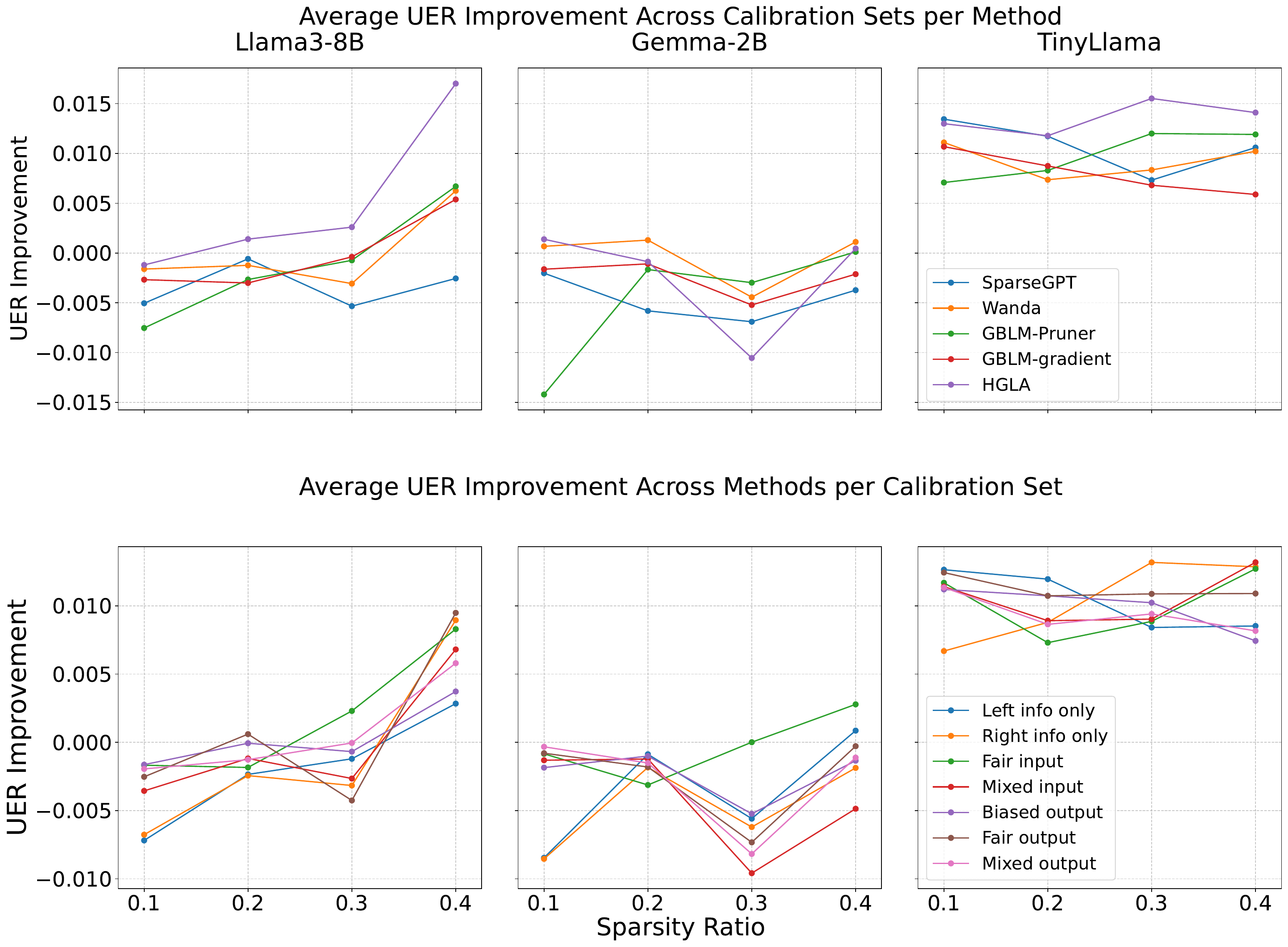}
        \caption{Political tweet summarisation - UER}
        \label{fig:political_by_method_calibration_uer}
    \end{subfigure}
    \hfill
    \begin{subfigure}[t]{0.45\textwidth}
        \centering
        \includegraphics[width=0.95\linewidth]{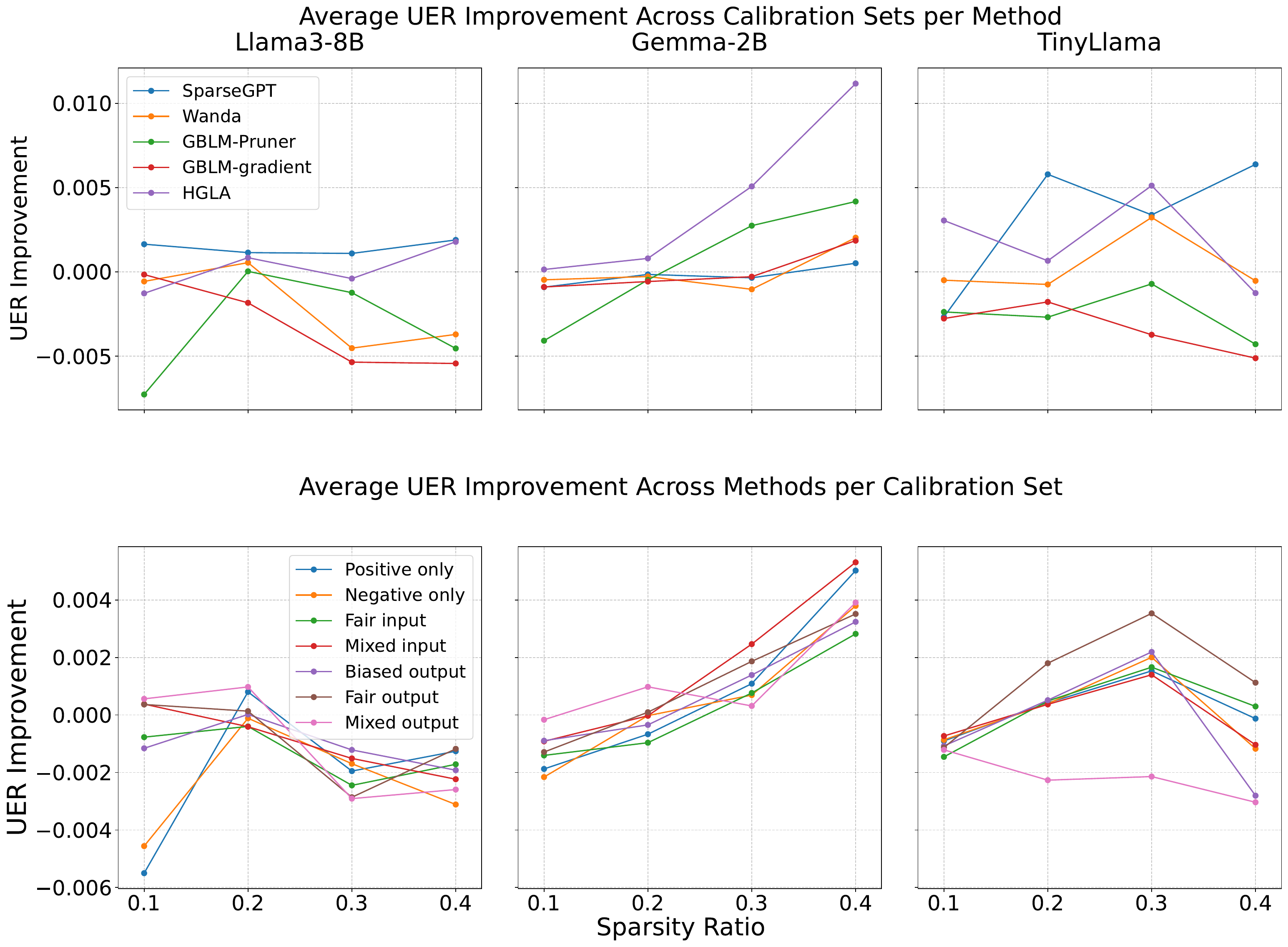}
        \caption{Review summarisation - UER}
        \label{fig:review_by_method_calibration_uer}
    \end{subfigure}
    \caption{Impact on fairness from pruning methods and calibration sets. The figures in the first row show the average model fairness improvement across calibration sets for each pruning method. The figures in the second row display the average fairness improvement across pruning methods for each calibration set. We observe that the pruning method has a bigger impact on model fairness than calibration sets. }
    \label{fig:combined_by_method_calibration}
\end{figure*}

% In our experiments, we evaluate model fairness using multiple metrics mentioned in Section~\ref{sec:evaluation}. We analyse these metrics for both vanilla and pruned models using equal inputs where we expect equal outputs. 
% We evaluate fairness improvements using these metrics. The improvement is quantified by calculating the absolute difference between the vanilla model's and pruned model's metrics. A model demonstrating positive impact on fairness should exhibit differences fall between zero and the metrics value of its corresponding vanilla model.
% To provide a comprehensive analysis and visualise the complex interplay between pruning procedures, calibration sets, and pruning ratios, we present results in Figure~\ref{fig:combined_by_method_calibration} in two ways: averaging across different pruning procedures for each calibration set, and averaging across calibration sets for each pruning procedure.
% Detailed model performance and fairness results are provided in Appendix~\ref{sec:model_performance_various_prune_calibration} and Appendix~\ref{sec:model_fairness_various_prune_calibration}.
We evaluate fairness using multiple metrics mentioned in Section~\ref{sec:evaluation} on both vanilla and pruned models, quantifying improvements through the absolute difference between their respective metrics. Improvements are considered positive when this difference falls between zero and the vanilla model's metric value. 
To provide a comprehensive analysis and visualise the complex interplay between pruning methods, calibration sets, and pruning ratios, we present results in Figure~\ref{fig:combined_by_method_calibration} in two ways: averaging across different pruning methods for each calibration set, and averaging across calibration sets for each pruning method.
Detailed model performance and fairness results are provided in Appendix~\ref{sec:model_performance_various_prune_calibration} and Appendix~\ref{sec:model_fairness_various_prune_calibration}.

From the calibration set perspective, our analysis reveals interesting patterns in fairness improvements. When examining the average improvement across methods per calibration set (shown in the first row of each sub-graph in Figure~\ref{fig:combined_by_method_calibration}), we observe that the trend lines from different calibration sets (shown in the second row of each sub-graph in Figure~\ref{fig:combined_by_method_calibration}) exhibit frequent intersections and maintain close proximity to each other. 
The variations across calibration sets are small, while differences across pruning methods are more pronounced, indicating that pruning methods have a greater impact on model fairness. This observation is quantitatively supported by variance analysis detailed in Appendix~\ref{sec:var_pruning_calibration}.

% From the pruning method perspective, increasing sparsity ratios generally leads to decreased model fairness across different metrics in both political tweet and review summarisation. Wanda is the preferred pruning method because it demonstrates more stable performance across different sparsity ratios, maintains better or less degradation in fairness scores compared to other methods, and shows consistent behaviour across both tasks and different models, offering a better balance between pruning and maintaining fairness compared to other methods.
From the pruning method perspective, increasing sparsity ratios generally leads to decreased model fairness across different metrics in both political tweet and review summarisation. 
% Examining the y-axis values, which represent enhanced fairness metrics, reveals HGLA's particular strength in summarisation while maintaining competitive performance across varying sparsity ratios. 
Our analysis demonstrates that HGLA consistently outperforms or matches alternative methods for fairness improvement across different summarisation tasks and calibration sets (evidenced by its position in the top two performance lines across metrics and summarisation datasets).
Wanda also demonstrates competitive performance as a pruning method, exhibiting stable results across different sparsity ratios compared to other approaches. It maintains better fairness scores or shows less degradation as sparsity increases.

Overall, the relationship between model pruning and fairness shows general trends towards decreased fairness with increased pruning across summarisation tasks and metrics.
The magnitude of this decline varies across tasks, methods, and evaluation metrics, with HGLA demonstrating enhanced fairness preservation and Wanda demonstrating more stable performance compared to other pruning methods.
% However, this pattern varies across tasks, methods, and evaluation metrics, with Wanda demonstrating more stable performance compared to other pruning methods. 
Suggesting these factors interact in complex and sometimes counterintuitive ways, with certain combinations of pruning methods and tasks showing unexpected preservation or degradation of fairness metrics. The results indicate that pruning without awareness of fairness implications could potentially harm model fairness, highlighting the importance of careful pruning method selection and continuous monitoring of fairness metrics during the pruning process.

In contrast to previous study ~\cite{chrysostomou2024investigating}, which found that increased pruning reduced hallucination by making models rely more on their original input, our findings suggest a different pattern for opinionated text summarisation. We found that pruning more does not necessarily lead to less biased summaries of opinionated text. One possible explanation is that model compression can cause models to forget or perform worse on minority classes and edge cases, which may amplify existing biases or create new disparities across different social groups~\cite{hooker2019compressed, hooker2020characterising, misra2024uncovering}. Without explicit guidance from fairness metrics during pruning, unconstrained pruning could potentially compromise model fairness~\cite{zayed2024fairness}.

\subsection{Fairness Pruned with HGLA}

\begin{table*}[htbp]
\definecolor{veryLightGreen}{rgb}{0.85, 0.95, 0.85}
\definecolor{lightGreen}{rgb}{0.70, 0.85, 0.70}
\definecolor{mediumGreen}{rgb}{0.40, 0.70, 0.40}
\definecolor{mediumDarkGreen}{rgb}{0.20, 0.60, 0.20}
\definecolor{darkGreen}{rgb}{0.07, 0.53, 0.03}

\newcommand{\hlc}[2]{%
  \ifdim#1pt>0pt
    \cellcolor{%
      \ifnum\pdfstrcmp{\fpeval{#1/#2}}{\fpeval{0.2}}<0 veryLightGreen\else
      \ifnum\pdfstrcmp{\fpeval{#1/#2}}{\fpeval{0.4}}<0 lightGreen\else
      \ifnum\pdfstrcmp{\fpeval{#1/#2}}{\fpeval{0.6}}<0 mediumGreen\else
      \ifnum\pdfstrcmp{\fpeval{#1/#2}}{\fpeval{0.8}}<0 mediumDarkGreen\else
      darkGreen\fi\fi\fi\fi
    }%
  \fi
  #1%
}
\centering
\tiny
\begin{subtable}[t]{\textwidth}
    \centering
    \resizebox{\textwidth}{!}{
    \begin{tabular}{c cccccccc cccccccc cccccccc}
    \toprule
    & \multicolumn{8}{c}{Llama3-8B } & \multicolumn{8}{c}{Gemma-2B} & \multicolumn{8}{c}{TinyLlama} \\
    \cmidrule(lr){2-9} \cmidrule(lr){10-17} \cmidrule(lr){18-25}
    & \multicolumn{2}{c}{SPD} & \multicolumn{2}{c}{SOF} & \multicolumn{2}{c}{BUR} & \multicolumn{2}{c}{UER} & 
      \multicolumn{2}{c}{SPD} & \multicolumn{2}{c}{SOF} & \multicolumn{2}{c}{BUR} & \multicolumn{2}{c}{UER} &
      \multicolumn{2}{c}{SPD} & \multicolumn{2}{c}{SOF} & \multicolumn{2}{c}{BUR} & \multicolumn{2}{c}{UER} \\
    \cmidrule(lr){2-3} \cmidrule(lr){4-5} \cmidrule(lr){6-7} \cmidrule(lr){8-9}
    \cmidrule(lr){10-11} \cmidrule(lr){12-13} \cmidrule(lr){14-15} \cmidrule(lr){16-17}
    \cmidrule(lr){18-19} \cmidrule(lr){20-21} \cmidrule(lr){22-23} \cmidrule(lr){24-25}
    Ratio & Wanda & HGLA & Wanda & HGLA & Wanda & HGLA & Wanda & HGLA & 
           Wanda & HGLA & Wanda & HGLA & Wanda & HGLA & Wanda & HGLA & 
           Wanda & HGLA & Wanda & HGLA & Wanda & HGLA & Wanda & HGLA \\
    \midrule
    0.1 & \hlc{-0.106}{0.187} & \hlc{-0.060}{0.187} & \hlc{-0.005}{0.011} & \hlc{-0.006}{0.011} & \hlc{-0.113}{0.240} & \hlc{-0.107}{0.240} & \hlc{-0.005}{0.036} & \hlc{0.006}{0.036} & 
         \hlc{-0.060}{0.270} & \hlc{-0.046}{0.270} & \hlc{0.005}{0.013} & \hlc{0.005}{0.013} & \hlc{0.000}{0.393} & \hlc{0.013}{0.393} & \hlc{0.000}{0.058} & \hlc{0.002}{0.058} & 
         \hlc{-0.013}{0.088} & \hlc{0.038}{0.088} & \hlc{-0.012}{0.015} & \hlc{-0.007}{0.015} & \hlc{0.027}{0.253} & \hlc{0.040}{0.253} & \hlc{0.013}{0.037} & \hlc{0.011}{0.037} \\
    0.2 & \hlc{-0.142}{0.187} & \hlc{-0.004}{0.187} & \hlc{-0.001}{0.011} & \hlc{-0.004}{0.011} & \hlc{-0.053}{0.240} & \hlc{-0.040}{0.240} & \hlc{-0.004}{0.036} & \hlc{0.000}{0.036} & 
         \hlc{0.043}{0.270} & \hlc{0.045}{0.270} & \hlc{0.007}{0.013} & \hlc{0.010}{0.013} & \hlc{0.007}{0.393} & \hlc{-0.027}{0.393} & \hlc{0.001}{0.058} & \hlc{-0.002}{0.058} & 
         \hlc{-0.077}{0.088} & \hlc{0.077}{0.088} & \hlc{-0.004}{0.015} & \hlc{0.003}{0.015} & \hlc{0.013}{0.253} & \hlc{0.040}{0.253} & \hlc{0.006}{0.037} & \hlc{0.010}{0.037} \\
    0.3 & \hlc{0.082}{0.187} & \hlc{0.119}{0.187} & \hlc{-0.003}{0.011} & \hlc{-0.005}{0.011} & \hlc{-0.007}{0.240} & \hlc{-0.060}{0.240} & \hlc{-0.005}{0.036} & \hlc{-0.001}{0.036} & 
         \hlc{0.225}{0.270} & \hlc{-0.052}{0.270} & \hlc{0.007}{0.013} & \hlc{0.004}{0.013} & \hlc{-0.040}{0.393} & \hlc{-0.053}{0.393} & \hlc{0.002}{0.058} & \hlc{-0.013}{0.058} & 
         \hlc{0.006}{0.088} & \hlc{-0.111}{0.088} & \hlc{-0.010}{0.015} & \hlc{0.004}{0.015} & \hlc{0.013}{0.253} & \hlc{0.147}{0.253} & \hlc{0.008}{0.037} & \hlc{0.024}{0.037} \\
    0.4 & \hlc{-0.128}{0.187} & \hlc{0.170}{0.187} & \hlc{0.003}{0.011} & \hlc{0.003}{0.011} & \hlc{-0.020}{0.240} & \hlc{0.100}{0.240} & \hlc{0.008}{0.036} & \hlc{0.015}{0.036} & 
         \hlc{0.038}{0.270} & \hlc{0.112}{0.270} & \hlc{-0.003}{0.013} & \hlc{0.015}{0.013} & \hlc{0.013}{0.393} & \hlc{0.047}{0.393} & \hlc{-0.006}{0.058} & \hlc{0.004}{0.058} & 
         \hlc{0.006}{0.088} & \hlc{-0.258}{0.088} & \hlc{-0.003}{0.015} & \hlc{0.010}{0.015} & \hlc{0.020}{0.253} & \hlc{0.067}{0.253} & \hlc{0.015}{0.037} & \hlc{0.011}{0.037} \\
    \bottomrule
    \end{tabular}
    }
    \caption{Political - Right information only pruned}
\end{subtable}

\begin{subtable}[t]{\textwidth}
    \centering
    \resizebox{\textwidth}{!}{
    \begin{tabular}{c cccccccc cccccccc cccccccc}
    \toprule
    & \multicolumn{8}{c}{Llama3-8B} & \multicolumn{8}{c}{Gemma-2B} & \multicolumn{8}{c}{TinyLlama} \\
    \cmidrule(lr){2-9} \cmidrule(lr){10-17} \cmidrule(lr){18-25}
    & \multicolumn{2}{c}{SPD} & \multicolumn{2}{c}{SOF} & \multicolumn{2}{c}{BUR} & \multicolumn{2}{c}{UER} & 
      \multicolumn{2}{c}{SPD} & \multicolumn{2}{c}{SOF} & \multicolumn{2}{c}{BUR} & \multicolumn{2}{c}{UER} &
      \multicolumn{2}{c}{SPD} & \multicolumn{2}{c}{SOF} & \multicolumn{2}{c}{BUR} & \multicolumn{2}{c}{UER} \\
    \cmidrule(lr){2-3} \cmidrule(lr){4-5} \cmidrule(lr){6-7} \cmidrule(lr){8-9}
    \cmidrule(lr){10-11} \cmidrule(lr){12-13} \cmidrule(lr){14-15} \cmidrule(lr){16-17}
    \cmidrule(lr){18-19} \cmidrule(lr){20-21} \cmidrule(lr){22-23} \cmidrule(lr){24-25}
    Ratio & Wanda & HGLA & Wanda & HGLA & Wanda & HGLA & Wanda & HGLA & 
           Wanda & HGLA & Wanda & HGLA & Wanda & HGLA & Wanda & HGLA & 
           Wanda & HGLA & Wanda & HGLA & Wanda & HGLA & Wanda & HGLA \\
    \midrule
    0.1 & \hlc{-0.132}{0.187} & \hlc{-0.232}{0.187} & \hlc{0.001}{0.011} & \hlc{-0.003}{0.011} & \hlc{-0.100}{0.240} & \hlc{-0.033}{0.240} & \hlc{-0.008}{0.036} & \hlc{-0.001}{0.036} & 
         \hlc{-0.014}{0.270} & \hlc{-0.035}{0.270} & \hlc{0.006}{0.013} & \hlc{0.005}{0.013} & \hlc{0.000}{0.393} & \hlc{-0.007}{0.393} & \hlc{0.001}{0.058} & \hlc{0.001}{0.058} & 
         \hlc{0.006}{0.088} & \hlc{0.065}{0.088} & \hlc{-0.012}{0.015} & \hlc{-0.012}{0.015} & \hlc{0.060}{0.253} & \hlc{0.073}{0.253} & \hlc{0.009}{0.037} & \hlc{0.015}{0.037} \\
    0.2 & \hlc{-0.140}{0.187} & \hlc{-0.134}{0.187} & \hlc{-0.004}{0.011} & \hlc{0.004}{0.011} & \hlc{-0.060}{0.240} & \hlc{-0.020}{0.240} & \hlc{-0.002}{0.036} & \hlc{0.002}{0.036} & 
         \hlc{0.039}{0.270} & \hlc{-0.059}{0.270} & \hlc{-0.002}{0.013} & \hlc{0.015}{0.013} & \hlc{0.020}{0.393} & \hlc{0.007}{0.393} & \hlc{0.003}{0.058} & \hlc{0.000}{0.058} & 
         \hlc{-0.025}{0.088} & \hlc{-0.037}{0.088} & \hlc{-0.012}{0.015} & \hlc{-0.003}{0.015} & \hlc{0.040}{0.253} & \hlc{0.073}{0.253} & \hlc{0.005}{0.037} & \hlc{0.015}{0.037} \\
    0.3 & \hlc{0.183}{0.187} & \hlc{0.043}{0.187} & \hlc{-0.006}{0.011} & \hlc{-0.002}{0.011} & \hlc{-0.100}{0.240} & \hlc{0.007}{0.240} & \hlc{0.001}{0.036} & \hlc{0.002}{0.036} & 
         \hlc{0.095}{0.270} & \hlc{-0.205}{0.270} & \hlc{0.007}{0.013} & \hlc{0.009}{0.013} & \hlc{0.000}{0.393} & \hlc{-0.053}{0.393} & \hlc{-0.004}{0.058} & \hlc{-0.008}{0.058} & 
         \hlc{-0.015}{0.088} & \hlc{0.037}{0.088} & \hlc{-0.011}{0.015} & \hlc{-0.007}{0.015} & \hlc{0.073}{0.253} & \hlc{0.067}{0.253} & \hlc{0.006}{0.037} & \hlc{0.010}{0.037} \\
    0.4 & \hlc{-0.156}{0.187} & \hlc{0.092}{0.187} & \hlc{-0.001}{0.011} & \hlc{0.001}{0.011} & \hlc{0.040}{0.240} & \hlc{0.100}{0.240} & \hlc{0.003}{0.036} & \hlc{0.019}{0.036} & 
         \hlc{-0.030}{0.270} & \hlc{-0.153}{0.270} & \hlc{0.004}{0.013} & \hlc{0.017}{0.013} & \hlc{-0.027}{0.393} & \hlc{-0.007}{0.393} & \hlc{0.005}{0.058} & \hlc{0.002}{0.058} & 
         \hlc{-0.082}{0.088} & \hlc{0.040}{0.088} & \hlc{-0.009}{0.015} & \hlc{-0.004}{0.015} & \hlc{0.053}{0.253} & \hlc{0.087}{0.253} & \hlc{0.007}{0.037} & \hlc{0.010}{0.037} \\
    \bottomrule
    \end{tabular}
    }
    \caption{Political - Left information only pruned}
\end{subtable}

\begin{subtable}[t]{\textwidth}
    \centering
    \resizebox{\textwidth}{!}{
    \begin{tabular}{c cccccccc cccccccc cccccccc}
    \toprule
    & \multicolumn{8}{c}{Llama3-8B} & \multicolumn{8}{c}{Gemma-2B} & \multicolumn{8}{c}{TinyLlama} \\
    \cmidrule(lr){2-9} \cmidrule(lr){10-17} \cmidrule(lr){18-25}
    & \multicolumn{2}{c}{SPD} & \multicolumn{2}{c}{SOF} & \multicolumn{2}{c}{BUR} & \multicolumn{2}{c}{UER} & 
      \multicolumn{2}{c}{SPD} & \multicolumn{2}{c}{SOF} & \multicolumn{2}{c}{BUR} & \multicolumn{2}{c}{UER} &
      \multicolumn{2}{c}{SPD} & \multicolumn{2}{c}{SOF} & \multicolumn{2}{c}{BUR} & \multicolumn{2}{c}{UER} \\
    \cmidrule(lr){2-3} \cmidrule(lr){4-5} \cmidrule(lr){6-7} \cmidrule(lr){8-9}
    \cmidrule(lr){10-11} \cmidrule(lr){12-13} \cmidrule(lr){14-15} \cmidrule(lr){16-17}
    \cmidrule(lr){18-19} \cmidrule(lr){20-21} \cmidrule(lr){22-23} \cmidrule(lr){24-25}
    Ratio & Wanda & HGLA & Wanda & HGLA & Wanda & HGLA & Wanda & HGLA & 
           Wanda & HGLA & Wanda & HGLA & Wanda & HGLA & Wanda & HGLA & 
           Wanda & HGLA & Wanda & HGLA & Wanda & HGLA & Wanda & HGLA \\
    \midrule
    0.1 & \hlc{0.081}{0.496} & \hlc{0.020}{0.496} & \hlc{-0.004}{0.005} & \hlc{0.000}{0.005} & \hlc{0.000}{0.387} & \hlc{-0.008}{0.387} & \hlc{-0.001}{0.062} & \hlc{-0.003}{0.062} & 
         \hlc{-0.001}{0.785} & \hlc{-0.013}{0.785} & \hlc{0.000}{0.005} & \hlc{-0.001}{0.005} & \hlc{0.021}{0.297} & \hlc{0.020}{0.297} & \hlc{0.002}{0.048} & \hlc{0.001}{0.048} & 
         \hlc{0.016}{0.382} & \hlc{0.031}{0.382} & \hlc{-0.001}{0.003} & \hlc{-0.001}{0.003} & \hlc{-0.010}{0.413} & \hlc{0.018}{0.413} & \hlc{-0.001}{0.059} & \hlc{0.003}{0.059} \\
    0.2 & \hlc{0.130}{0.496} & \hlc{0.097}{0.496} & \hlc{0.002}{0.005} & \hlc{-0.003}{0.005} & \hlc{0.044}{0.387} & \hlc{0.034}{0.387} & \hlc{0.001}{0.062} & \hlc{0.004}{0.062} & 
         \hlc{0.029}{0.785} & \hlc{-0.028}{0.785} & \hlc{-0.002}{0.005} & \hlc{0.002}{0.005} & \hlc{-0.016}{0.297} & \hlc{0.016}{0.297} & \hlc{-0.001}{0.048} & \hlc{0.000}{0.048} & 
         \hlc{0.045}{0.382} & \hlc{0.077}{0.382} & \hlc{0.001}{0.003} & \hlc{-0.001}{0.003} & \hlc{-0.021}{0.413} & \hlc{0.032}{0.413} & \hlc{-0.002}{0.059} & \hlc{0.002}{0.059} \\
    0.3 & \hlc{-0.004}{0.496} & \hlc{0.077}{0.496} & \hlc{-0.002}{0.005} & \hlc{-0.003}{0.005} & \hlc{-0.039}{0.387} & \hlc{0.016}{0.387} & \hlc{-0.004}{0.062} & \hlc{0.000}{0.062} & 
         \hlc{0.015}{0.785} & \hlc{-0.032}{0.785} & \hlc{-0.003}{0.005} & \hlc{-0.001}{0.005} & \hlc{-0.017}{0.297} & \hlc{0.004}{0.297} & \hlc{0.000}{0.048} & \hlc{0.005}{0.048} & 
         \hlc{0.070}{0.382} & \hlc{0.064}{0.382} & \hlc{-0.001}{0.003} & \hlc{-0.001}{0.003} & \hlc{0.014}{0.413} & \hlc{0.039}{0.413} & \hlc{0.002}{0.059} & \hlc{0.005}{0.059} \\
    0.4 & \hlc{-0.045}{0.496} & \hlc{-0.040}{0.496} & \hlc{-0.003}{0.005} & \hlc{0.001}{0.005} & \hlc{-0.019}{0.387} & \hlc{0.010}{0.387} & \hlc{-0.004}{0.062} & \hlc{0.003}{0.062} & 
         \hlc{0.008}{0.785} & \hlc{0.140}{0.785} & \hlc{-0.003}{0.005} & \hlc{-0.003}{0.005} & \hlc{0.008}{0.297} & \hlc{0.051}{0.297} & \hlc{0.005}{0.048} & \hlc{0.012}{0.048} & 
         \hlc{0.126}{0.382} & \hlc{0.061}{0.382} & \hlc{-0.002}{0.003} & \hlc{-0.001}{0.003} & \hlc{0.010}{0.413} & \hlc{-0.013}{0.413} & \hlc{0.000}{0.059} & \hlc{-0.001}{0.059} \\
    \bottomrule
    \end{tabular}
    }
    \caption{Review - Positive information only pruned}
\end{subtable}

\begin{subtable}[t]{\textwidth}
    \centering
    \resizebox{\textwidth}{!}{
    \begin{tabular}{c cccccccc cccccccc cccccccc}
    \toprule
    & \multicolumn{8}{c}{Llama3-8B} & \multicolumn{8}{c}{Gemma-2B} & \multicolumn{8}{c}{TinyLlama} \\
    \cmidrule(lr){2-9} \cmidrule(lr){10-17} \cmidrule(lr){18-25}
    & \multicolumn{2}{c}{SPD} & \multicolumn{2}{c}{SOF} & \multicolumn{2}{c}{BUR} & \multicolumn{2}{c}{UER} & 
      \multicolumn{2}{c}{SPD} & \multicolumn{2}{c}{SOF} & \multicolumn{2}{c}{BUR} & \multicolumn{2}{c}{UER} &
      \multicolumn{2}{c}{SPD} & \multicolumn{2}{c}{SOF} & \multicolumn{2}{c}{BUR} & \multicolumn{2}{c}{UER} \\
    \cmidrule(lr){2-3} \cmidrule(lr){4-5} \cmidrule(lr){6-7} \cmidrule(lr){8-9}
    \cmidrule(lr){10-11} \cmidrule(lr){12-13} \cmidrule(lr){14-15} \cmidrule(lr){16-17}
    \cmidrule(lr){18-19} \cmidrule(lr){20-21} \cmidrule(lr){22-23} \cmidrule(lr){24-25}
    Ratio & Wanda & HGLA & Wanda & HGLA & Wanda & HGLA & Wanda & HGLA & 
           Wanda & HGLA & Wanda & HGLA & Wanda & HGLA & Wanda & HGLA & 
           Wanda & HGLA & Wanda & HGLA & Wanda & HGLA & Wanda & HGLA \\
    \midrule
    0.1 & \hlc{0.115}{0.496} & \hlc{0.085}{0.496} & \hlc{-0.003}{0.005} & \hlc{-0.003}{0.005} & \hlc{0.003}{0.387} & \hlc{0.010}{0.387} & \hlc{-0.001}{0.062} & \hlc{-0.001}{0.062} & 
         \hlc{0.000}{0.785} & \hlc{-0.013}{0.785} & \hlc{0.001}{0.005} & \hlc{0.001}{0.005} & \hlc{0.020}{0.297} & \hlc{0.003}{0.297} & \hlc{0.001}{0.048} & \hlc{0.000}{0.048} & 
         \hlc{0.040}{0.382} & \hlc{0.029}{0.382} & \hlc{-0.001}{0.003} & \hlc{-0.002}{0.003} & \hlc{0.003}{0.413} & \hlc{0.022}{0.413} & \hlc{0.000}{0.059} & \hlc{0.003}{0.059} \\
    0.2 & \hlc{0.102}{0.496} & \hlc{0.145}{0.496} & \hlc{0.001}{0.005} & \hlc{-0.003}{0.005} & \hlc{0.017}{0.387} & \hlc{0.017}{0.387} & \hlc{-0.001}{0.062} & \hlc{0.000}{0.062} & 
         \hlc{0.033}{0.785} & \hlc{-0.034}{0.785} & \hlc{0.000}{0.005} & \hlc{0.001}{0.005} & \hlc{-0.017}{0.297} & \hlc{0.007}{0.297} & \hlc{-0.001}{0.048} & \hlc{0.000}{0.048} & 
         \hlc{0.038}{0.382} & \hlc{0.080}{0.382} & \hlc{0.001}{0.003} & \hlc{-0.001}{0.003} & \hlc{-0.006}{0.413} & \hlc{-0.004}{0.413} & \hlc{-0.002}{0.059} & \hlc{0.000}{0.059} \\
    0.3 & \hlc{0.001}{0.496} & \hlc{0.114}{0.496} & \hlc{-0.003}{0.005} & \hlc{0.001}{0.005} & \hlc{-0.023}{0.387} & \hlc{0.031}{0.387} & \hlc{-0.004}{0.062} & \hlc{0.001}{0.062} & 
         \hlc{-0.023}{0.785} & \hlc{-0.004}{0.785} & \hlc{0.000}{0.005} & \hlc{-0.003}{0.005} & \hlc{-0.007}{0.297} & \hlc{0.021}{0.297} & \hlc{-0.001}{0.048} & \hlc{0.005}{0.048} & 
         \hlc{0.092}{0.382} & \hlc{0.101}{0.382} & \hlc{-0.002}{0.003} & \hlc{-0.001}{0.003} & \hlc{0.039}{0.413} & \hlc{0.058}{0.413} & \hlc{0.003}{0.059} & \hlc{0.007}{0.059} \\
    0.4 & \hlc{-0.050}{0.496} & \hlc{-0.112}{0.496} & \hlc{-0.003}{0.005} & \hlc{-0.001}{0.005} & \hlc{-0.013}{0.387} & \hlc{-0.018}{0.387} & \hlc{-0.002}{0.062} & \hlc{0.001}{0.062} & 
         \hlc{0.027}{0.785} & \hlc{0.219}{0.785} & \hlc{0.002}{0.005} & \hlc{-0.001}{0.005} & \hlc{-0.016}{0.297} & \hlc{0.033}{0.297} & \hlc{0.001}{0.048} & \hlc{0.007}{0.048} & 
         \hlc{0.072}{0.382} & \hlc{-0.065}{0.382} & \hlc{0.001}{0.003} & \hlc{0.000}{0.003} & \hlc{-0.018}{0.413} & \hlc{-0.013}{0.413} & \hlc{-0.002}{0.059} & \hlc{0.000}{0.059} \\
    \bottomrule
    \end{tabular}
    }
    \caption{Review - Negative information only pruned}
\end{subtable}

\caption{Fairness improvement through pruning based on HGLA and Wanda, darker colours indicating greater fairness improvement. Compared to Wanda \cite{sun2023simple}, HGLA brings greater benefits when using single-sided input opposite to the model's intrinsic bias, enabling deeper pruning (40\% vs 30\% for Llama3-8B) and showing improvements where Wanda produced minimal changes (e.g., Gemma-2B in review summarisation).}

% Fairness improvement through pruning models based on high gradients and low activations, darker colours indicating greater fairness improvement. Compared to Wanda \cite{sun2023simple}, HGLA demonstrates greater advantages across various metrics when using single-sided input, particularly input that opposes the model's intrinsic bias.
% For example, when pruning for political tweet summarisation, Llama3-8B showed greater improvement with HGLA compared to Wanda, allowing pruning up to 40\% rather than 30\%. In review summarisation, while Gemma-2B showed no significant fairness change when using Wanda, HGLA demonstrated substantial improvement.

% For example, when summarising political tweets, Llama3-8B showed greater improvement with HGLA compared to WANDA, allowing for pruning up to 40\% rather than 30\%. The effect on TinyLlama was comparable to that observed in Gemma-2B when using anti-bias information.
% In review summarisation, Gemma-2B showed no significant changes in fairness metrics when using WANDA. However, it demonstrated substantial improvements when pruned using HGLA, while other models maintained similar fairness levels.
\label{tab:hgla_fairness}
\end{table*}

Existing SOTA post-training pruning methods that focus solely on model performance risk sacrificing fairness, as demonstrated by our findings in Section~\ref{sec:pruning_calibration_fairness}. This limitation becomes particularly crucial in opinion summarisation, where models can maintain performance while amplifying biases. Our method addresses this by targeting parameters that exhibit high-gradient and low-activation (HGLA) values, those that are redundant in processing input but sensitive to generation.

Section~\ref{sec:pruning_calibration_fairness} presented aggregated results averaged across calibration sets to identify broad patterns in how pruning methods affect fairness. While this aggregation was necessary to identify general trends, it potentially obscures important nuances about specific calibration-pruning interactions. Table~\ref{tab:hgla_fairness} offers a closer look at different calibration sets coupled with the preferred methods we found earlier---HGLA and Wanda~\cite{sun2023simple}. Specifically, we use single-sided information to isolate weights that are redundant in input processing of a particular side but whose changes would modify the output, allowing us to target parameters based on specific calibration conditions.

Our analysis reveals that HGLA demonstrates superior performance in fairness metrics compared to Wanda across tested models in both political and review summarisation. 
In political summarisation, HGLA shows particularly strong improvements, evidenced by greater improvements in both right-leaning and left-leaning information pruning overall. Specifically, it shows greater improvement in Gemma-2B and TinyLlama, while also enabling Llama3-8B to be pruned up to 40\%.

In review summarisation, while the improvements are more modest, HGLA maintains consistent performance across both positive and negative information pruning. In pruning using positive information, HGLA achieves stable improvements, particularly in Gemma-2B and TinyLlama. Similarly, in pruning using negative information, HGLA demonstrates reliable fairness preservation, with notable improvements in metrics across all three models, suggesting its robust capability in maintaining fairness during pruning.

\subsection{Human Evaluation}
\begin{table}[h]
\centering
\small
\begin{tabular}{lcc}
\hline
Rank & Model & Rating \\
\hline
1 & HGLA & 1450.6 \\
2 & SparseGPT & 1420.3 \\
3 & Wanda & 1404.2 \\
4 & GBLM-Pruner & 1374.9 \\
5 & GBLM-Gradient & 1350.0 \\
\hline
\end{tabular}
\caption{Elo ratings of pruning methods based on pairwise fairness comparisons. HGLA achieves the highest fairness rating, followed by activation-based methods such as SparseGPT and Wanda, and gradient-based GBLM variants in the lowest tier.
}
\label{tab:model-ratings}
\end{table}

We systematically compare output fairness across five pruning methods through human evaluation, using TinyLlama summaries with 0.3 sparsity ratio and negative-only review calibration, as this configuration demonstrated improved fairness. For our initial evaluation, we create 20 direct comparison pairs by randomly selecting outputs from different pruning methods to verify human annotators' reliability. Following \citet{shandilya2020fairness}, annotators identified distinct positive and negative opinions in each input, then assess which summary better preserves the original opinion distribution, achieving substantial inter-annotator agreement with a Fleiss' Kappa coefficient of 0.555~\cite{fleiss1971measuring}. Given the strong agreement, we extend the evaluation to 100 random comparison pairs to ensure a minimum of 30 comparisons per method, based on evidence that rating convergence occurs after approximately 30 comparative assessments~\cite{eloratings}. Detailed information is provided in Appendix~\ref{sec:human_evaluation_detail}.

We analyse the pairwise comparisons using the Elo rating system~\cite{elo1967proposed}, a framework originally designed for chess rankings. The system dynamically adjusts ratings based on comparison outcomes and competitors' relative strength, with victories against higher-rated opponents earning more points. We initialise all methods with a default rating of 1400 and use a K-factor of 16, determining winners through majority voting across the three annotations per comparison.

As shown in Table~\ref{tab:model-ratings}, by aggregating human evaluations of 30 direct comparisons of summaries generated by each pruning method, the Elo rating system shows that HGLA ranks highest among all methods, followed by activation-based methods such as SparseGPT and Wanda, with gradient-based GBLM variants in the lowest tier. The emergence of three distinct performance tiers suggests that the selection of pruning method significantly influences the preservation of fairness characteristics in the resulting models.

% The results can be found in Table~\ref{tab:model-ratings}, we find that HGLA outperformed other pruning methods, where the activation based methods cluster closer and ranked next. The gradient-based methods of the GBLM ranked last. The substantial gap between HGLA and other methods, validated through multiple independent judgments from highly qualified annotators with good agreement, suggests that HGLA consistently demonstrates superior fairness characteristics.

% An interesting observation that the pruning methods formed three distinct performance tiers suggesting that the selection of pruning methodology significantly influences the preservation of fairness characteristics in the resulting models.

\section{Conclusion}
% In this study, we investigate post-training pruning across large language models to assess their impact on model fairness in opinion summarisation. While prior work has shown calibration sets significantly affect model performance, our findings reveal that pruning methods have a better effect on fairness than calibration selection. In contrast to model hallucination, where pruning leads to less hallucination, we observe decreasing fairness when pruning more with SOTA methods. Not all pruning methods equally preserve model performance and fairness, highlighting the risk of overlooking fairness considerations when focusing solely on performance.
% To address these limitations, we introduce High Gradient Low Activation (HGLA) pruning, targeting parameters redundant in input processing but influential in output generation. Our examination shows promise for improving fairness across different models and tasks where existing methods fall short. Human evaluation reveals that outputs generated by HGLA achieve better fairness compared to those produced by existing SOTA post-training pruning methods.
In this study, we examined how post-training pruning affects LLM fairness in opinion summarisation, finding pruning methods impact fairness more than calibration set selection despite prior research emphasising calibration's importance. Unlike hallucination reduction from pruning, fairness decreases with increased pruning using SOTA methods. Not all pruning methods equally preserve model performance and fairness, highlighting the risk of overlooking fairness considerations when focusing solely on performance.
To address these limitations, we introduce High Gradient Low Activation (HGLA) pruning, targeting parameters redundant in input processing but influential in output generation. Our examination shows promise for improving fairness across models and tasks where existing methods fall short. Human evaluation reveals that outputs generated by HGLA achieve better fairness compared to existing pruning methods.
Our work emphasises that efficiency should not compromise fairness, and future research should explore the relationship between efficiency, performance, and fairness.

\section*{Limitations}
We primarily concentrate on open-source language models due to the restricted access to parameters in closed-source LLMs. However, it's worth noting that our approach to assessing fairness in LLM-generated summaries of text with opinions remains relevant for researchers with access to closed-source models. 
Our research deliberately focuses on post-training pruning as it uniquely allows us to investigate how selectively removing specific parameters affects model bias---a question that other compression techniques cannot address in the same way. Unlike distillation or quantisation, which transform representations or reduce precision uniformly, pruning provides a distinctive analytical lens to observe how different regions of the parameter space influence biased behaviour. Our core research question examines whether it is possible to strategically remove parts of a model to reduce bias---a question fundamentally aligned with pruning's selective removal approach.
Additionally, our work focus on unstructured pruning methods only since unstructured pruning offers finer control over weight retention, allowing preservation of crucial features that may be important for fair representation of minority groups or underrepresented data points. Notice that the same evaluation can be applied on semi-structure and structured pruning. 

Our exclusive focus on fairness is not arbitrary, but fundamentally critical in the context of opinion summarisation. As our paper highlights, summaries of opinionated text have profound implications for how audiences interpret and process information, particularly in sensitive domains like political discourse and product reviews. Fairness becomes paramount because biased summaries can disproportionately represent or misrepresent diverse perspectives, potentially shaping public opinion or consumer decisions. Using metrics that don't properly relate to the bias concept being investigated, along with a lack of distinction between conceptual definitions and their practical implementations, present challenges to actionability to discussed bias~\cite{delobelle2024metrics}.

Finally, our study exclusively uses English-based models, tasks, and calibration data. We hypothesise that the efficacy of the LLM pruning techniques we test is largely independent of language. However, we acknowledge the crucial role of linguistic diversity. Consequently, we suggest that in future studies researchers investigate the effectiveness of LLM pruning methods across a wide spectrum of language families, including those with limited resources.

\section*{Ethical Considerations}
This study followed ethical principles and guidelines. The authors of this paper by no means suggest that language models are intentionally biased. We highly encourage readers to investigate and evaluate the findings for themselves. Overall, the goal of our research is to promote awareness of bias in summarising social media text since it is critical to understand what is summarised and whether it represents actual public opinions. Our work contributes to understanding the biases of summarisation models when summarising social media text, which is crucial for ethical use.

Our approach relies on predefined labels in datasets to measure bias. These labels are assigned based on established policies. However, if the labelling policy itself is inaccurate, our procedure might measure bias incorrectly. Therefore, we recommend using our technique only with datasets that have undergone careful review and construction to ensure accurate labelling.

\bibliography{main}

\begin{thebibliography}{56}
\providecommand{\natexlab}[1]{#1}

\bibitem[{Barbieri et~al.(2020)Barbieri, Camacho-Collados, Anke, and Neves}]{barbieri2020tweeteval}
Francesco Barbieri, Jose Camacho-Collados, Luis~Espinosa Anke, and Leonardo Neves. 2020.
\newblock Tweeteval: Unified benchmark and comparative evaluation for tweet classification.
\newblock In \emph{Findings of the Association for Computational Linguistics: EMNLP 2020}, pages 1644--1650.

\bibitem[{Bilal et~al.(2022)Bilal, Wang, Tsakalidis, Nguyen, Procter, and Liakata}]{bilal2022template}
Iman~Munire Bilal, Bo~Wang, Adam Tsakalidis, Dong Nguyen, Rob Procter, and Maria Liakata. 2022.
\newblock Template-based abstractive microblog opinion summarization.
\newblock \emph{Transactions of the Association for Computational Linguistics}, 10:1229--1248.

\bibitem[{Blodgett et~al.(2016)Blodgett, Green, and O{'}Connor}]{blodgett-etal-2016-demographic}
Su~Lin Blodgett, Lisa Green, and Brendan O{'}Connor. 2016.
\newblock \href {https://doi.org/10.18653/v1/D16-1120} {Demographic dialectal variation in social media: A case study of {A}frican-{A}merican {E}nglish}.
\newblock In \emph{Proceedings of the 2016 Conference on Empirical Methods in Natural Language Processing}, pages 1119--1130, Austin, Texas. Association for Computational Linguistics.

\bibitem[{Bra{\v{z}}inskas et~al.(2021)Bra{\v{z}}inskas, Lapata, and Titov}]{bravzinskas2021learning}
Arthur Bra{\v{z}}inskas, Mirella Lapata, and Ivan Titov. 2021.
\newblock Learning opinion summarizers by selecting informative reviews.
\newblock \emph{arXiv preprint arXiv:2109.04325}.

\bibitem[{Bra{\v{z}}inskas et~al.(2022)Bra{\v{z}}inskas, Nallapati, Bansal, and Dreyer}]{bravzinskas2022efficient}
Arthur Bra{\v{z}}inskas, Ramesh Nallapati, Mohit Bansal, and Markus Dreyer. 2022.
\newblock Efficient few-shot fine-tuning for opinion summarization.
\newblock \emph{arXiv preprint arXiv:2205.02170}.

\bibitem[{Brown(2020)}]{brown2020language}
Tom~B Brown. 2020.
\newblock Language models are few-shot learners.
\newblock \emph{arXiv preprint arXiv:2005.14165}.

\bibitem[{Chowdhery et~al.(2023)Chowdhery, Narang, Devlin, Bosma, Mishra, Roberts, Barham, Chung, Sutton, Gehrmann et~al.}]{chowdhery2023palm}
Aakanksha Chowdhery, Sharan Narang, Jacob Devlin, Maarten Bosma, Gaurav Mishra, Adam Roberts, Paul Barham, Hyung~Won Chung, Charles Sutton, Sebastian Gehrmann, et~al. 2023.
\newblock Palm: Scaling language modeling with pathways.
\newblock \emph{Journal of Machine Learning Research}, 24(240):1--113.

\bibitem[{Chrysostomou et~al.(2024)Chrysostomou, Zhao, Williams, and Aletras}]{chrysostomou2024investigating}
George Chrysostomou, Zhixue Zhao, Miles Williams, and Nikolaos Aletras. 2024.
\newblock Investigating hallucinations in pruned large language models for abstractive summarization.
\newblock \emph{Transactions of the Association for Computational Linguistics}, 12:1163--1181.

\bibitem[{Das et~al.(2023)Das, Ma, and Shen}]{das2023beyond}
Rocktim~Jyoti Das, Liqun Ma, and Zhiqiang Shen. 2023.
\newblock Beyond size: How gradients shape pruning decisions in large language models.
\newblock \emph{arXiv preprint arXiv:2311.04902}.

\bibitem[{Dash et~al.(2019)Dash, Shandilya, Biswas, Ghosh, Ghosh, and Chakraborty}]{dash2019summarizing}
Abhisek Dash, Anurag Shandilya, Arindam Biswas, Kripabandhu Ghosh, Saptarshi Ghosh, and Abhijnan Chakraborty. 2019.
\newblock Summarizing user-generated textual content: Motivation and methods for fairness in algorithmic summaries.
\newblock \emph{Proceedings of the ACM on Human-Computer Interaction}, 3(CSCW):1--28.

\bibitem[{Delobelle et~al.(2024)Delobelle, Attanasio, Nozza, Blodgett, Talat et~al.}]{delobelle2024metrics}
Pieter Delobelle, Giuseppe Attanasio, Debora Nozza, Su~Lin Blodgett, Zeerak Talat, et~al. 2024.
\newblock Metrics for what, metrics for whom: assessing actionability of bias evaluation metrics in nlp.
\newblock In \emph{Proceedings of the 2024 Conference on Empirical Methods in Natural Language Processing}. Association for Computational Linguistics.

\bibitem[{Devlin(2018)}]{devlin2018bert}
Jacob Devlin. 2018.
\newblock Bert: Pre-training of deep bidirectional transformers for language understanding.
\newblock \emph{arXiv preprint arXiv:1810.04805}.

\bibitem[{Dubey et~al.(2024)Dubey, Jauhri, Pandey, Kadian, Al-Dahle, Letman, Mathur, Schelten, Yang, Fan et~al.}]{dubey2024llama}
Abhimanyu Dubey, Abhinav Jauhri, Abhinav Pandey, Abhishek Kadian, Ahmad Al-Dahle, Aiesha Letman, Akhil Mathur, Alan Schelten, Amy Yang, Angela Fan, et~al. 2024.
\newblock The llama 3 herd of models.
\newblock \emph{arXiv preprint arXiv:2407.21783}.

\bibitem[{Durmus et~al.(2023)Durmus, Nguyen, Liao, Schiefer, Askell, Bakhtin, Chen, Hatfield-Dodds, Hernandez, Joseph et~al.}]{durmus2023towards}
Esin Durmus, Karina Nguyen, Thomas~I Liao, Nicholas Schiefer, Amanda Askell, Anton Bakhtin, Carol Chen, Zac Hatfield-Dodds, Danny Hernandez, Nicholas Joseph, et~al. 2023.
\newblock Towards measuring the representation of subjective global opinions in language models.
\newblock \emph{arXiv preprint arXiv:2306.16388}.

\bibitem[{Elo(1967)}]{elo1967proposed}
Arpad~E Elo. 1967.
\newblock The proposed uscf rating system, its development, theory, and applications.
\newblock \emph{Chess life}, 22(8):242--247.

\bibitem[{Epstein et~al.(2023)Epstein, Hertzmann, of~Human~Creativity, Akten, Farid, Fjeld, Frank, Groh, Herman, Leach et~al.}]{epstein2023art}
Ziv Epstein, Aaron Hertzmann, Investigators of~Human~Creativity, Memo Akten, Hany Farid, Jessica Fjeld, Morgan~R Frank, Matthew Groh, Laura Herman, Neil Leach, et~al. 2023.
\newblock Art and the science of generative ai.
\newblock \emph{Science}, 380(6650):1110--1111.

\bibitem[{Feng et~al.(2023)Feng, Park, Liu, and Tsvetkov}]{feng-etal-2023-pretraining}
Shangbin Feng, Chan~Young Park, Yuhan Liu, and Yulia Tsvetkov. 2023.
\newblock \href {https://doi.org/10.18653/v1/2023.acl-long.656} {From pretraining data to language models to downstream tasks: Tracking the trails of political biases leading to unfair {NLP} models}.
\newblock In \emph{Proceedings of the 61st Annual Meeting of the Association for Computational Linguistics (Volume 1: Long Papers)}, pages 11737--11762, Toronto, Canada. Association for Computational Linguistics.

\bibitem[{Fleiss(1971)}]{fleiss1971measuring}
Joseph~L Fleiss. 1971.
\newblock Measuring nominal scale agreement among many raters.
\newblock \emph{Psychological bulletin}, 76(5):378.

\bibitem[{Frantar and Alistarh(2023)}]{frantar2023sparsegpt}
Elias Frantar and Dan Alistarh. 2023.
\newblock Sparsegpt: Massive language models can be accurately pruned in one-shot.
\newblock In \emph{International Conference on Machine Learning}, pages 10323--10337. PMLR.

\bibitem[{Gallegos et~al.(2024)Gallegos, Rossi, Barrow, Tanjim, Kim, Dernoncourt, Yu, Zhang, and Ahmed}]{gallegos2024bias}
Isabel~O Gallegos, Ryan~A Rossi, Joe Barrow, Md~Mehrab Tanjim, Sungchul Kim, Franck Dernoncourt, Tong Yu, Ruiyi Zhang, and Nesreen~K Ahmed. 2024.
\newblock Bias and fairness in large language models: A survey.
\newblock \emph{Computational Linguistics}, pages 1--79.

\bibitem[{Han et~al.(2015)Han, Pool, Tran, and Dally}]{han2015learning}
Song Han, Jeff Pool, John Tran, and William Dally. 2015.
\newblock Learning both weights and connections for efficient neural network.
\newblock \emph{Advances in neural information processing systems}, 28.

\bibitem[{Hooker et~al.(2019)Hooker, Courville, Clark, Dauphin, and Frome}]{hooker2019compressed}
Sara Hooker, Aaron Courville, Gregory Clark, Yann Dauphin, and Andrea Frome. 2019.
\newblock What do compressed deep neural networks forget?
\newblock \emph{arXiv preprint arXiv:1911.05248}.

\bibitem[{Hooker et~al.(2020)Hooker, Moorosi, Clark, Bengio, and Denton}]{hooker2020characterising}
Sara Hooker, Nyalleng Moorosi, Gregory Clark, Samy Bengio, and Emily Denton. 2020.
\newblock Characterising bias in compressed models.
\newblock \emph{arXiv preprint arXiv:2010.03058}.

\bibitem[{Hou et~al.(2024)Hou, Li, He, Yan, Chen, and McAuley}]{hou2024bridging}
Yupeng Hou, Jiacheng Li, Zhankui He, An~Yan, Xiusi Chen, and Julian McAuley. 2024.
\newblock Bridging language and items for retrieval and recommendation.
\newblock \emph{arXiv preprint arXiv:2403.03952}.

\bibitem[{Huang et~al.(2024)Huang, Fayek, and Zhang}]{huang-etal-2024-bias}
Nannan Huang, Haytham Fayek, and Xiuzhen Zhang. 2024.
\newblock \href {https://aclanthology.org/2024.eacl-long.63} {Bias in opinion summarisation from pre-training to adaptation: A case study in political bias}.
\newblock In \emph{Proceedings of the 18th Conference of the European Chapter of the Association for Computational Linguistics (Volume 1: Long Papers)}, pages 1041--1055, St. Julian{'}s, Malta. Association for Computational Linguistics.

\bibitem[{Huang et~al.(2023)Huang, Tian, Fayek, and Zhang}]{huang-etal-2023-examining}
Nannan Huang, Lin Tian, Haytham Fayek, and Xiuzhen Zhang. 2023.
\newblock \href {https://aclanthology.org/2023.wassa-1.14} {Examining bias in opinion summarisation through the perspective of opinion diversity}.
\newblock In \emph{Proceedings of the 13th Workshop on Computational Approaches to Subjectivity, Sentiment, {\&} Social Media Analysis}, pages 149--161, Toronto, Canada. Association for Computational Linguistics.

\bibitem[{Jaiswal et~al.(2023)Jaiswal, Gan, Du, Zhang, Wang, and Yang}]{jaiswal2023compressing}
Ajay Jaiswal, Zhe Gan, Xianzhi Du, Bowen Zhang, Zhangyang Wang, and Yinfei Yang. 2023.
\newblock Compressing llms: The truth is rarely pure and never simple.
\newblock \emph{arXiv preprint arXiv:2310.01382}.

\bibitem[{Jakesch et~al.(2023)Jakesch, Bhat, Buschek, Zalmanson, and Naaman}]{jakesch2023co}
Maurice Jakesch, Advait Bhat, Daniel Buschek, Lior Zalmanson, and Mor Naaman. 2023.
\newblock Co-writing with opinionated language models affects users’ views.
\newblock In \emph{Proceedings of the 2023 CHI conference on human factors in computing systems}, pages 1--15.

\bibitem[{Ladhak et~al.(2023)Ladhak, Durmus, Suzgun, Zhang, Jurafsky, Mckeown, and Hashimoto}]{ladhak2023pre}
Faisal Ladhak, Esin Durmus, Mirac Suzgun, Tianyi Zhang, Dan Jurafsky, Kathleen Mckeown, and Tatsunori~B Hashimoto. 2023.
\newblock When do pre-training biases propagate to downstream tasks? a case study in text summarization.
\newblock In \emph{Proceedings of the 17th Conference of the European Chapter of the Association for Computational Linguistics}, pages 3198--3211.

\bibitem[{Le~Scao et~al.(2023)Le~Scao, Fan, Akiki, Pavlick, Ili{\'c}, Hesslow, Castagn{\'e}, Luccioni, Yvon, Gall{\'e} et~al.}]{le2023bloom}
Teven Le~Scao, Angela Fan, Christopher Akiki, Ellie Pavlick, Suzana Ili{\'c}, Daniel Hesslow, Roman Castagn{\'e}, Alexandra~Sasha Luccioni, Fran{\c{c}}ois Yvon, Matthias Gall{\'e}, et~al. 2023.
\newblock Bloom: A 176b-parameter open-access multilingual language model.

\bibitem[{Li et~al.(2023)Li, Du, Song, Wang, and Wang}]{li2023survey}
Yingji Li, Mengnan Du, Rui Song, Xin Wang, and Ying Wang. 2023.
\newblock A survey on fairness in large language models.
\newblock \emph{arXiv preprint arXiv:2308.10149}.

\bibitem[{Liang et~al.(2021)Liang, Wu, Morency, and Salakhutdinov}]{liang2021towards}
Paul~Pu Liang, Chiyu Wu, Louis-Philippe Morency, and Ruslan Salakhutdinov. 2021.
\newblock Towards understanding and mitigating social biases in language models.
\newblock In \emph{International Conference on Machine Learning}, pages 6565--6576. PMLR.

\bibitem[{Lin(2004)}]{lin2004rouge}
Chin-Yew Lin. 2004.
\newblock Rouge: A package for automatic evaluation of summaries.
\newblock In \emph{Text summarization branches out}, pages 74--81.

\bibitem[{Liu et~al.(2019)Liu, Ott, Goyal, Du, Joshi, Chen, Levy, Lewis, Zettlemoyer, and Stoyanov}]{liu2019roberta}
Yinhan Liu, Myle Ott, Naman Goyal, Jingfei Du, Mandar Joshi, Danqi Chen, Omer Levy, Mike Lewis, Luke Zettlemoyer, and Veselin Stoyanov. 2019.
\newblock Roberta: A robustly optimized bert pretraining approach.
\newblock \emph{arXiv preprint arXiv:1907.11692}.

\bibitem[{Menghani(2023)}]{menghani2023efficient}
Gaurav Menghani. 2023.
\newblock Efficient deep learning: A survey on making deep learning models smaller, faster, and better.
\newblock \emph{ACM Computing Surveys}, 55(12):1--37.

\bibitem[{Misra et~al.(2024)Misra, Chaudhary, Goyal, Runwal, and Chen}]{misra2024uncovering}
Diganta Misra, Muawiz Chaudhary, Agam Goyal, Bharat Runwal, and Pin~Yu Chen. 2024.
\newblock Uncovering the hidden cost of model compression.
\newblock In \emph{Proceedings of the IEEE/CVF Conference on Computer Vision and Pattern Recognition}, pages 1611--1621.

\bibitem[{Radford et~al.(2019)Radford, Wu, Child, Luan, Amodei, Sutskever et~al.}]{radford2019language}
Alec Radford, Jeffrey Wu, Rewon Child, David Luan, Dario Amodei, Ilya Sutskever, et~al. 2019.
\newblock Language models are unsupervised multitask learners.
\newblock \emph{OpenAI blog}, 1(8):9.

\bibitem[{Raffel et~al.(2020)Raffel, Shazeer, Roberts, Lee, Narang, Matena, Zhou, Li, and Liu}]{raffel2020exploring}
Colin Raffel, Noam Shazeer, Adam Roberts, Katherine Lee, Sharan Narang, Michael Matena, Yanqi Zhou, Wei Li, and Peter~J Liu. 2020.
\newblock Exploring the limits of transfer learning with a unified text-to-text transformer.
\newblock \emph{Journal of machine learning research}, 21(140):1--67.

\bibitem[{Ratings(2024)}]{eloratings}
World Football~Elo Ratings. 2024.
\newblock \href {https://eloratings.net/about} {About {Elo} ratings}.
\newblock Accessed: 2024-02-10.

\bibitem[{Santurkar et~al.(2023)Santurkar, Durmus, Ladhak, Lee, Liang, and Hashimoto}]{santurkar2023whose}
Shibani Santurkar, Esin Durmus, Faisal Ladhak, Cinoo Lee, Percy Liang, and Tatsunori Hashimoto. 2023.
\newblock Whose opinions do language models reflect?
\newblock \emph{arXiv preprint arXiv:2303.17548}.

\bibitem[{Shandilya et~al.(2020)Shandilya, Dash, Chakraborty, Ghosh, and Ghosh}]{shandilya2020fairness}
Anurag Shandilya, Abhisek Dash, Abhijnan Chakraborty, Kripabandhu Ghosh, and Saptarshi Ghosh. 2020.
\newblock Fairness for whom? understanding the reader’s perception of fairness in text summarization.
\newblock In \emph{2020 IEEE International Conference on Big Data (Big Data)}, pages 3692--3701. IEEE.

\bibitem[{Shandilya et~al.(2018)Shandilya, Ghosh, and Ghosh}]{shandilya2018fairness}
Anurag Shandilya, Kripabandhu Ghosh, and Saptarshi Ghosh. 2018.
\newblock Fairness of extractive text summarization.
\newblock In \emph{Companion Proceedings of the The Web Conference 2018}, pages 97--98.

\bibitem[{Sheng et~al.(2019)Sheng, Chang, Natarajan, and Peng}]{sheng2019woman}
Emily Sheng, Kai-Wei Chang, Prem Natarajan, and Nanyun Peng. 2019.
\newblock The woman worked as a babysitter: On biases in language generation.
\newblock In \emph{Proceedings of the 2019 Conference on Empirical Methods in Natural Language Processing and the 9th International Joint Conference on Natural Language Processing (EMNLP-IJCNLP)}, pages 3407--3412.

\bibitem[{Sun et~al.(2023)Sun, Liu, Bair, and Kolter}]{sun2023simple}
Mingjie Sun, Zhuang Liu, Anna Bair, and J~Zico Kolter. 2023.
\newblock A simple and effective pruning approach for large language models.
\newblock \emph{arXiv preprint arXiv:2306.11695}.

\bibitem[{Team et~al.(2024)Team, Mesnard, Hardin, Dadashi, Bhupatiraju, Pathak, Sifre, Rivi{\`e}re, Kale, Love et~al.}]{team2024gemma}
Gemma Team, Thomas Mesnard, Cassidy Hardin, Robert Dadashi, Surya Bhupatiraju, Shreya Pathak, Laurent Sifre, Morgane Rivi{\`e}re, Mihir~Sanjay Kale, Juliette Love, et~al. 2024.
\newblock Gemma: Open models based on gemini research and technology.
\newblock \emph{arXiv preprint arXiv:2403.08295}.

\bibitem[{Touvron et~al.(2023)Touvron, Lavril, Izacard, Martinet, Lachaux, Lacroix, Rozi{\`e}re, Goyal, Hambro, Azhar et~al.}]{touvron2023llama}
Hugo Touvron, Thibaut Lavril, Gautier Izacard, Xavier Martinet, Marie-Anne Lachaux, Timoth{\'e}e Lacroix, Baptiste Rozi{\`e}re, Naman Goyal, Eric Hambro, Faisal Azhar, et~al. 2023.
\newblock Llama: Open and efficient foundation language models.
\newblock \emph{arXiv preprint arXiv:2302.13971}.

\bibitem[{Vig et~al.(2020)Vig, Gehrmann, Belinkov, Qian, Nevo, Sakenis, Huang, Singer, and Shieber}]{vig2020causal}
Jesse Vig, Sebastian Gehrmann, Yonatan Belinkov, Sharon Qian, Daniel Nevo, Simas Sakenis, Jason Huang, Yaron Singer, and Stuart Shieber. 2020.
\newblock Causal mediation analysis for interpreting neural nlp: The case of gender bias.
\newblock \emph{arXiv preprint arXiv:2004.12265}.

\bibitem[{Williams and Aletras(2024)}]{williams2024impact}
Miles Williams and Nikolaos Aletras. 2024.
\newblock On the impact of calibration data in post-training quantization and pruning.
\newblock In \emph{Proceedings of the 62nd Annual Meeting of the Association for Computational Linguistics (Volume 1: Long Papers)}, pages 10100--10118.

\bibitem[{Wolf et~al.(2020)Wolf, Debut, Sanh, Chaumond, Delangue, Moi, Cistac, Rault, Louf, Funtowicz, Davison, Shleifer, von Platen, Ma, Jernite, Plu, Xu, Le~Scao, Gugger, Drame, Lhoest, and Rush}]{wolf-etal-2020-transformers}
Thomas Wolf, Lysandre Debut, Victor Sanh, Julien Chaumond, Clement Delangue, Anthony Moi, Pierric Cistac, Tim Rault, Remi Louf, Morgan Funtowicz, Joe Davison, Sam Shleifer, Patrick von Platen, Clara Ma, Yacine Jernite, Julien Plu, Canwen Xu, Teven Le~Scao, Sylvain Gugger, Mariama Drame, Quentin Lhoest, and Alexander Rush. 2020.
\newblock \href {https://doi.org/10.18653/v1/2020.emnlp-demos.6} {Transformers: State-of-the-art natural language processing}.
\newblock In \emph{Proceedings of the 2020 Conference on Empirical Methods in Natural Language Processing: System Demonstrations}, pages 38--45, Online. Association for Computational Linguistics.

\bibitem[{Wu et~al.(2022)Wu, Raghavendra, Gupta, Acun, Ardalani, Maeng, Chang, Aga, Huang, Bai et~al.}]{wu2022sustainable}
Carole-Jean Wu, Ramya Raghavendra, Udit Gupta, Bilge Acun, Newsha Ardalani, Kiwan Maeng, Gloria Chang, Fiona Aga, Jinshi Huang, Charles Bai, et~al. 2022.
\newblock Sustainable ai: Environmental implications, challenges and opportunities.
\newblock \emph{Proceedings of Machine Learning and Systems}, 4:795--813.

\bibitem[{Yuan et~al.(2021)Yuan, Neubig, and Liu}]{yuan2021bartscore}
Weizhe Yuan, Graham Neubig, and Pengfei Liu. 2021.
\newblock Bartscore: Evaluating generated text as text generation.
\newblock \emph{Advances in Neural Information Processing Systems}, 34:27263--27277.

\bibitem[{Zayed et~al.(2024)Zayed, Mordido, Shabanian, Baldini, and Chandar}]{zayed2024fairness}
Abdelrahman Zayed, Gon{\c{c}}alo Mordido, Samira Shabanian, Ioana Baldini, and Sarath Chandar. 2024.
\newblock Fairness-aware structured pruning in transformers.
\newblock In \emph{Proceedings of the AAAI Conference on Artificial Intelligence}, volume~38, pages 22484--22492.

\bibitem[{Zhang et~al.(2024)Zhang, Zeng, Wang, and Lu}]{zhang2024tinyllama}
Peiyuan Zhang, Guangtao Zeng, Tianduo Wang, and Wei Lu. 2024.
\newblock Tinyllama: An open-source small language model.
\newblock \emph{arXiv preprint arXiv:2401.02385}.

\bibitem[{Zhang et~al.(2019)Zhang, Kishore, Wu, Weinberger, and Artzi}]{zhang2019bertscore}
Tianyi Zhang, Varsha Kishore, Felix Wu, Kilian~Q Weinberger, and Yoav Artzi. 2019.
\newblock Bertscore: Evaluating text generation with bert.
\newblock \emph{arXiv preprint arXiv:1904.09675}.

\bibitem[{Zhang et~al.(2023)Zhang, Zhang, Liu, Fabbri, Liu, Kamoi, Lu, Xiong, Zhao, Radev et~al.}]{zhang2023fair}
Yusen Zhang, Nan Zhang, Yixin Liu, Alexander Fabbri, Junru Liu, Ryo Kamoi, Xiaoxin Lu, Caiming Xiong, Jieyu Zhao, Dragomir Radev, et~al. 2023.
\newblock Fair abstractive summarization of diverse perspectives.
\newblock \emph{arXiv preprint arXiv:2311.07884}.

\bibitem[{Zhu et~al.(2023)Zhu, Li, Liu, Ma, and Wang}]{zhu2023survey}
Xunyu Zhu, Jian Li, Yong Liu, Can Ma, and Weiping Wang. 2023.
\newblock A survey on model compression for large language models.
\newblock \emph{arXiv preprint arXiv:2308.07633}.

\end{thebibliography}

\appendix

\section{Appendix}
\label{sec:appendix}
\subsection{Steps and Classification Models for Fairness Evaluation}
\label{sec:fairness_evaluation_classification}
To calculate SPD, we adopt sentence splitting function\footnote{\url{https://www.nltk.org/api/nltk.tokenize.html}} to split summaries into sentences and then classify sentences from the generated summaries using the same classification model for political tweet classification provided by \citet{huang-etal-2024-bias} that is a RoBERTa \cite{liu2019roberta} further pretrained on the tweet dataset \cite{barbieri2020tweeteval}\footnote{\url{https://huggingface.co/cardiffnlp/twitter-roberta-base}.} and then fine-tuned using the political partition of the dataset provided by \citet{dash2019summarizing}. The average accuracy and macro F1 scores of the model are 0.9162 and 0.9031 respectively. 
For review sentiment classification, we use a BERT \cite{devlin2018bert} base model finetuned for sentiment analysis on product reviews in six languages\footnote{\url{https://huggingface.co/nlptown/bert-base-multilingual-uncased-sentiment}} and then further fine-tuned on an Amazon US Customer Reviews Dataset\footnote{\url{https://huggingface.co/LiYuan/amazon-review-sentiment-analysis}}. We formulate our question as a binary classification where we convert ratings 1 and 2 as negative and 3 to 5 as positive and test it using the provided input dataset by FewSum \cite{bravzinskas2022efficient} and AmaSum \cite{bravzinskas2021learning}. The average accuracy and macro F1 scores of the model are 0.9297 and 0.887 respectively.

\subsection{Fairness Evaluation Test Set Generation}
\label{sec:fairness_evaluation_test_set}
For model fairness evaluation, we manually generate test sets with 100 input collections.
For political tweet summarisation, we use the political partition in FairSumm for evaluating fairness in summarising political tweets. Similar to the input collections used by \citet{bilal2022template}, where each input collection contains roughly 30 tweets, when generating the test set for fairness evaluation, we select 30 tweets for each example in the test set.
For review summarisation fairness, we use the Amazon reviews 2023. We selected reviews with a minimum of 30 words and a maximum of 120 words. This matches the review length in FewSum \cite{bravzinskas2022efficient} and AmaSum \cite{bravzinskas2021learning}, where each input collection contains 8 reviews and the average length of each review is between 30 and 120 words. 
For the different calibration sets that we are generating in Section~\ref{sec:pruning_calibration_fairness}, we further filter products that have at least 8 reviews representing each side, ensuring we have products that satisfy these requirements to select from when creating these sets.
The test sets include equal representations of inputs only (i.e., 50\% positive and 50\% negative reviews). 

\subsection{Prompt Template}
\label{sec:promt}
For all these prompt-based LLMs, we follow \citet{zhang2023fair} to generate summaries and use the template provided in their work for both political tweet summarisation and review summarisation ``Reviews about \{TOPIC\}. Each review is separated by || : \{SOURCE\} Please write a short text containing the salient information, i.e. a summary. The summary of the reviews is:".

\subsection{Random Calibration Set Construction}
\label{sec:random_set_construct}
For political tweet summarisation, we generate the calibration set by random sampling from FairSumm \cite{dash2019summarizing} with the length of the average length in the political partition of MOS \cite{bilal2022template}. 
The review calibration set is randomly sampled using the Amazon 2023 dataset \cite{hou2024bridging}.
The filtering process is the same as mentioned in Appendix~\ref{sec:fairness_evaluation_test_set}, where we filtered reviews to have a minimum of 30 words and a maximum of 120 words, and ensured that the collection of reviews for each product has over 8 reviews for each side.
Adopting from existing post-training pruning methods we generate 128 calibration examples in our calibration set \cite{sun2023simple}.

\subsection{Model Performance Using Random Calibration Set}
\label{sec:model_performance_ramdom_calibration}
\begin{figure*}[htbp]
    \centering
    \begin{subfigure}[t]{0.45\textwidth}
        \centering
        \includegraphics[width=0.95\linewidth]{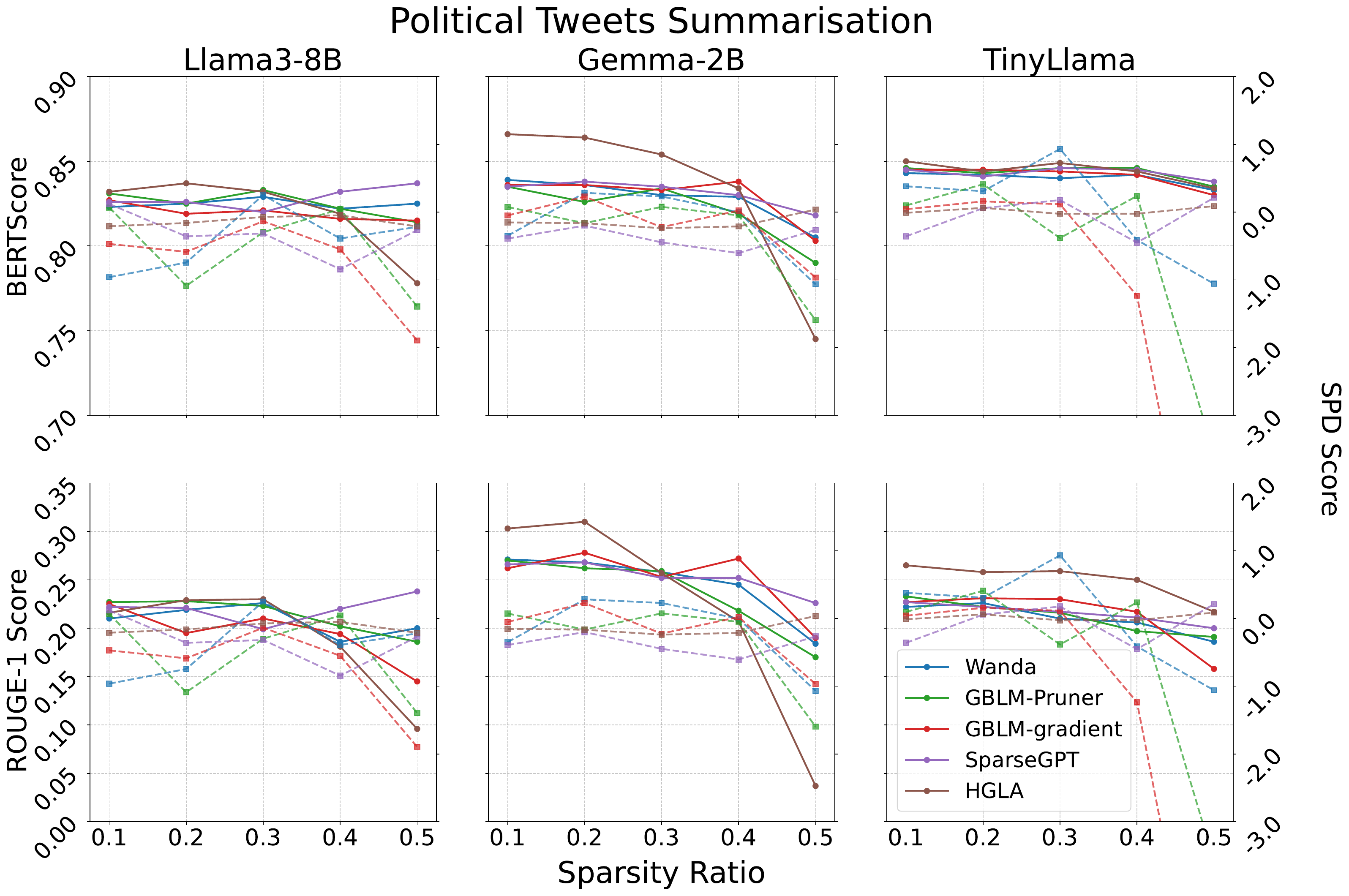}
        \caption{Political tweet summarisation - SPD}
        \label{fig:political_performance_spd_tradeoff}
    \end{subfigure}
    \hfill
    \begin{subfigure}[t]{0.45\textwidth}
        \centering
        \includegraphics[width=0.95\linewidth]{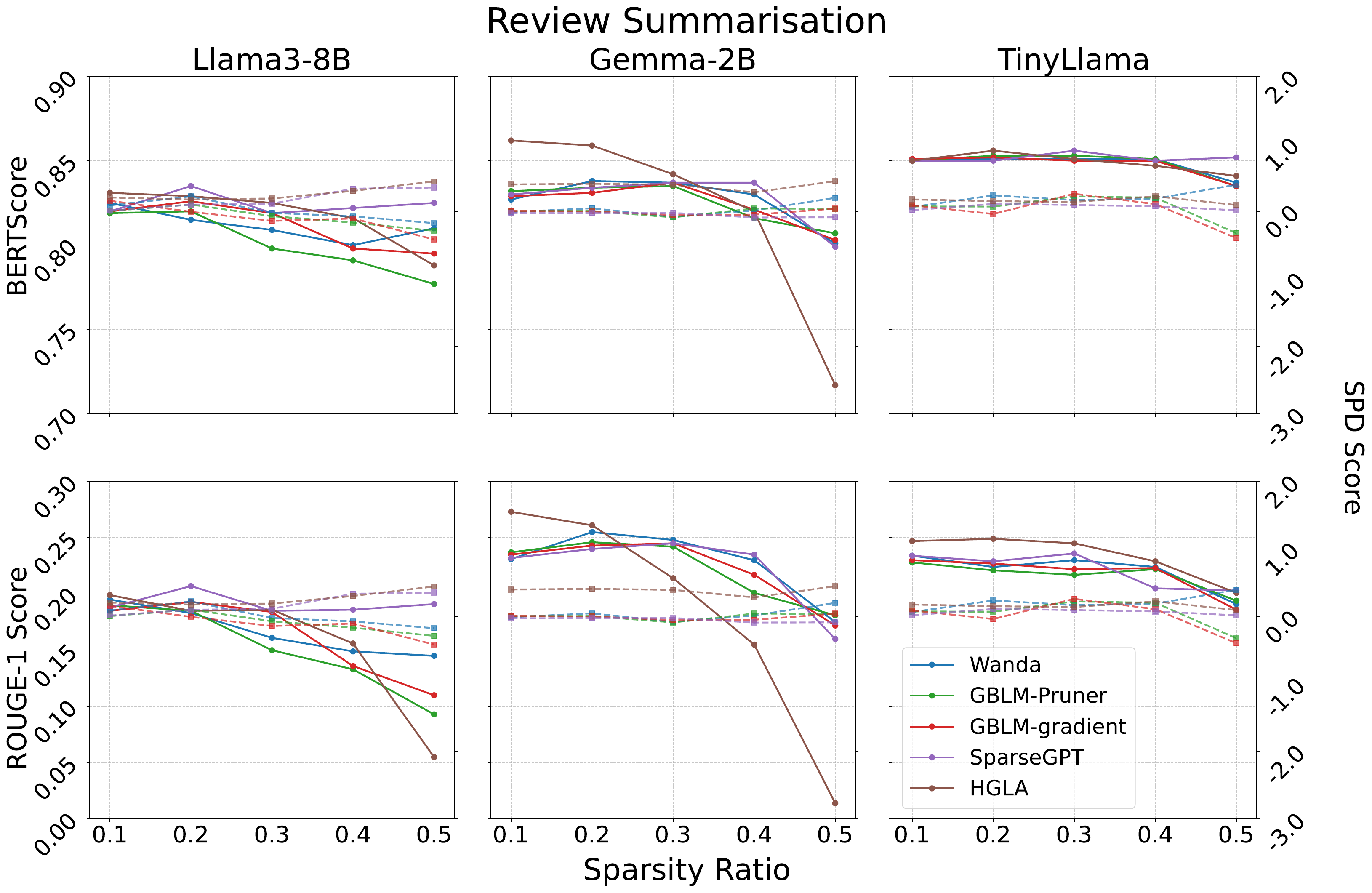}
        \caption{Review summarisation - SPD}
        \label{fig:review_performance_spd_tradeoff}
    \end{subfigure}
    
    \centering
    
    \begin{subfigure}[t]{0.45\textwidth}
        \centering
        \includegraphics[width=0.95\linewidth]{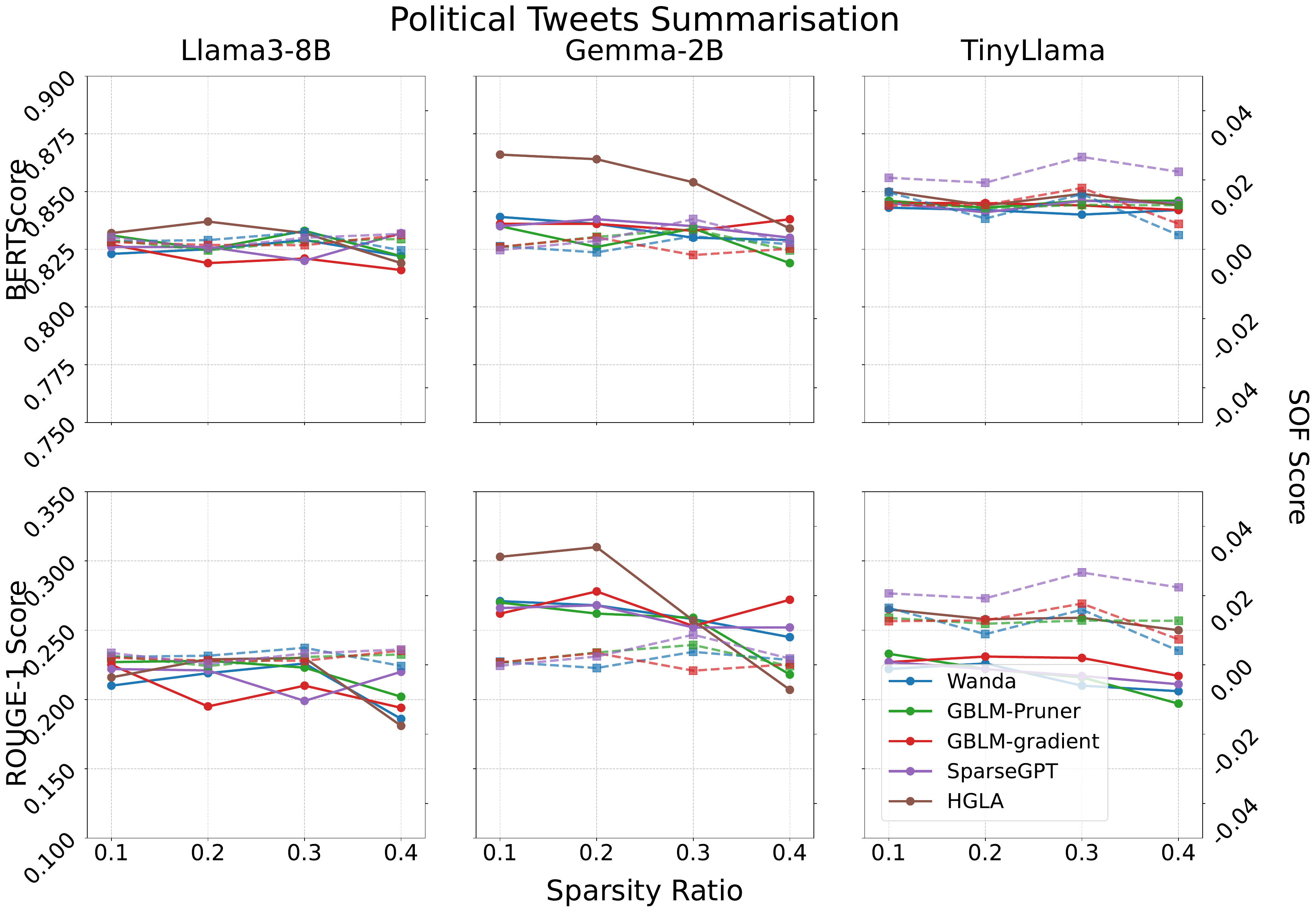}
        \caption{Political tweet summarisation - SOF}
        \label{fig:political_performance_sof_tradeoff}
    \end{subfigure}
    \hfill
    \begin{subfigure}[t]{0.45\textwidth}
        \centering
        \includegraphics[width=0.95\linewidth]{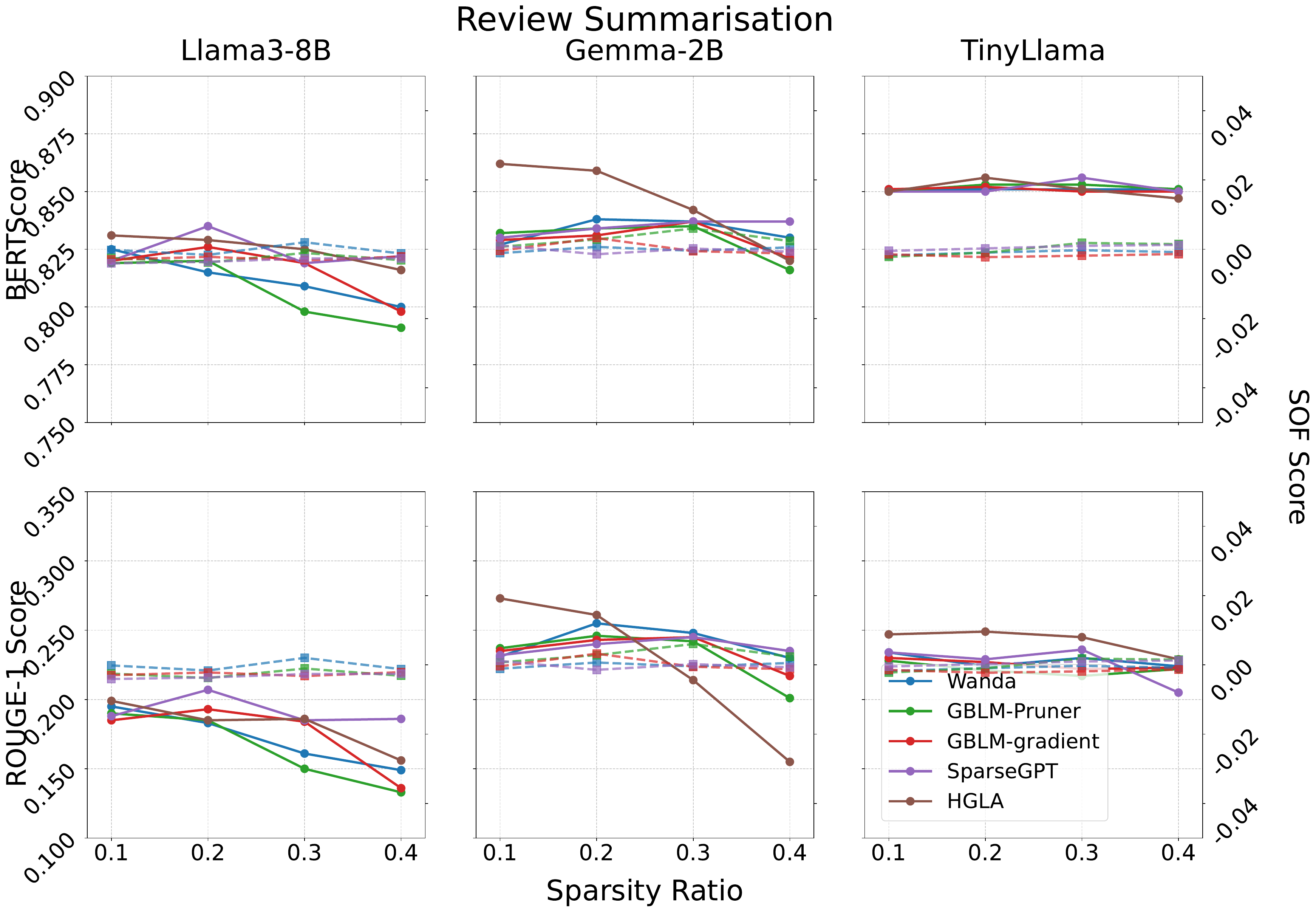}
        \caption{Review summarisation - SOF}
        \label{fig:review_performance_sof_tradeoff}
    \end{subfigure}
    
    \begin{subfigure}[t]{0.45\textwidth}
        \centering
        \includegraphics[width=0.95\linewidth]{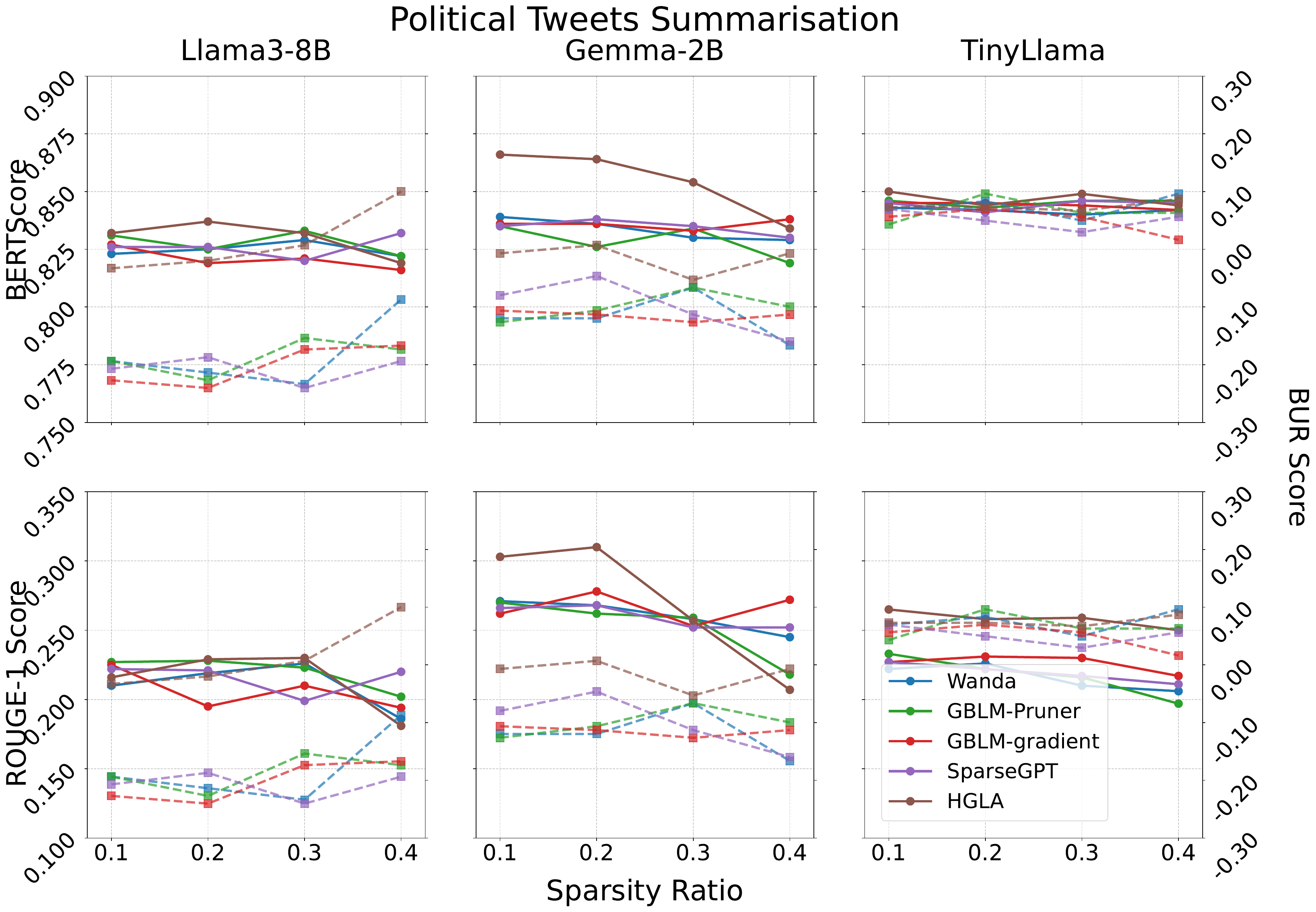}
        \caption{Political tweet summarisation - BUR}
        \label{fig:political_performance_bur_tradeoff}
    \end{subfigure}
    \hfill
    \begin{subfigure}[t]{0.45\textwidth}
        \centering
        \includegraphics[width=0.95\linewidth]{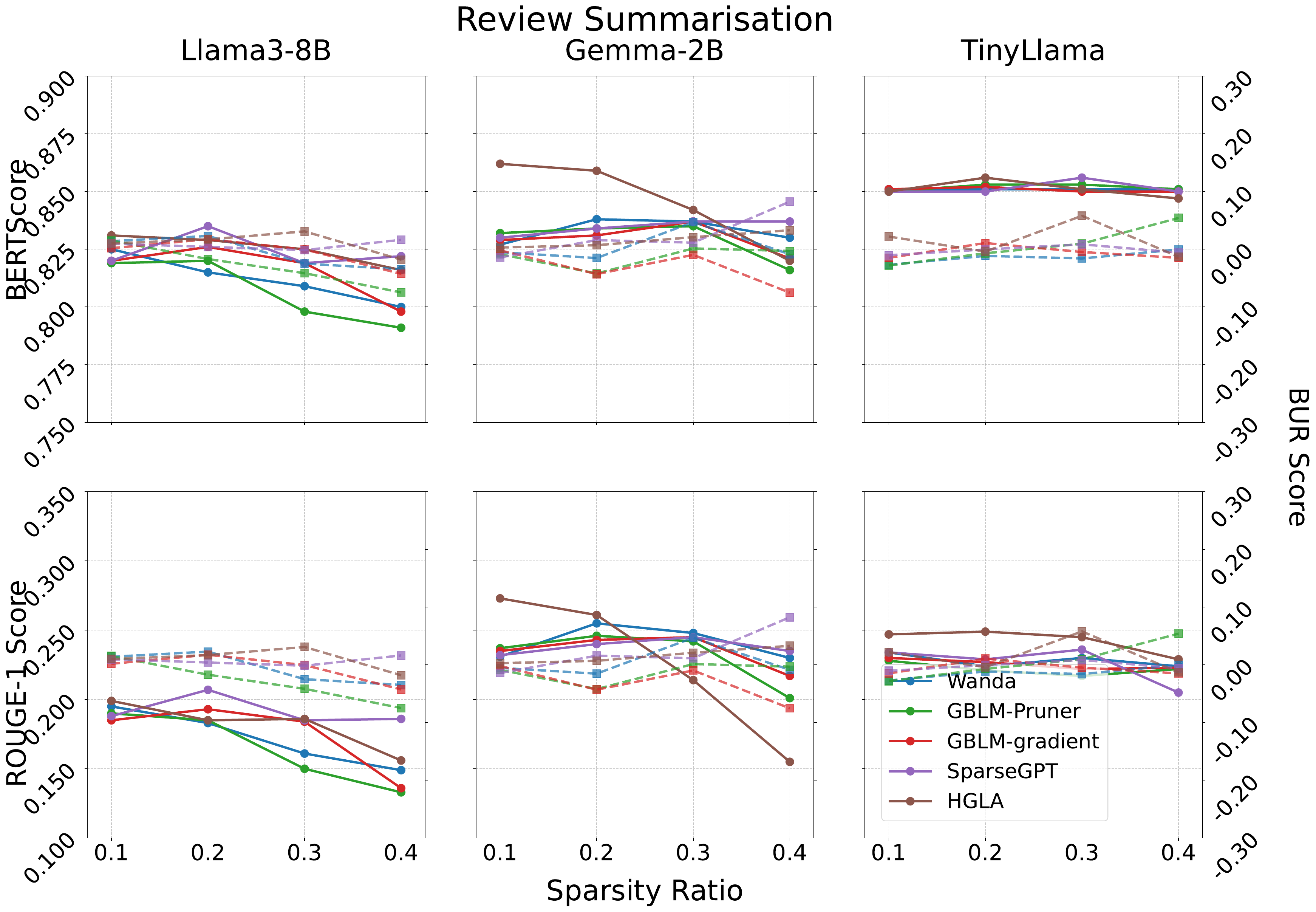}
        \caption{Review summarisation - BUR}
        \label{fig:review_performance_bur_tradeoff}
    \end{subfigure}
    
    \begin{subfigure}[t]{0.45\textwidth}
        \centering
        \includegraphics[width=0.95\linewidth]{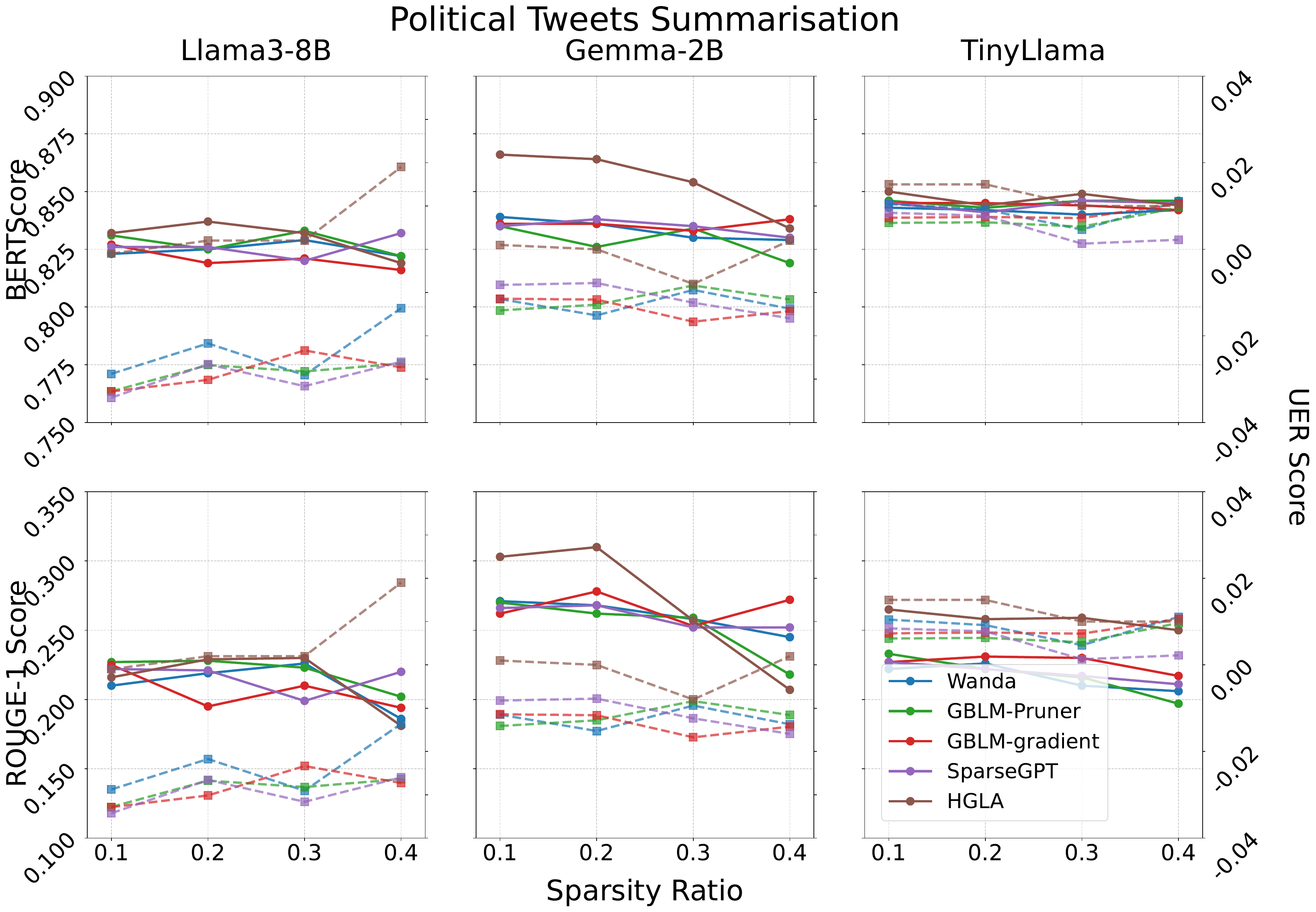}
        \caption{Political tweet summarisation - UER}
        \label{fig:political_performance_uer_tradeoff}
    \end{subfigure}
    \hfill
    \begin{subfigure}[t]{0.45\textwidth}
        \centering
        \includegraphics[width=0.95\linewidth]{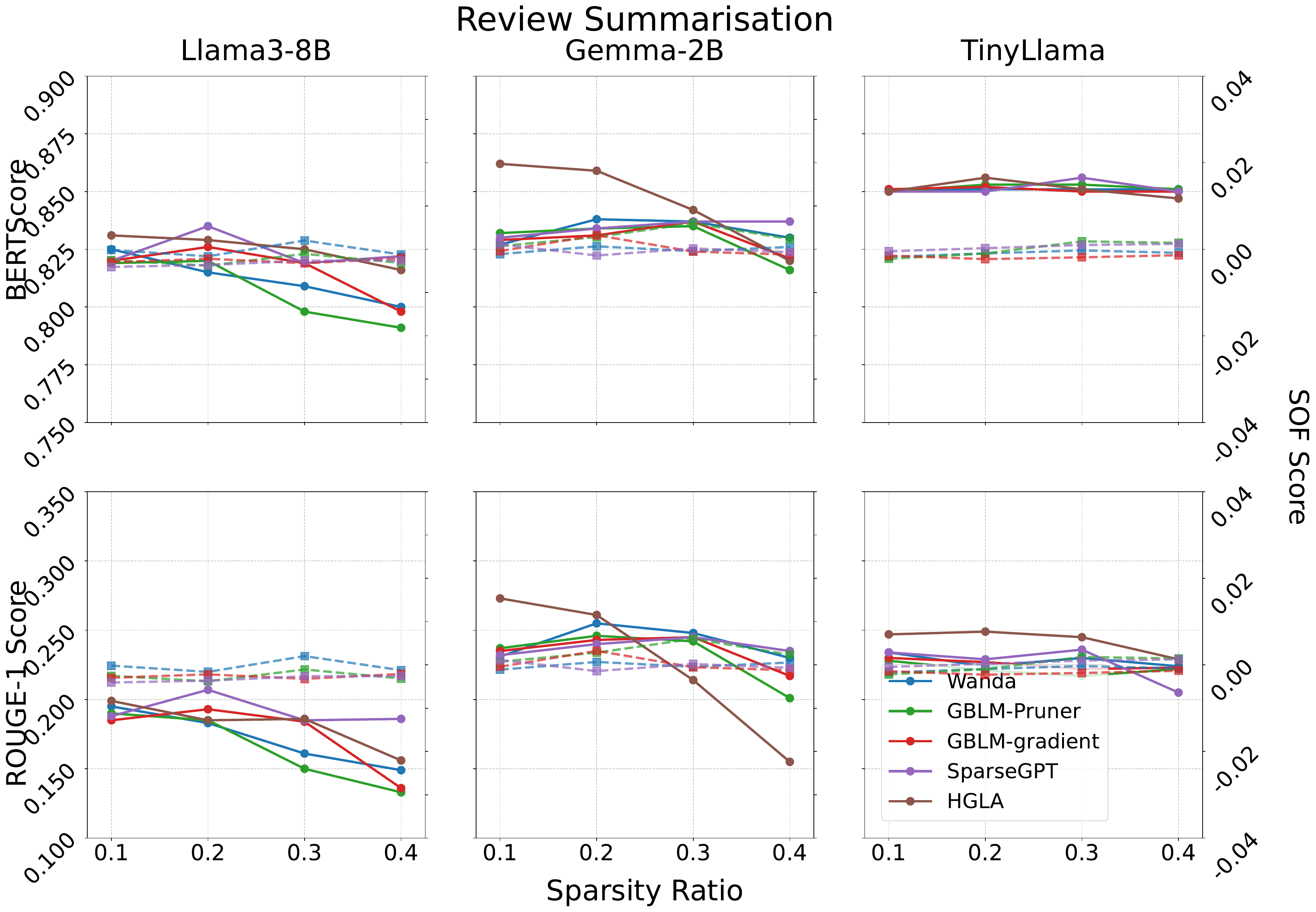}
        \caption{Review summarisation - UER}
        \label{fig:review_performance_uer_tradeoff}
    \end{subfigure}
    \caption{Performance-fairness tradeoff: solid lines represent model performance and dotted lines represent model fairness, evaluated across four fairness metrics (SPD, SOF, BUR, UER) on political tweet and review summarisation. The x-axis shows the sparsity ratio, with higher values indicating pruning more model parameters.}
    \label{fig:performance_fairness_tradeoff}
\end{figure*}

% Model performance across different pruning methods using random calibration sets are reported in Table~\ref{tab:performance_random} with 
Performance-fairness tradeoff visualisations are shown in Figure~\ref{fig:performance_fairness_tradeoff}. The relationship between model pruning and fairness shows general trends toward decreased fairness with increased pruning ratios across summarisation tasks and metrics. However, this pattern varies across tasks, methods, and evaluation metrics, with Wanda demonstrating more stable performance compared to other pruning methods. The relationship between model performance and fairness shows complex interactions - while pruning leads to gradual performance decline across models, fairness metric changes vary by task and model. Wanda exhibits the most balanced trade-off, maintaining both performance and fairness stability, while GBLM-gradient shows more dramatic fairness changes even with stable performance. These factors interact in complex and sometimes counterintuitive ways, with certain combinations of pruning methods and tasks showing unexpected preservation or degradation of fairness metrics. The results indicate that pruning without awareness of fairness implications could potentially harm model fairness, highlighting the importance of careful pruning method selection and continuous monitoring of fairness metrics during the pruning process.

\subsection{Model Performance Across Pruning Methods and Calibration Sets}
\label{sec:model_performance_various_prune_calibration}
Model performance across different pruning methods using different calibration sets are reported in Table~\ref{tab:performance_political} and Table~\ref{tab:performance_review} for summarising political tweets and reviews respectively.

\begin{sidewaystable}
\tiny
\setlength{\tabcolsep}{2pt}
\renewcommand{\arraystretch}{0.9}
\centering
\resizebox{\textheight}{!}{%
\begin{tabular}{l*{29}{r}}
\toprule
& \multicolumn{9}{c}{Llama3-8B} & \multicolumn{9}{c}{Gemma-2B} & \multicolumn{9}{c}{TinyLlama} \\
\cmidrule(lr){3-10} \cmidrule(lr){11-18} \cmidrule(lr){19-26}
&  & \multicolumn{2}{c}{SparseGPT} & \multicolumn{2}{c}{Wanda} & \multicolumn{2}{c}{GBLM-Pruner} & \multicolumn{2}{c}{GBLM-Gradient} 
& \multicolumn{2}{c}{SparseGPT} & \multicolumn{2}{c}{Wanda} & \multicolumn{2}{c}{GBLM-Pruner} & \multicolumn{2}{c}{GBLM-Gradient}
& \multicolumn{2}{c}{SparseGPT} & \multicolumn{2}{c}{Wanda} & \multicolumn{2}{c}{GBLM-Pruner} & \multicolumn{2}{c}{GBLM-Gradient} \\
Sparsity Ratio & Calibration Set & ROUGE1/2/L & BERTScore & ROUGE1/2/L & BERTScore & ROUGE1/2/L & BERTScore & ROUGE1/2/L & BERTScore
& ROUGE1/2/L & BERTScore & ROUGE1/2/L & BERTScore & ROUGE1/2/L & BERTScore & ROUGE1/2/L & BERTScore & ROUGE1/2/L & BERTScore
& ROUGE1/2/L & BERTScore & ROUGE1/2/L & BERTScore & ROUGE1/2/L & BERTScore && \\
\midrule
\multirow{7}{*}{10\%} & Left only & 0.224/0.068/0.155 & 0.829 & 0.236/0.073/0.161 & 0.830 & 0.220/0.066/0.154 & 0.824 & 0.222/0.069/0.154 & 0.822 & 0.265/0.079/0.193 & 0.837 & 0.264/0.085/0.196 & 0.838 & 0.269/0.088/0.201 & 0.838 & 0.262/0.087/0.195 & 0.834 & 0.233/0.075/0.158 & 0.846 & 0.223/0.072/0.155 & 0.844 & 0.229/0.077/0.164 & 0.845 & 0.225/0.074/0.155 & 0.844 \\
 & Right only & 0.219/0.068/0.150 & 0.825 & 0.212/0.065/0.150 & 0.825 & 0.226/0.071/0.162 & 0.831 & 0.231/0.068/0.162 & 0.834 & 0.262/0.081/0.189 & 0.833 & 0.264/0.082/0.194 & 0.837 & 0.271/0.090/0.201 & 0.836 & 0.272/0.094/0.206 & 0.840 & 0.239/0.079/0.163 & 0.845 & 0.220/0.071/0.159 & 0.843 & 0.224/0.075/0.159 & 0.845 & 0.227/0.073/0.158 & 0.845 \\
 & Fair input & 0.218/0.067/0.147 & 0.826 & 0.214/0.068/0.152 & 0.823 & 0.234/0.071/0.165 & 0.832 & 0.225/0.073/0.162 & 0.827 & 0.274/0.090/0.198 & 0.836 & 0.266/0.083/0.197 & 0.839 & 0.259/0.082/0.187 & 0.833 & 0.260/0.084/0.188 & 0.833 & 0.229/0.076/0.159 & 0.845 & 0.223/0.073/0.160 & 0.844 & 0.220/0.071/0.151 & 0.847 & 0.212/0.070/0.149 & 0.845 \\
 & Mixed input & 0.209/0.065/0.146 & 0.823 & 0.223/0.071/0.162 & 0.836 & 0.223/0.069/0.152 & 0.825 & 0.199/0.060/0.145 & 0.821 & 0.266/0.084/0.194 & 0.834 & 0.262/0.081/0.194 & 0.835 & 0.261/0.088/0.193 & 0.836 & 0.261/0.086/0.192 & 0.835 & 0.228/0.075/0.157 & 0.844 & 0.222/0.072/0.157 & 0.845 & 0.216/0.066/0.153 & 0.842 & 0.211/0.068/0.149 & 0.843 \\
 & Biased output & 0.221/0.071/0.152 & 0.828 & 0.222/0.072/0.160 & 0.823 & 0.219/0.065/0.157 & 0.830 & 0.223/0.066/0.152 & 0.826 & 0.264/0.084/0.192 & 0.832 & 0.269/0.087/0.199 & 0.839 & 0.270/0.087/0.198 & 0.837 & 0.271/0.093/0.201 & 0.836 & 0.229/0.074/0.156 & 0.845 & 0.221/0.071/0.157 & 0.844 & 0.225/0.076/0.156 & 0.847 & 0.221/0.070/0.152 & 0.843 \\
 & Fair output & 0.217/0.068/0.149 & 0.827 & 0.209/0.063/0.152 & 0.826 & 0.213/0.063/0.150 & 0.825 & 0.232/0.072/0.162 & 0.832 & 0.271/0.088/0.197 & 0.834 & 0.270/0.086/0.200 & 0.838 & 0.273/0.089/0.200 & 0.839 & 0.256/0.082/0.188 & 0.835 & 0.234/0.075/0.158 & 0.845 & 0.219/0.071/0.157 & 0.843 & 0.228/0.076/0.159 & 0.843 & 0.227/0.076/0.160 & 0.845 \\
 & Mixed output & 0.216/0.070/0.150 & 0.826 & 0.233/0.073/0.166 & 0.829 & 0.234/0.071/0.156 & 0.828 & 0.239/0.074/0.163 & 0.832 & 0.266/0.085/0.193 & 0.834 & 0.270/0.083/0.198 & 0.838 & 0.267/0.087/0.194 & 0.833 & 0.270/0.088/0.201 & 0.837 & 0.232/0.078/0.162 & 0.845 & 0.222/0.072/0.159 & 0.844 & 0.220/0.071/0.154 & 0.843 & 0.220/0.074/0.153 & 0.844 \\
  \midrule
 \multirow{7}{*}{20\%} 
 & Left only & 0.211/0.065/0.150 & 0.819 & 0.213/0.061/0.151 & 0.828 & 0.224/0.067/0.157 & 0.829 & 0.230/0.078/0.171 & 0.831 & 0.268/0.086/0.188 & 0.838 & 0.264/0.085/0.191 & 0.837 & 0.274/0.088/0.193 & 0.836 & 0.269/0.083/0.190 & 0.836 & 0.224/0.069/0.157 & 0.843 & 0.225/0.072/0.165 & 0.843 & 0.231/0.075/0.160 & 0.842 & 0.227/0.073/0.159 & 0.845 \\
 & Right only & 0.215/0.070/0.150 & 0.825 & 0.219/0.063/0.163 & 0.829 & 0.209/0.068/0.150 & 0.824 & 0.202/0.062/0.147 & 0.819 & 0.260/0.078/0.184 & 0.836 & 0.256/0.083/0.185 & 0.831 & 0.276/0.087/0.195 & 0.838 & 0.278/0.096/0.204 & 0.837 & 0.225/0.074/0.163 & 0.841 & 0.222/0.069/0.161 & 0.841 & 0.227/0.070/0.155 & 0.844 & 0.217/0.071/0.157 & 0.843 \\
 & Fair input & 0.230/0.072/0.168 & 0.830 & 0.197/0.062/0.145 & 0.824 & 0.205/0.067/0.144 & 0.821 & 0.213/0.063/0.144 & 0.822 & 0.265/0.083/0.188 & 0.835 & 0.265/0.086/0.192 & 0.840 & 0.251/0.080/0.181 & 0.829 & 0.271/0.094/0.195 & 0.840 & 0.222/0.068/0.158 & 0.842 & 0.222/0.072/0.162 & 0.842 & 0.234/0.075/0.171 & 0.848 & 0.204/0.066/0.154 & 0.843 \\
 & Mixed input & 0.218/0.071/0.155 & 0.820 & 0.222/0.071/0.157 & 0.827 & 0.218/0.068/0.151 & 0.827 & 0.213/0.068/0.151 & 0.819 & 0.266/0.088/0.189 & 0.836 & 0.261/0.083/0.187 & 0.832 & 0.246/0.079/0.181 & 0.832 & 0.251/0.083/0.184 & 0.830 & 0.219/0.066/0.153 & 0.842 & 0.230/0.075/0.165 & 0.843 & 0.230/0.069/0.162 & 0.846 & 0.221/0.068/0.158 & 0.845 \\
 & Biased output & 0.214/0.068/0.153 & 0.822 & 0.230/0.075/0.160 & 0.831 & 0.213/0.067/0.156 & 0.828 & 0.232/0.074/0.165 & 0.829 & 0.264/0.084/0.186 & 0.836 & 0.269/0.090/0.196 & 0.836 & 0.263/0.083/0.186 & 0.828 & 0.257/0.090/0.190 & 0.833 & 0.225/0.072/0.156 & 0.844 & 0.223/0.071/0.159 & 0.844 & 0.231/0.074/0.168 & 0.842 & 0.231/0.071/0.162 & 0.844 \\
 & Fair output & 0.209/0.063/0.148 & 0.822 & 0.234/0.069/0.162 & 0.833 & 0.227/0.069/0.165 & 0.829 & 0.224/0.076/0.161 & 0.827 & 0.272/0.086/0.192 & 0.838 & 0.264/0.087/0.191 & 0.837 & 0.268/0.081/0.189 & 0.835 & 0.271/0.091/0.198 & 0.838 & 0.217/0.069/0.154 & 0.841 & 0.231/0.075/0.166 & 0.844 & 0.231/0.074/0.161 & 0.841 & 0.226/0.071/0.159 & 0.843 \\
 & Mixed output & 0.218/0.069/0.156 & 0.827 & 0.214/0.065/0.154 & 0.826 & 0.246/0.076/0.162 & 0.833 & 0.215/0.059/0.145 & 0.824 & 0.268/0.085/0.191 & 0.836 & 0.269/0.087/0.193 & 0.839 & 0.231/0.068/0.171 & 0.823 & 0.251/0.084/0.180 & 0.827 & 0.226/0.073/0.164 & 0.842 & 0.225/0.072/0.162 & 0.842 & 0.229/0.078/0.165 & 0.846 & 0.228/0.074/0.169 & 0.845 \\
  \midrule
  \multirow{7}{*}{30\%} 
 & Left only & 0.208/0.066/0.156 & 0.824 & 0.215/0.061/0.149 & 0.828 & 0.204/0.065/0.143 & 0.818 & 0.202/0.062/0.148 & 0.821 & 0.266/0.082/0.191 & 0.834 & 0.259/0.085/0.191 & 0.835 & 0.265/0.091/0.192 & 0.837 & 0.266/0.086/0.191 & 0.835 & 0.231/0.077/0.167 & 0.844 & 0.223/0.073/0.165 & 0.843 & 0.215/0.079/0.154 & 0.845 & 0.214/0.077/0.163 & 0.846 \\
 & Right only & 0.214/0.067/0.158 & 0.830 & 0.226/0.071/0.160 & 0.827 & 0.210/0.066/0.155 & 0.817 & 0.211/0.067/0.159 & 0.823 & 0.253/0.082/0.183 & 0.829 & 0.256/0.082/0.186 & 0.830 & 0.243/0.078/0.186 & 0.830 & 0.258/0.078/0.190 & 0.839 & 0.229/0.083/0.166 & 0.847 & 0.230/0.078/0.169 & 0.845 & 0.215/0.070/0.156 & 0.847 & 0.225/0.076/0.161 & 0.848 \\
 & Fair input & 0.217/0.065/0.154 & 0.824 & 0.209/0.064/0.153 & 0.822 & 0.204/0.068/0.147 & 0.822 & 0.168/0.053/0.130 & 0.813 & 0.246/0.079/0.180 & 0.829 & 0.262/0.085/0.194 & 0.835 & 0.232/0.079/0.177 & 0.821 & 0.262/0.089/0.196 & 0.835 & 0.233/0.074/0.164 & 0.843 & 0.210/0.071/0.159 & 0.842 & 0.211/0.070/0.164 & 0.844 & 0.206/0.066/0.166 & 0.840 \\
 & Mixed input & 0.198/0.063/0.149 & 0.819 & 0.210/0.057/0.147 & 0.820 & 0.171/0.054/0.128 & 0.808 & 0.183/0.051/0.132 & 0.813 & 0.254/0.080/0.186 & 0.829 & 0.267/0.087/0.196 & 0.834 & 0.239/0.078/0.175 & 0.822 & 0.248/0.083/0.189 & 0.831 & 0.211/0.074/0.157 & 0.842 & 0.226/0.075/0.169 & 0.842 & 0.213/0.072/0.165 & 0.847 & 0.217/0.074/0.172 & 0.843 \\
 & Biased output & 0.207/0.068/0.151 & 0.828 & 0.199/0.058/0.144 & 0.818 & 0.200/0.060/0.148 & 0.822 & 0.213/0.062/0.151 & 0.825 & 0.249/0.088/0.180 & 0.827 & 0.267/0.086/0.194 & 0.835 & 0.250/0.082/0.186 & 0.826 & 0.267/0.093/0.198 & 0.836 & 0.228/0.080/0.172 & 0.846 & 0.225/0.073/0.170 & 0.842 & 0.212/0.066/0.153 & 0.844 & 0.221/0.072/0.167 & 0.847 \\
 & Fair output & 0.200/0.065/0.150 & 0.820 & 0.207/0.064/0.145 & 0.821 & 0.198/0.058/0.143 & 0.818 & 0.205/0.061/0.146 & 0.824 & 0.245/0.075/0.177 & 0.830 & 0.271/0.089/0.199 & 0.838 & 0.250/0.085/0.186 & 0.823 & 0.263/0.088/0.192 & 0.837 & 0.218/0.071/0.160 & 0.845 & 0.223/0.077/0.169 & 0.842 & 0.213/0.074/0.156 & 0.847 & 0.213/0.073/0.165 & 0.843 \\
 & Mixed output & 0.203/0.062/0.148 & 0.821 & 0.201/0.061/0.150 & 0.822 & 0.191/0.056/0.133 & 0.820 & 0.148/0.041/0.111 & 0.794 & 0.254/0.078/0.184 & 0.832 & 0.264/0.083/0.196 & 0.835 & 0.239/0.067/0.170 & 0.827 & 0.267/0.086/0.200 & 0.833 & 0.221/0.079/0.166 & 0.848 & 0.221/0.074/0.166 & 0.843 & 0.200/0.065/0.161 & 0.840 & 0.212/0.069/0.167 & 0.844 \\
  \midrule
  \multirow{7}{*}{40\%} 
 & Left only & 0.232/0.073/0.162 & 0.831 & 0.184/0.056/0.131 & 0.815 & 0.192/0.068/0.148 & 0.821 & 0.202/0.065/0.149 & 0.816 & 0.248/0.079/0.183 & 0.825 & 0.246/0.081/0.179 & 0.826 & 0.209/0.069/0.160 & 0.812 & 0.259/0.083/0.191 & 0.832 & 0.201/0.073/0.153 & 0.843 & 0.201/0.071/0.170 & 0.843 & 0.191/0.067/0.155 & 0.842 & 0.189/0.066/0.157 & 0.844 \\
 & Right only & 0.207/0.069/0.150 & 0.821 & 0.188/0.064/0.143 & 0.815 & 0.226/0.075/0.166 & 0.831 & 0.199/0.063/0.152 & 0.824 & 0.247/0.084/0.184 & 0.826 & 0.255/0.081/0.188 & 0.828 & 0.223/0.079/0.168 & 0.815 & 0.243/0.084/0.184 & 0.825 & 0.223/0.075/0.170 & 0.846 & 0.201/0.072/0.167 & 0.842 & 0.209/0.072/0.175 & 0.843 & 0.189/0.061/0.158 & 0.842 \\
 & Fair input & 0.222/0.066/0.155 & 0.833 & 0.194/0.057/0.148 & 0.817 & 0.160/0.049/0.138 & 0.818 & 0.123/0.031/0.106 & 0.802 & 0.262/0.082/0.192 & 0.834 & 0.241/0.077/0.180 & 0.830 & 0.220/0.076/0.171 & 0.823 & 0.229/0.083/0.177 & 0.821 & 0.225/0.082/0.173 & 0.850 & 0.198/0.068/0.163 & 0.841 & 0.169/0.052/0.151 & 0.832 & 0.178/0.055/0.155 & 0.835 \\
 & Mixed input & 0.219/0.069/0.152 & 0.826 & 0.194/0.059/0.146 & 0.823 & 0.169/0.049/0.133 & 0.808 & 0.129/0.033/0.108 & 0.795 & 0.243/0.078/0.182 & 0.822 & 0.250/0.083/0.185 & 0.829 & 0.205/0.071/0.155 & 0.807 & 0.236/0.070/0.169 & 0.819 & 0.198/0.072/0.154 & 0.848 & 0.218/0.075/0.179 & 0.842 & 0.171/0.061/0.153 & 0.831 & 0.182/0.062/0.155 & 0.836 \\
 & Biased output & 0.206/0.070/0.153 & 0.824 & 0.218/0.067/0.159 & 0.822 & 0.216/0.068/0.164 & 0.830 & 0.185/0.062/0.141 & 0.812 & 0.251/0.082/0.191 & 0.828 & 0.255/0.082/0.190 & 0.834 & 0.230/0.078/0.171 & 0.826 & 0.253/0.084/0.184 & 0.832 & 0.230/0.085/0.179 & 0.851 & 0.211/0.066/0.174 & 0.840 & 0.208/0.073/0.160 & 0.845 & 0.209/0.073/0.173 & 0.844 \\
 & Fair output & 0.214/0.075/0.157 & 0.829 & 0.200/0.068/0.152 & 0.817 & 0.218/0.075/0.162 & 0.824 & 0.179/0.066/0.137 & 0.807 & 0.232/0.073/0.174 & 0.818 & 0.246/0.079/0.179 & 0.828 & 0.212/0.070/0.159 & 0.814 & 0.261/0.082/0.188 & 0.836 & 0.211/0.077/0.154 & 0.850 & 0.212/0.070/0.168 & 0.843 & 0.215/0.070/0.168 & 0.844 & 0.199/0.069/0.162 & 0.839 \\
 & Mixed output & 0.212/0.069/0.156 & 0.828 & 0.209/0.064/0.154 & 0.821 & 0.165/0.059/0.140 & 0.810 & 0.135/0.045/0.117 & 0.795 & 0.251/0.080/0.182 & 0.828 & 0.245/0.077/0.181 & 0.829 & 0.237/0.082/0.182 & 0.820 & 0.235/0.081/0.173 & 0.821 & 0.197/0.070/0.150 & 0.844 & 0.217/0.075/0.179 & 0.843 & 0.190/0.060/0.164 & 0.840 & 0.167/0.050/0.144 & 0.837 \\
 
\bottomrule
\end{tabular}%
}
\caption{Model performance across pruning methods and calibration sets---Political Tweet Summarisation}
\label{tab:performance_political}
\end{sidewaystable}

\begin{sidewaystable}
\tiny
\setlength{\tabcolsep}{2pt}
\renewcommand{\arraystretch}{0.9}
\centering
\resizebox{\textheight}{!}{%
\begin{tabular}{l*{29}{r}}
\toprule
& \multicolumn{9}{c}{Llama3-8B} & \multicolumn{9}{c}{Gemma-2B} & \multicolumn{9}{c}{TinyLlama} \\
\cmidrule(lr){3-10} \cmidrule(lr){11-18} \cmidrule(lr){19-26}
&  & \multicolumn{2}{c}{SparseGPT} & \multicolumn{2}{c}{Wanda} & \multicolumn{2}{c}{GBLM-Pruner} & \multicolumn{2}{c}{GBLM-Gradient} 
& \multicolumn{2}{c}{SparseGPT} & \multicolumn{2}{c}{Wanda} & \multicolumn{2}{c}{GBLM-Pruner} & \multicolumn{2}{c}{GBLM-Gradient}
& \multicolumn{2}{c}{SparseGPT} & \multicolumn{2}{c}{Wanda} & \multicolumn{2}{c}{GBLM-Pruner} & \multicolumn{2}{c}{GBLM-Gradient} \\
Sparsity Ratio & Calibration Set & ROUGE1/2/L & BERTScore & ROUGE1/2/L & BERTScore & ROUGE1/2/L & BERTScore & ROUGE1/2/L & BERTScore
& ROUGE1/2/L & BERTScore & ROUGE1/2/L & BERTScore & ROUGE1/2/L & BERTScore & ROUGE1/2/L & BERTScore & ROUGE1/2/L & BERTScore
& ROUGE1/2/L & BERTScore & ROUGE1/2/L & BERTScore & ROUGE1/2/L & BERTScore && \\
\midrule
\multirow{7}{*}{10\%} & Left only & 0.187/0.028/0.114 & 0.817 & 0.186/0.028/0.113 & 0.821 & 0.193/0.029/0.117 & 0.824 & 0.171/0.025/0.106 & 0.814 & 0.236/0.037/0.157 & 0.831 & 0.234/0.038/0.154 & 0.828 & 0.238/0.036/0.159 & 0.833 & 0.235/0.036/0.155 & 0.831 & 0.231/0.041/0.147 & 0.850 & 0.229/0.041/0.144 & 0.849 & 0.232/0.042/0.147 & 0.851 & 0.234/0.041/0.147 & 0.852\\
 & Right only & 0.210/0.031/0.126 & 0.834 & 0.193/0.028/0.116 & 0.822 & 0.206/0.031/0.125 & 0.832 & 0.184/0.027/0.112 & 0.815 & 0.233/0.037/0.155 & 0.829 & 0.232/0.037/0.153 & 0.827 & 0.235/0.037/0.158 & 0.833 & 0.234/0.037/0.155 & 0.831 & 0.238/0.042/0.150 & 0.851 & 0.230/0.041/0.146 & 0.850 & 0.235/0.043/0.148 & 0.852 & 0.243/0.045/0.151 & 0.854 \\
 & Fair input & 0.198/0.031/0.119 & 0.824 & 0.181/0.026/0.111 & 0.820 & 0.196/0.028/0.119 & 0.826 & 0.210/0.032/0.125 & 0.834 & 0.233/0.037/0.156 & 0.830 & 0.232/0.037/0.152 & 0.826 & 0.235/0.037/0.156 & 0.831 & 0.239/0.038/0.158 & 0.830 & 0.230/0.040/0.145 & 0.850 & 0.230/0.041/0.146 & 0.850 & 0.226/0.040/0.144 & 0.851 & 0.233/0.043/0.148 & 0.852 \\
 & Mixed input & 0.199/0.031/0.122 & 0.828 & 0.182/0.025/0.109 & 0.821 & 0.191/0.029/0.115 & 0.821 & 0.186/0.028/0.114 & 0.819 & 0.235/0.036/0.158 & 0.831 & 0.229/0.036/0.151 & 0.826 & 0.237/0.036/0.155 & 0.832 & 0.228/0.036/0.153 & 0.827 & 0.233/0.041/0.147 & 0.850 & 0.247/0.045/0.157 & 0.851 & 0.236/0.044/0.150 & 0.853 & 0.230/0.043/0.146 & 0.853 \\
 & Biased output & 0.187/0.026/0.113 & 0.819 & 0.195/0.027/0.118 & 0.828 & 0.202/0.029/0.121 & 0.830 & 0.180/0.027/0.109 & 0.814 & 0.231/0.035/0.154 & 0.829 & 0.234/0.037/0.153 & 0.828 & 0.229/0.035/0.154 & 0.828 & 0.237/0.038/0.158 & 0.831 & 0.231/0.041/0.146 & 0.850 & 0.232/0.041/0.147 & 0.850 & 0.236/0.043/0.149 & 0.852 & 0.235/0.043/0.150 & 0.852 \\
 & Fair output & 0.206/0.029/0.124 & 0.832 & 0.179/0.026/0.110 & 0.814 & 0.183/0.027/0.112 & 0.817 & 0.182/0.027/0.113 & 0.819 & 0.234/0.038/0.155 & 0.830 & 0.232/0.037/0.152 & 0.827 & 0.239/0.037/0.160 & 0.833 & 0.235/0.038/0.158 & 0.832 & 0.235/0.042/0.149 & 0.850 & 0.229/0.041/0.146 & 0.850 & 0.228/0.041/0.145 & 0.850 & 0.234/0.043/0.148 & 0.852 \\
 & Mixed output & 0.187/0.026/0.116 & 0.819 & 0.197/0.030/0.117 & 0.828 & 0.213/0.033/0.126 & 0.832 & 0.199/0.028/0.118 & 0.827 & 0.230/0.036/0.154 & 0.828 & 0.232/0.036/0.151 & 0.827 & 0.240/0.040/0.158 & 0.833 & 0.232/0.035/0.153 & 0.831 & 0.238/0.042/0.149 & 0.852 & 0.234/0.041/0.148 & 0.850 & 0.233/0.041/0.147 & 0.852 & 0.232/0.043/0.147 & 0.852 \\
  \midrule
 \multirow{7}{*}{20\%} 
 & Left only & 0.193/0.027/0.118 & 0.827 & 0.183/0.029/0.109 & 0.817 & 0.191/0.029/0.118 & 0.819 & 0.181/0.026/0.109 & 0.816 & 0.234/0.036/0.157 & 0.831 & 0.244/0.040/0.162 & 0.837 & 0.242/0.038/0.161 & 0.832 & 0.240/0.037/0.162 & 0.835 & 0.233/0.042/0.145 & 0.851 & 0.223/0.040/0.140 & 0.852 & 0.226/0.040/0.142 & 0.853 & 0.220/0.040/0.138 & 0.853 \\
 & Right only & 0.207/0.031/0.126 & 0.833 & 0.194/0.030/0.115 & 0.824 & 0.200/0.030/0.119 & 0.826 & 0.181/0.027/0.113 & 0.819 & 0.241/0.038/0.161 & 0.834 & 0.244/0.039/0.161 & 0.834 & 0.239/0.039/0.160 & 0.831 & 0.236/0.036/0.157 & 0.833 & 0.240/0.042/0.146 & 0.850 & 0.218/0.038/0.138 & 0.851 & 0.232/0.041/0.146 & 0.854 & 0.230/0.040/0.144 & 0.852 \\
 & Fair input & 0.207/0.029/0.121 & 0.831 & 0.205/0.031/0.120 & 0.826 & 0.209/0.031/0.128 & 0.832 & 0.198/0.029/0.123 & 0.827 & 0.246/0.039/0.167 & 0.838 & 0.244/0.040/0.162 & 0.835 & 0.242/0.038/0.161 & 0.834 & 0.242/0.037/0.159 & 0.832 & 0.237/0.040/0.144 & 0.850 & 0.222/0.039/0.140 & 0.852 & 0.228/0.042/0.141 & 0.855 & 0.226/0.040/0.141 & 0.853 \\
 & Mixed input & 0.204/0.032/0.118 & 0.827 & 0.197/0.030/0.118 & 0.825 & 0.175/0.026/0.110 & 0.816 & 0.176/0.027/0.112 & 0.818 & 0.242/0.038/0.166 & 0.836 & 0.244/0.038/0.161 & 0.837 & 0.241/0.037/0.161 & 0.835 & 0.240/0.037/0.157 & 0.831 & 0.237/0.038/0.143 & 0.851 & 0.246/0.045/0.155 & 0.854 & 0.237/0.042/0.151 & 0.852 & 0.245/0.046/0.154 & 0.853 \\
 & Biased output & 0.203/0.030/0.120 & 0.826 & 0.191/0.029/0.113 & 0.820 & 0.209/0.032/0.127 & 0.833 & 0.192/0.029/0.118 & 0.824 & 0.244/0.039/0.163 & 0.836 & 0.243/0.039/0.160 & 0.835 & 0.244/0.040/0.162 & 0.835 & 0.244/0.037/0.160 & 0.837 & 0.231/0.039/0.143 & 0.849 & 0.218/0.039/0.139 & 0.851 & 0.224/0.040/0.141 & 0.853 & 0.218/0.038/0.141 & 0.851 \\
 & Fair output & 0.190/0.028/0.115 & 0.820 & 0.207/0.031/0.124 & 0.832 & 0.183/0.026/0.112 & 0.820 & 0.177/0.026/0.109 & 0.817 & 0.241/0.038/0.160 & 0.831 & 0.245/0.038/0.161 & 0.835 & 0.235/0.039/0.160 & 0.831 & 0.244/0.040/0.162 & 0.836 & 0.235/0.041/0.148 & 0.850 & 0.217/0.037/0.137 & 0.850 & 0.219/0.039/0.134 & 0.852 & 0.227/0.040/0.142 & 0.852 \\
 & Mixed output & 0.194/0.031/0.114 & 0.823 & 0.182/0.026/0.106 & 0.814 & 0.189/0.028/0.117 & 0.824 & 0.206/0.033/0.124 & 0.830 & 0.239/0.038/0.159 & 0.833 & 0.249/0.040/0.164 & 0.836 & 0.231/0.032/0.153 & 0.833 & 0.231/0.037/0.156 & 0.833 & 0.232/0.039/0.141 & 0.850 & 0.218/0.038/0.137 & 0.851 & 0.230/0.042/0.150 & 0.851 & 0.237/0.045/0.155 & 0.854 \\
  \midrule
  \multirow{7}{*}{30\%} 
 & Left only & 0.206/0.030/0.120 & 0.832 & 0.159/0.023/0.097 & 0.806 & 0.168/0.026/0.107 & 0.808 & 0.165/0.024/0.105 & 0.810 & 0.232/0.035/0.156 & 0.828 & 0.240/0.039/0.160 & 0.832 & 0.234/0.035/0.158 & 0.829 & 0.243/0.039/0.160 & 0.830 & 0.233/0.042/0.147 & 0.853 & 0.231/0.041/0.142 & 0.851 & 0.222/0.040/0.140 & 0.851 & 0.210/0.039/0.135 & 0.850 \\
 & Right only & 0.193/0.030/0.114 & 0.824 & 0.150/0.022/0.088 & 0.797 & 0.169/0.023/0.105 & 0.812 & 0.163/0.024/0.101 & 0.809 & 0.245/0.040/0.163 & 0.835 & 0.250/0.043/0.166 & 0.839 & 0.233/0.036/0.156 & 0.832 & 0.248/0.039/0.164 & 0.834 & 0.218/0.039/0.140 & 0.854 & 0.227/0.041/0.140 & 0.851 & 0.224/0.040/0.140 & 0.850 & 0.221/0.042/0.140 & 0.851 \\
 & Fair input & 0.198/0.032/0.120 & 0.825 & 0.142/0.021/0.086 & 0.792 & 0.160/0.022/0.101 & 0.805 & 0.159/0.023/0.100 & 0.804 & 0.237/0.037/0.158 & 0.834 & 0.248/0.042/0.166 & 0.836 & 0.235/0.037/0.162 & 0.833 & 0.237/0.037/0.157 & 0.831 & 0.233/0.042/0.145 & 0.852 & 0.230/0.043/0.142 & 0.851 & 0.221/0.038/0.145 & 0.852 & 0.221/0.042/0.148 & 0.852 \\
 & Mixed input & 0.185/0.028/0.114 & 0.820 & 0.174/0.026/0.103 & 0.812 & 0.178/0.023/0.115 & 0.822 & 0.177/0.025/0.109 & 0.818 & 0.239/0.040/0.161 & 0.833 & 0.249/0.042/0.167 & 0.838 & 0.173/0.023/0.125 & 0.807 & 0.220/0.032/0.152 & 0.830 & 0.241/0.044/0.149 & 0.854 & 0.246/0.045/0.152 & 0.853 & 0.212/0.038/0.144 & 0.844 & 0.219/0.043/0.153 & 0.848 \\
 & Biased output & 0.182/0.029/0.111 & 0.818 & 0.140/0.020/0.085 & 0.791 & 0.141/0.020/0.090 & 0.795 & 0.177/0.024/0.112 & 0.815 & 0.241/0.039/0.160 & 0.831 & 0.253/0.043/0.167 & 0.840 & 0.224/0.033/0.155 & 0.830 & 0.244/0.038/0.161 & 0.837 & 0.235/0.042/0.146 & 0.854 & 0.233/0.042/0.145 & 0.851 & 0.224/0.041/0.142 & 0.855 & 0.223/0.040/0.142 & 0.852 \\
 & Fair output & 0.192/0.029/0.120 & 0.825 & 0.152/0.022/0.092 & 0.797 & 0.147/0.019/0.095 & 0.802 & 0.152/0.019/0.099 & 0.809 & 0.237/0.038/0.160 & 0.831 & 0.247/0.040/0.165 & 0.838 & 0.220/0.032/0.152 & 0.829 & 0.240/0.037/0.159 & 0.833 & 0.231/0.041/0.144 & 0.852 & 0.232/0.043/0.144 & 0.851 & 0.212/0.037/0.135 & 0.849 & 0.221/0.039/0.144 & 0.851 \\
 & Mixed output & 0.197/0.030/0.117 & 0.824 & 0.155/0.022/0.093 & 0.801 & 0.181/0.024/0.117 & 0.819 & 0.162/0.021/0.106 & 0.813 & 0.245/0.040/0.165 & 0.834 & 0.245/0.039/0.161 & 0.835 & 0.194/0.029/0.138 & 0.819 & 0.233/0.035/0.160 & 0.831 & 0.223/0.041/0.138 & 0.852 & 0.231/0.041/0.143 & 0.851 & 0.208/0.041/0.155 & 0.845 & 0.228/0.047/0.164 & 0.851 \\
  \midrule
  \multirow{7}{*}{40\%} 
 & Left only & 0.187/0.027/0.115 & 0.821 & 0.131/0.018/0.085 & 0.790 & 0.152/0.018/0.098 & 0.806 & 0.174/0.025/0.110 & 0.818 & 0.234/0.034/0.157 & 0.834 & 0.235/0.034/0.155 & 0.833 & 0.193/0.029/0.138 & 0.811 & 0.207/0.030/0.142 & 0.821 & 0.219/0.039/0.138 & 0.854 & 0.221/0.040/0.147 & 0.850 & 0.226/0.041/0.151 & 0.849 & 0.222/0.040/0.149 & 0.852 \\
 & Right only & 0.212/0.031/0.125 & 0.835 & 0.125/0.016/0.080 & 0.788 & 0.158/0.022/0.099 & 0.805 & 0.132/0.016/0.087 & 0.794 & 0.238/0.036/0.164 & 0.833 & 0.237/0.037/0.161 & 0.831 & 0.214/0.030/0.149 & 0.824 & 0.226/0.034/0.153 & 0.830 & 0.202/0.033/0.128 & 0.850 & 0.217/0.040/0.148 & 0.849 & 0.224/0.043/0.149 & 0.851 & 0.209/0.040/0.144 & 0.847 \\
 & Fair input & 0.140/0.019/0.089 & 0.794 & 0.100/0.011/0.068 & 0.777 & 0.128/0.017/0.090 & 0.796 & 0.130/0.017/0.087 & 0.791 & 0.243/0.038/0.163 & 0.835 & 0.232/0.036/0.155 & 0.829 & 0.207/0.029/0.146 & 0.820 & 0.242/0.036/0.161 & 0.836 & 0.208/0.036/0.129 & 0.850 & 0.222/0.041/0.148 & 0.851 & 0.222/0.043/0.148 & 0.852 & 0.206/0.039/0.151 & 0.848 \\
 & Mixed input & 0.186/0.028/0.112 & 0.820 & 0.127/0.016/0.077 & 0.790 & 0.117/0.014/0.088 & 0.792 & 0.103/0.011/0.085 & 0.784 & 0.238/0.039/0.161 & 0.830 & 0.224/0.034/0.152 & 0.826 & 0.138/0.014/0.109 & 0.800 & 0.163/0.020/0.123 & 0.800 & 0.214/0.038/0.136 & 0.852 & 0.236/0.046/0.159 & 0.852 & 0.163/0.029/0.130 & 0.831 & 0.185/0.032/0.146 & 0.833 \\
 & Biased output & 0.214/0.032/0.124 & 0.835 & 0.118/0.014/0.077 & 0.782 & 0.108/0.014/0.073 & 0.782 & 0.123/0.013/0.079 & 0.786 & 0.246/0.041/0.167 & 0.837 & 0.233/0.036/0.155 & 0.828 & 0.228/0.033/0.157 & 0.825 & 0.238/0.039/0.162 & 0.834 & 0.217/0.037/0.138 & 0.850 & 0.228/0.043/0.152 & 0.852 & 0.232/0.045/0.155 & 0.852 & 0.217/0.043/0.154 & 0.849 \\
 & Fair output & 0.210/0.031/0.123 & 0.836 & 0.134/0.017/0.083 & 0.790 & 0.148/0.021/0.099 & 0.803 & 0.163/0.021/0.108 & 0.814 & 0.235/0.035/0.158 & 0.832 & 0.238/0.035/0.161 & 0.834 & 0.210/0.031/0.145 & 0.821 & 0.226/0.031/0.153 & 0.829 & 0.212/0.035/0.133 & 0.851 & 0.220/0.042/0.149 & 0.851 & 0.219/0.041/0.147 & 0.849 & 0.226/0.040/0.145 & 0.848 \\
 & Mixed output & 0.208/0.034/0.123 & 0.830 & 0.136/0.017/0.086 & 0.794 & 0.123/0.015/0.096 & 0.796 & 0.092/0.010/0.072 & 0.788 & 0.241/0.040/0.164 & 0.830 & 0.236/0.034/0.154 & 0.831 & 0.141/0.017/0.112 & 0.796 & 0.183/0.025/0.135 & 0.812 & 0.219/0.038/0.139 & 0.849 & 0.235/0.044/0.154 & 0.851 & 0.169/0.028/0.135 & 0.831 & 0.176/0.034/0.146 & 0.835 \\
 
\bottomrule
\end{tabular}%
}
\caption{Model performance across pruning methods and calibration sets---Review Summarisation}
\label{tab:performance_review}
\end{sidewaystable}

\subsection{Model Fairness Across Pruning Methods and Calibration Sets}
\label{sec:model_fairness_various_prune_calibration}
In our fairness evaluation, we employ four metrics to comprehensively assess bias in summarisation models: Second-Order SPD (SPD), Binary Unfair Rate (BUR), Unfair Error Rate (UER), and Second-Order Fairness (SOF). For political tweet summarisation, these metrics reveal the model's tendency to represent different political perspectives, while in review summarisation, they expose potential sentiment biases.

The Second-Order SPD (SPD) directly measures the proportion of right versus left-leaning opinions, with a positive value indicating a right-leaning bias in political tweet summaries or a positive bias in review summaries. Complementing SPD, the Binary Unfair Rate (BUR) quantifies overall fairness by calculating the ratio of fair summaries to the total number of generated summaries. A lower BUR suggests more consistent representation across summaries.The Unfair Error Rate (UER) provides deeper insight by measuring the average discrepancy between the target and generated social value distributions. This metric captures the extent of underrepresentation, revealing how closely the generated summaries reflect the diverse perspectives present in source documents. Meanwhile, the Second-Order Fairness (SOF) examines the variance of UER across different values, highlighting which specific social perspectives experience more significant representation challenges.

Consistent with previous research \cite{zhang2023fair, huang-etal-2024-bias}, our analysis reveals an inherent bias in models towards left-leaning or positive opinions. We report the metric value for the vanilla model alongside each model name in Table~\ref{tab:various_input_fairness}, Table~\ref{tab:various_input_fairness_sof}, Table~\ref{tab:various_input_fairness_bur} and Table~\ref{tab:various_input_fairness_uer}, providing a clear baseline for comparison.
To assess fairness improvements, we calculate the absolute difference between the vanilla and pruned model's SPD, BUR, UER, and SOF metrics. A model demonstrating fairness enhancement will show a reduction in these metrics, indicating more balanced representation.

The results of pruned models, generated using various calibration sets, are detailed in Table~\ref{tab:various_input_fairness}, Table~\ref{tab:various_input_fairness_sof}, Table~\ref{tab:various_input_fairness_bur} and Table~\ref{tab:various_input_fairness_uer}. We specifically highlight instances where the model achieves meaningful improvements in fairness across multiple metrics, providing a multi-dimensional view of bias mitigation in summarisation models.

% For political tweet summarisation, we measure the difference between the proportions of right and left-leaning opinions with positive SPD indicating right-leaning bias and vice versa. For review summarisation, we measure the difference between the proportions of positive and negative opinions with positive SPD indicating positive bias and vice versa. The SPD value of the vanilla model is reported in brackets next to the model name in Table~\ref{tab:various_input_fairness}. Consistent with existing work \cite{zhang2023fair, huang-etal-2024-bias}, we find that models are inherently biased towards left-leaning or positive opinions, i.e., including more left-leaning opinions when summarising political tweets and more positive opinions when summarising reviews.

% We report the fairness improvement by calculating the absolute difference between the SPD of the vanilla model and the SPD of the pruned model. A model demonstrating a positive impact on fairness should have a difference ranging from 0 to its vanilla SPD.
% Results of pruned models using different calibration sets and their change compared to the original model is in Table~\ref{tab:various_input_fairness}. We highlighted areas where the model has an improvement in fairness compared to the original model generated summaries.

\begin{table*}[htbp]
\definecolor{veryLightGreen}{rgb}{0.85, 0.95, 0.85}
\definecolor{lightGreen}{rgb}{0.70, 0.85, 0.70}
\definecolor{mediumGreen}{rgb}{0.40, 0.70, 0.40}
\definecolor{mediumDarkGreen}{rgb}{0.20, 0.60, 0.20}
\definecolor{darkGreen}{rgb}{0.07, 0.53, 0.03}

\newcommand{\hlc}[2]{%
  \ifdim#1pt>0pt
    \ifdim#1pt<#2pt
      \cellcolor{%
        \ifnum\pdfstrcmp{\fpeval{#1/#2}}{\fpeval{0.2}}<0 veryLightGreen\else
        \ifnum\pdfstrcmp{\fpeval{#1/#2}}{\fpeval{0.4}}<0 lightGreen\else
        \ifnum\pdfstrcmp{\fpeval{#1/#2}}{\fpeval{0.6}}<0 mediumGreen\else
        \ifnum\pdfstrcmp{\fpeval{#1/#2}}{\fpeval{0.8}}<0 mediumDarkGreen\else
        darkGreen\fi\fi\fi\fi
      }%
    \fi
  \fi
  #1%
}
\centering
\tiny
\setlength{\tabcolsep}{4pt} % Reduce space between columns

\begin{subtable}{\textwidth}
\begin{tabular}{c c r r r r r r r r r r r r r r r}
\toprule
\multicolumn{2}{c}{} & \multicolumn{5}{c}{Llama3-8B (0.496)} & \multicolumn{5}{c}{Gemma-2B (0.785)} & \multicolumn{5}{c}{TinyLlama (0.382)} \\
\cmidrule(lr){3-7} \cmidrule(lr){8-12} \cmidrule(lr){13-17}
Sparsity & Calibration & \multicolumn{1}{c}{Sparse} & \multicolumn{1}{c}{} & \multicolumn{1}{c}{GBLM} & \multicolumn{1}{c}{GBLM} & \multicolumn{1}{c}{} & \multicolumn{1}{c}{Sparse} & \multicolumn{1}{c}{} & \multicolumn{1}{c}{GBLM} & \multicolumn{1}{c}{GBLM} & \multicolumn{1}{c}{} & \multicolumn{1}{c}{Sparse} & \multicolumn{1}{c}{} & \multicolumn{1}{c}{GBLM} & \multicolumn{1}{c}{GBLM} & \multicolumn{1}{c}{} \\
Ratio & Sets & \multicolumn{1}{c}{GPT} & \multicolumn{1}{c}{Wanda} & \multicolumn{1}{c}{Pruner} & \multicolumn{1}{c}{Gradient} & \multicolumn{1}{c}{HGLA} & \multicolumn{1}{c}{GPT} & \multicolumn{1}{c}{Wanda} & \multicolumn{1}{c}{Pruner} & \multicolumn{1}{c}{Gradient} & \multicolumn{1}{c}{HGLA} & \multicolumn{1}{c}{GPT} & \multicolumn{1}{c}{Wanda} & \multicolumn{1}{c}{Pruner} & \multicolumn{1}{c}{Gradient} & \multicolumn{1}{c}{HGLA} \\
\hline
\multirow{7}{*}{10\%} & Left only & \hlc{-0.081}{0.187} & \hlc{-0.132}{0.187} & \hlc{-0.179}{0.187} & \hlc{-0.178}{0.187} & \hlc{-0.232}{0.187} & \hlc{-0.059}{0.270} & \hlc{-0.014}{0.270} & \hlc{-0.107}{0.270} & \hlc{-0.043}{0.270} & \hlc{-0.035}{0.270} & \hlc{0.071}{0.088} & \hlc{0.006}{0.088} & \hlc{0.084}{0.088} & \hlc{-0.037}{0.088} & \hlc{0.065}{0.088} \\
 & Right only & \hlc{-0.117}{0.187} & \hlc{-0.106}{0.187} & \hlc{-0.060}{0.187} & \hlc{-0.211}{0.187} & \hlc{-0.060}{0.187} & \hlc{-0.081}{0.270} & \hlc{-0.060}{0.270} & \hlc{-0.004}{0.270} & \hlc{-0.097}{0.270} & \hlc{-0.046}{0.270} & \hlc{0.018}{0.088} & \hlc{-0.013}{0.088} & \hlc{0.071}{0.088} & \hlc{-0.052}{0.088} & \hlc{0.038}{0.088} \\
 & Fair input & \hlc{-0.082}{0.187} & \hlc{-0.130}{0.187} & \hlc{-0.268}{0.187} & \hlc{-0.189}{0.187} & \hlc{0.051}{0.187} & \hlc{-0.013}{0.270} & \hlc{-0.080}{0.270} & \hlc{-0.155}{0.270} & \hlc{0.041}{0.270} & \hlc{0.146}{0.270} & \hlc{-0.064}{0.088} & \hlc{0.045}{0.088} & \hlc{0.022}{0.088} & \hlc{0.039}{0.088} & \hlc{0.060}{0.088} \\
 & Mixed input & \hlc{-0.004}{0.187} & \hlc{-0.142}{0.187} & \hlc{-0.037}{0.187} & \hlc{-0.144}{0.187} & \hlc{0.056}{0.187} & \hlc{-0.106}{0.270} & \hlc{-0.035}{0.270} & \hlc{0.051}{0.270} & \hlc{-0.087}{0.270} & \hlc{0.130}{0.270} & \hlc{-0.056}{0.088} & \hlc{0.002}{0.088} & \hlc{0.062}{0.088} & \hlc{-0.029}{0.088} & \hlc{0.038}{0.088} \\
 & Biased output & \hlc{-0.077}{0.187} & \hlc{-0.073}{0.187} & \hlc{-0.128}{0.187} & \hlc{0.031}{0.187} & \hlc{-0.020}{0.187} & \hlc{0.025}{0.270} & \hlc{-0.080}{0.270} & \hlc{0.014}{0.270} & \hlc{-0.077}{0.270} & \hlc{0.145}{0.270} & \hlc{-0.053}{0.088} & \hlc{0.005}{0.088} & \hlc{-0.015}{0.088} & \hlc{-0.077}{0.088} & \hlc{0.063}{0.088} \\
 & Fair output & \hlc{-0.028}{0.187} & \hlc{-0.123}{0.187} & \hlc{-0.098}{0.187} & \hlc{-0.202}{0.187} & \hlc{-0.050}{0.187} & \hlc{-0.169}{0.270} & \hlc{-0.008}{0.270} & \hlc{-0.063}{0.270} & \hlc{-0.003}{0.270} & \hlc{0.152}{0.270} & \hlc{0.027}{0.088} & \hlc{0.018}{0.088} & \hlc{-0.036}{0.088} & \hlc{0.069}{0.088} & \hlc{0.032}{0.088} \\
 & Mixed output & \hlc{-0.075}{0.187} & \hlc{-0.119}{0.187} & \hlc{-0.134}{0.187} & \hlc{-0.108}{0.187} & \hlc{-0.030}{0.187} & \hlc{-0.123}{0.270} & \hlc{-0.105}{0.270} & \hlc{0.022}{0.270} & \hlc{-0.030}{0.270} & \hlc{0.104}{0.270} & \hlc{0.057}{0.088} & \hlc{-0.016}{0.088} & \hlc{-0.063}{0.088} & \hlc{0.013}{0.088} & \hlc{0.022}{0.088} \\
\hline
\multirow{7}{*}{20\%} & Left only & \hlc{-0.112}{0.187} & \hlc{-0.140}{0.187} & \hlc{-0.050}{0.187} & \hlc{-0.087}{0.187} & \hlc{-0.134}{0.187} & \hlc{-0.097}{0.270} & \hlc{0.039}{0.270} & \hlc{-0.059}{0.270} & \hlc{-0.047}{0.270} & \hlc{-0.059}{0.270} & \hlc{0.084}{0.088} & \hlc{-0.025}{0.088} & \hlc{-0.020}{0.088} & \hlc{0.052}{0.088} & \hlc{-0.037}{0.088} \\
 & Right only & \hlc{-0.238}{0.187} & \hlc{-0.142}{0.187} & \hlc{-0.245}{0.187} & \hlc{-0.218}{0.187} & \hlc{-0.004}{0.187} & \hlc{-0.157}{0.270} & \hlc{0.043}{0.270} & \hlc{0.032}{0.270} & \hlc{-0.073}{0.270} & \hlc{0.045}{0.270} & \hlc{-0.064}{0.088} & \hlc{-0.007}{0.088} & \hlc{-0.032}{0.088} & \hlc{0.027}{0.088} & \hlc{0.077}{0.088} \\
 & Fair input & \hlc{-0.125}{0.187} & \hlc{-0.093}{0.187} & \hlc{-0.117}{0.187} & \hlc{-0.100}{0.187} & \hlc{0.050}{0.187} & \hlc{-0.064}{0.270} & \hlc{0.083}{0.270} & \hlc{-0.112}{0.270} & \hlc{0.070}{0.270} & \hlc{0.151}{0.270} & \hlc{0.038}{0.088} & \hlc{0.013}{0.088} & \hlc{0.019}{0.088} & \hlc{0.046}{0.088} & \hlc{0.006}{0.088} \\
 & Mixed input & \hlc{-0.152}{0.187} & \hlc{-0.073}{0.187} & \hlc{-0.190}{0.187} & \hlc{-0.068}{0.187} & \hlc{0.000}{0.187} & \hlc{-0.205}{0.270} & \hlc{0.057}{0.270} & \hlc{0.107}{0.270} & \hlc{-0.005}{0.270} & \hlc{0.114}{0.270} & \hlc{-0.163}{0.088} & \hlc{0.023}{0.088} & \hlc{0.022}{0.088} & \hlc{-0.026}{0.088} & \hlc{0.080}{0.088} \\
 & Biased output & \hlc{-0.140}{0.187} & \hlc{-0.129}{0.187} & \hlc{-0.052}{0.187} & \hlc{-0.207}{0.187} & \hlc{0.042}{0.187} & \hlc{-0.145}{0.270} & \hlc{0.057}{0.270} & \hlc{-0.164}{0.270} & \hlc{0.091}{0.270} & \hlc{0.148}{0.270} & \hlc{0.020}{0.088} & \hlc{0.052}{0.088} & \hlc{0.016}{0.088} & \hlc{-0.103}{0.088} & \hlc{0.071}{0.088} \\
 & Fair output & \hlc{-0.216}{0.187} & \hlc{-0.300}{0.187} & \hlc{-0.017}{0.187} & \hlc{-0.222}{0.187} & \hlc{0.037}{0.187} & \hlc{-0.078}{0.270} & \hlc{0.058}{0.270} & \hlc{-0.010}{0.270} & \hlc{0.034}{0.270} & \hlc{0.111}{0.270} & \hlc{-0.037}{0.088} & \hlc{-0.039}{0.088} & \hlc{-0.013}{0.088} & \hlc{0.002}{0.088} & \hlc{0.016}{0.088} \\
 & Mixed output & \hlc{-0.282}{0.187} & \hlc{-0.074}{0.187} & \hlc{0.098}{0.187} & \hlc{0.006}{0.187} & \hlc{0.060}{0.187} & \hlc{-0.041}{0.270} & \hlc{0.087}{0.270} & \hlc{-0.107}{0.270} & \hlc{0.076}{0.270} & \hlc{0.118}{0.270} & \hlc{-0.134}{0.088} & \hlc{0.003}{0.088} & \hlc{-0.042}{0.088} & \hlc{0.042}{0.088} & \hlc{0.071}{0.088} \\
\hline
\multirow{7}{*}{30\%} & Left only & \hlc{-0.157}{0.187} & \hlc{0.183}{0.187} & \hlc{-0.191}{0.187} & \hlc{0.076}{0.187} & \hlc{0.043}{0.187} & \hlc{0.087}{0.270} & \hlc{0.095}{0.270} & \hlc{0.031}{0.270} & \hlc{0.061}{0.270} & \hlc{-0.205}{0.270} & \hlc{-0.112}{0.088} & \hlc{-0.015}{0.088} & \hlc{0.056}{0.088} & \hlc{0.046}{0.088} & \hlc{0.037}{0.088} \\
 & Right only & \hlc{0.140}{0.187} & \hlc{0.082}{0.187} & \hlc{-0.052}{0.187} & \hlc{-0.098}{0.187} & \hlc{0.119}{0.187} & \hlc{-0.083}{0.270} & \hlc{0.225}{0.270} & \hlc{-0.140}{0.270} & \hlc{-0.028}{0.270} & \hlc{-0.052}{0.270} & \hlc{0.049}{0.088} & \hlc{0.006}{0.088} & \hlc{-0.113}{0.088} & \hlc{-0.102}{0.088} & \hlc{-0.111}{0.088} \\
 & Fair input & \hlc{-0.053}{0.187} & \hlc{-0.058}{0.187} & \hlc{-0.160}{0.187} & \hlc{-0.130}{0.187} & \hlc{0.167}{0.187} & \hlc{-0.093}{0.270} & \hlc{0.123}{0.270} & \hlc{0.058}{0.270} & \hlc{0.006}{0.270} & \hlc{0.169}{0.270} & \hlc{-0.058}{0.088} & \hlc{0.013}{0.088} & \hlc{0.062}{0.088} & \hlc{0.075}{0.088} & \hlc{-0.076}{0.088} \\
 & Mixed input & \hlc{0.043}{0.187} & \hlc{-0.011}{0.187} & \hlc{0.132}{0.187} & \hlc{-0.395}{0.187} & \hlc{0.100}{0.187} & \hlc{-0.060}{0.270} & \hlc{0.129}{0.270} & \hlc{-0.084}{0.270} & \hlc{0.087}{0.270} & \hlc{0.057}{0.270} & \hlc{0.051}{0.088} & \hlc{0.000}{0.088} & \hlc{0.061}{0.088} & \hlc{0.053}{0.088} & \hlc{0.070}{0.088} \\
 & Biased output & \hlc{-0.078}{0.187} & \hlc{0.098}{0.187} & \hlc{-0.074}{0.187} & \hlc{-0.278}{0.187} & \hlc{0.135}{0.187} & \hlc{-0.115}{0.270} & \hlc{0.110}{0.270} & \hlc{-0.001}{0.270} & \hlc{0.070}{0.270} & \hlc{0.129}{0.270} & \hlc{-0.163}{0.088} & \hlc{0.071}{0.088} & \hlc{-0.045}{0.088} & \hlc{-0.042}{0.088} & \hlc{-0.024}{0.088} \\
 & Fair output & \hlc{-0.066}{0.187} & \hlc{0.128}{0.187} & \hlc{-0.118}{0.187} & \hlc{-0.134}{0.187} & \hlc{0.110}{0.187} & \hlc{-0.123}{0.270} & \hlc{0.120}{0.270} & \hlc{-0.135}{0.270} & \hlc{0.025}{0.270} & \hlc{0.075}{0.270} & \hlc{0.074}{0.088} & \hlc{0.075}{0.088} & \hlc{-0.139}{0.088} & \hlc{0.030}{0.088} & \hlc{0.009}{0.088} \\
 & Mixed output & \hlc{-0.151}{0.187} & \hlc{0.004}{0.187} & \hlc{0.095}{0.187} & \hlc{-0.261}{0.187} & \hlc{0.091}{0.187} & \hlc{0.049}{0.270} & \hlc{0.043}{0.270} & \hlc{0.074}{0.270} & \hlc{0.167}{0.270} & \hlc{0.076}{0.270} & \hlc{-0.131}{0.088} & \hlc{0.016}{0.088} & \hlc{0.084}{0.088} & \hlc{-0.216}{0.088} & \hlc{0.023}{0.088} \\
\hline
\multirow{7}{*}{40\%} & Left only & \hlc{0.122}{0.187} & \hlc{-0.156}{0.187} & \hlc{-0.092}{0.187} & \hlc{-0.070}{0.187} & \hlc{0.092}{0.187} & \hlc{-0.086}{0.270} & \hlc{-0.030}{0.270} & \hlc{0.025}{0.270} & \hlc{-0.019}{0.270} & \hlc{-0.153}{0.270} & \hlc{-0.124}{0.088} & \hlc{-0.082}{0.088} & \hlc{-0.229}{0.088} & \hlc{0.022}{0.088} & \hlc{0.040}{0.088} \\
 & Right only & \hlc{0.093}{0.187} & \hlc{-0.128}{0.187} & \hlc{0.129}{0.187} & \hlc{-0.110}{0.187} & \hlc{0.170}{0.187} & \hlc{0.097}{0.270} & \hlc{0.038}{0.270} & \hlc{-0.026}{0.270} & \hlc{-0.097}{0.270} & \hlc{0.112}{0.270} & \hlc{-0.243}{0.088} & \hlc{0.006}{0.088} & \hlc{-0.079}{0.088} & \hlc{0.055}{0.088} & \hlc{-0.258}{0.088} \\
 & Fair input & \hlc{0.074}{0.187} & \hlc{-0.128}{0.187} & \hlc{-0.456}{0.187} & \hlc{-0.483}{0.187} & \hlc{0.009}{0.187} & \hlc{-0.003}{0.270} & \hlc{0.079}{0.270} & \hlc{0.010}{0.270} & \hlc{-0.042}{0.270} & \hlc{0.173}{0.270} & \hlc{-0.004}{0.088} & \hlc{0.044}{0.088} & \hlc{-0.523}{0.088} & \hlc{-0.229}{0.088} & \hlc{0.042}{0.088} \\
 & Mixed input & \hlc{0.103}{0.187} & \hlc{-0.040}{0.187} & \hlc{-0.414}{0.187} & \hlc{-0.544}{0.187} & \hlc{0.181}{0.187} & \hlc{-0.024}{0.270} & \hlc{0.013}{0.270} & \hlc{-0.128}{0.270} & \hlc{-0.048}{0.270} & \hlc{0.101}{0.270} & \hlc{0.010}{0.088} & \hlc{-0.062}{0.088} & \hlc{-0.350}{0.088} & \hlc{-0.326}{0.088} & \hlc{-0.076}{0.088} \\
 & Biased output & \hlc{-0.082}{0.187} & \hlc{-0.108}{0.187} & \hlc{-0.040}{0.187} & \hlc{0.002}{0.187} & \hlc{0.101}{0.187} & \hlc{-0.262}{0.270} & \hlc{-0.001}{0.270} & \hlc{-0.049}{0.270} & \hlc{-0.009}{0.270} & \hlc{0.106}{0.270} & \hlc{0.004}{0.088} & \hlc{0.080}{0.088} & \hlc{-0.073}{0.088} & \hlc{-0.040}{0.088} & \hlc{-0.136}{0.088} \\
 & Fair output & \hlc{-0.055}{0.187} & \hlc{-0.632}{0.187} & \hlc{-0.130}{0.187} & \hlc{-0.115}{0.187} & \hlc{0.101}{0.187} & \hlc{-0.139}{0.270} & \hlc{0.114}{0.270} & \hlc{-0.107}{0.270} & \hlc{-0.068}{0.270} & \hlc{0.179}{0.270} & \hlc{0.138}{0.088} & \hlc{0.084}{0.088} & \hlc{0.005}{0.088} & \hlc{-0.077}{0.088} & \hlc{-0.015}{0.088} \\
 & Mixed output & \hlc{-0.099}{0.187} & \hlc{-0.243}{0.187} & \hlc{-0.476}{0.187} & \hlc{-0.403}{0.187} & \hlc{0.146}{0.187} & \hlc{-0.089}{0.270} & \hlc{0.095}{0.270} & \hlc{0.123}{0.270} & \hlc{0.105}{0.270} & \hlc{0.160}{0.270} & \hlc{-0.055}{0.088} & \hlc{-0.047}{0.088} & \hlc{-0.168}{0.088} & \hlc{-0.228}{0.088} & \hlc{-0.036}{0.088} \\
\bottomrule
\end{tabular}
\caption{Political tweet summarisation}
\end{subtable}

\bigskip

\begin{subtable}{\textwidth}
\begin{tabular}{c c r r r r r r r r r r r r r r r}
\toprule
\multicolumn{2}{c}{} & \multicolumn{5}{c}{Llama3-8B (0.496)} & \multicolumn{5}{c}{Gemma-2B (0.785)} & \multicolumn{5}{c}{TinyLlama (0.382)} \\
\cmidrule(lr){3-7} \cmidrule(lr){8-12} \cmidrule(lr){13-17}
Sparsity & Calibration & \multicolumn{1}{c}{Sparse} & \multicolumn{1}{c}{} & \multicolumn{1}{c}{GBLM} & \multicolumn{1}{c}{GBLM} & \multicolumn{1}{c}{} & \multicolumn{1}{c}{Sparse} & \multicolumn{1}{c}{} & \multicolumn{1}{c}{GBLM} & \multicolumn{1}{c}{GBLM} & \multicolumn{1}{c}{} & \multicolumn{1}{c}{Sparse} & \multicolumn{1}{c}{} & \multicolumn{1}{c}{GBLM} & \multicolumn{1}{c}{GBLM} & \multicolumn{1}{c}{} \\
Ratio & Sets & \multicolumn{1}{c}{GPT} & \multicolumn{1}{c}{Wanda} & \multicolumn{1}{c}{Pruner} & \multicolumn{1}{c}{Gradient} & \multicolumn{1}{c}{HGLA} & \multicolumn{1}{c}{GPT} & \multicolumn{1}{c}{Wanda} & \multicolumn{1}{c}{Pruner} & \multicolumn{1}{c}{Gradient} & \multicolumn{1}{c}{HGLA} & \multicolumn{1}{c}{GPT} & \multicolumn{1}{c}{Wanda} & \multicolumn{1}{c}{Pruner} & \multicolumn{1}{c}{Gradient} & \multicolumn{1}{c}{HGLA} \\
\hline
\multirow{7}{*}{10\%} 
& Negative only & \hlc{0.025}{0.496} & \hlc{0.115}{0.496} & \hlc{0.080}{0.496} & \hlc{0.056}{0.496} & \hlc{0.085}{0.496} & \hlc{-0.009}{0.785} & \hlc{0.000}{0.785} & \hlc{-0.018}{0.785} & \hlc{-0.001}{0.785} & \hlc{-0.013}{0.785} & \hlc{0.008}{0.382} & \hlc{0.040}{0.382} & \hlc{0.025}{0.382} & \hlc{0.028}{0.382} & \hlc{0.029}{0.382}\\
 & Positive only & \hlc{0.002}{0.496} & \hlc{0.081}{0.496} & \hlc{0.110}{0.496} & \hlc{0.058}{0.496} & \hlc{0.020}{0.496} & \hlc{-0.021}{0.785} & \hlc{-0.001}{0.785} & \hlc{0.015}{0.785} & \hlc{0.014}{0.785} & \hlc{-0.013}{0.785} & \hlc{0.014}{0.382} & \hlc{0.016}{0.382} & \hlc{0.027}{0.382} & \hlc{0.002}{0.382} & \hlc{0.031}{0.382}\\
 & Fair input & \hlc{0.049}{0.496} & \hlc{0.014}{0.496} & \hlc{0.072}{0.496} & \hlc{0.062}{0.496} & \hlc{0.298}{0.496} & \hlc{-0.008}{0.785} & \hlc{-0.001}{0.785} & \hlc{0.007}{0.785} & \hlc{0.015}{0.785} & \hlc{0.378}{0.785} & \hlc{0.025}{0.382} & \hlc{0.016}{0.382} & \hlc{0.021}{0.382} & \hlc{-0.013}{0.382} & \hlc{0.205}{0.382} \\
 & Mixed input & \hlc{0.038}{0.496} & \hlc{0.082}{0.496} & \hlc{0.041}{0.496} & \hlc{0.037}{0.496} & \hlc{0.283}{0.496} & \hlc{0.008}{0.785} & \hlc{0.001}{0.785} & \hlc{-0.008}{0.785} & \hlc{0.014}{0.785} & \hlc{0.394}{0.785} & \hlc{0.018}{0.382} & \hlc{0.020}{0.382} & \hlc{0.050}{0.382} & \hlc{0.002}{0.382} & \hlc{0.204}{0.382}\\
 & Biased output & \hlc{0.094}{0.496} & \hlc{0.056}{0.496} & \hlc{0.094}{0.496} & \hlc{0.062}{0.496} & \hlc{0.285}{0.496} & \hlc{-0.003}{0.785} & \hlc{-0.002}{0.785} & \hlc{0.008}{0.785} & \hlc{0.002}{0.785} & \hlc{0.378}{0.785} & \hlc{0.013}{0.382} & \hlc{0.041}{0.382} & \hlc{0.019}{0.382} & \hlc{0.029}{0.382} & \hlc{0.212}{0.382}\\
 & Fair output & \hlc{0.057}{0.496} & \hlc{0.073}{0.496} & \hlc{0.047}{0.496} & \hlc{0.072}{0.496} & \hlc{0.289}{0.496} & \hlc{-0.010}{0.785} & \hlc{-0.012}{0.785} & \hlc{0.002}{0.785} & \hlc{0.012}{0.785} & \hlc{0.393}{0.785} & \hlc{0.013}{0.382} & \hlc{0.017}{0.382} & \hlc{0.027}{0.382} & \hlc{0.015}{0.382} & \hlc{0.203}{0.382}\\
 & Mixed output & \hlc{0.079}{0.496} & \hlc{0.105}{0.496} & \hlc{0.070}{0.496} & \hlc{0.083}{0.496} & \hlc{0.291}{0.496} & \hlc{0.001}{0.785} & \hlc{-0.002}{0.785} & \hlc{0.019}{0.785} & \hlc{0.021}{0.785} & \hlc{0.390}{0.785} & \hlc{0.005}{0.382} & \hlc{0.021}{0.382} & \hlc{0.040}{0.382} & \hlc{0.000}{0.382} & \hlc{0.202}{0.382}\\
\hline
\multirow{7}{*}{20\%} & 
Negative only & \hlc{0.140}{0.496} & \hlc{0.102}{0.496} & \hlc{0.044}{0.496} & \hlc{0.060}{0.496} & \hlc{0.145}{0.496} & \hlc{-0.043}{0.785} & \hlc{0.033}{0.785} & \hlc{0.015}{0.785} & \hlc{0.057}{0.785} & \hlc{-0.034}{0.785} & \hlc{0.034}{0.382} & \hlc{0.038}{0.382} & \hlc{0.042}{0.382} & \hlc{0.075}{0.382} & \hlc{0.080}{0.382} \\
 & Positive only & \hlc{0.103}{0.496} & \hlc{0.130}{0.496} & \hlc{0.076}{0.496} & \hlc{0.051}{0.496} & \hlc{0.097}{0.496} & \hlc{-0.039}{0.785} & \hlc{0.029}{0.785} & \hlc{-0.036}{0.785} & \hlc{0.008}{0.785} & \hlc{-0.028}{0.785} & \hlc{0.050}{0.382} & \hlc{0.045}{0.382} & \hlc{0.065}{0.382} & \hlc{0.059}{0.382} & \hlc{0.077}{0.382} \\
 & Fair input & \hlc{0.125}{0.496} & \hlc{0.087}{0.496} & \hlc{0.060}{0.496} & \hlc{0.052}{0.496} & \hlc{0.299}{0.496} & \hlc{-0.018}{0.785} & \hlc{0.019}{0.785} & \hlc{-0.007}{0.785} & \hlc{0.034}{0.785} & \hlc{0.367}{0.785} & \hlc{0.030}{0.382} & \hlc{0.036}{0.382} & \hlc{0.060}{0.382} & \hlc{0.063}{0.382} & \hlc{0.244}{0.382} \\
 & Mixed input & \hlc{0.026}{0.496} & \hlc{0.126}{0.496} & \hlc{0.056}{0.496} & \hlc{-0.017}{0.496} & \hlc{0.306}{0.496} & \hlc{-0.020}{0.785} & \hlc{0.006}{0.785} & \hlc{-0.016}{0.785} & \hlc{-0.012}{0.785} & \hlc{0.387}{0.785} & \hlc{0.048}{0.382} & \hlc{0.067}{0.382} & \hlc{0.079}{0.382} & \hlc{0.050}{0.382} & \hlc{0.192}{0.382} \\
 & Biased output & \hlc{0.097}{0.496} & \hlc{0.135}{0.496} & \hlc{0.066}{0.496} & \hlc{0.049}{0.496} & \hlc{0.284}{0.496} & \hlc{-0.039}{0.785} & \hlc{0.036}{0.785} & \hlc{0.010}{0.785} & \hlc{0.014}{0.785} & \hlc{0.387}{0.785} & \hlc{0.031}{0.382} & \hlc{0.049}{0.382} & \hlc{0.023}{0.382} & \hlc{0.018}{0.382} & \hlc{0.219}{0.382} \\
 & Fair output & \hlc{0.093}{0.496} & \hlc{0.118}{0.496} & \hlc{0.083}{0.496} & \hlc{0.041}{0.496} & \hlc{0.269}{0.496} & \hlc{-0.015}{0.785} & \hlc{0.041}{0.785} & \hlc{-0.018}{0.785} & \hlc{0.060}{0.785} & \hlc{0.385}{0.785} & \hlc{0.035}{0.382} & \hlc{0.069}{0.382} & \hlc{0.012}{0.382} & \hlc{0.014}{0.382} & \hlc{0.232}{0.382} \\
 & Mixed output & \hlc{0.104}{0.496} & \hlc{0.080}{0.496} & \hlc{0.063}{0.496} & \hlc{0.102}{0.496} & \hlc{0.288}{0.496} & \hlc{-0.041}{0.785} & \hlc{0.041}{0.785} & \hlc{-0.077}{0.785} & \hlc{-0.033}{0.785} & \hlc{0.386}{0.785} & \hlc{0.042}{0.382} & \hlc{0.060}{0.382} & \hlc{0.120}{0.382} & \hlc{0.018}{0.382} & \hlc{0.206}{0.382} \\
\hline
\multirow{7}{*}{30\%} & Negative only & \hlc{0.056}{0.496} & \hlc{0.001}{0.496} & \hlc{-0.024}{0.496} & \hlc{-0.072}{0.496} & \hlc{0.114}{0.496} & \hlc{0.018}{0.785} & \hlc{-0.023}{0.785} & \hlc{-0.120}{0.785} & \hlc{-0.015}{0.785} & \hlc{-0.004}{0.785} & \hlc{0.018}{0.382} & \hlc{0.092}{0.382} & \hlc{0.150}{0.382} & \hlc{0.079}{0.382} & \hlc{0.101}{0.382} \\
 & Positive only & \hlc{0.102}{0.496} & \hlc{-0.004}{0.496} & \hlc{-0.010}{0.496} & \hlc{-0.015}{0.496} & \hlc{0.077}{0.496} & \hlc{0.010}{0.785} & \hlc{0.015}{0.785} & \hlc{-0.069}{0.785} & \hlc{-0.038}{0.785} & \hlc{-0.032}{0.785} & \hlc{-0.018}{0.382} & \hlc{0.070}{0.382} & \hlc{0.057}{0.382} & \hlc{0.049}{0.382} & \hlc{0.064}{0.382} \\
 & Fair input & \hlc{0.059}{0.496} & \hlc{-0.014}{0.496} & \hlc{-0.034}{0.496} & \hlc{-0.027}{0.496} & \hlc{0.307}{0.496} & \hlc{0.016}{0.785} & \hlc{-0.024}{0.785} & \hlc{-0.053}{0.785} & \hlc{-0.045}{0.785} & \hlc{0.373}{0.785} & \hlc{0.045}{0.382} & \hlc{0.068}{0.382} & \hlc{0.020}{0.382} & \hlc{0.011}{0.382} & \hlc{0.246}{0.382} \\
 & Mixed input & \hlc{0.043}{0.496} & \hlc{0.051}{0.496} & \hlc{0.046}{0.496} & \hlc{-0.079}{0.496} & \hlc{0.278}{0.496} & \hlc{-0.011}{0.785} & \hlc{-0.054}{0.785} & \hlc{0.006}{0.785} & \hlc{-0.009}{0.785} & \hlc{0.357}{0.785} & \hlc{0.082}{0.382} & \hlc{0.036}{0.382} & \hlc{0.026}{0.382} & \hlc{0.074}{0.382} & \hlc{0.232}{0.382} \\
 & Biased output & \hlc{0.126}{0.496} & \hlc{-0.004}{0.496} & \hlc{0.023}{0.496} & \hlc{0.000}{0.496} & \hlc{0.262}{0.496} & \hlc{-0.051}{0.785} & \hlc{-0.026}{0.785} & \hlc{-0.037}{0.785} & \hlc{-0.084}{0.785} & \hlc{0.406}{0.785} & \hlc{0.021}{0.382} & \hlc{0.071}{0.382} & \hlc{0.049}{0.382} & \hlc{0.017}{0.382} & \hlc{0.217}{0.382} \\
 & Fair output & \hlc{0.096}{0.496} & \hlc{0.016}{0.496} & \hlc{-0.107}{0.496} & \hlc{-0.048}{0.496} & \hlc{0.256}{0.496} & \hlc{0.015}{0.785} & \hlc{-0.026}{0.785} & \hlc{-0.067}{0.785} & \hlc{-0.038}{0.785} & \hlc{0.382}{0.785} & \hlc{0.055}{0.382} & \hlc{0.087}{0.382} & \hlc{0.019}{0.382} & \hlc{0.068}{0.382} & \hlc{0.245}{0.382} \\
 & Mixed output & \hlc{0.119}{0.496} & \hlc{0.015}{0.496} & \hlc{-0.058}{0.496} & \hlc{-0.098}{0.496} & \hlc{0.272}{0.496} & \hlc{-0.012}{0.785} & \hlc{-0.035}{0.785} & \hlc{-0.059}{0.785} & \hlc{-0.029}{0.785} & \hlc{0.354}{0.785} & \hlc{0.007}{0.382} & \hlc{0.093}{0.382} & \hlc{0.188}{0.382} & \hlc{0.145}{0.382} & \hlc{0.209}{0.382} \\
\hline
\multirow{7}{*}{40\%} & Negative only & \hlc{0.078}{0.496} & \hlc{-0.050}{0.496} & \hlc{-0.112}{0.496} & \hlc{-0.110}{0.496} & \hlc{-0.112}{0.496} & \hlc{-0.045}{0.785} & \hlc{0.027}{0.785} & \hlc{-0.011}{0.785} & \hlc{-0.033}{0.785} & \hlc{0.219}{0.785} & \hlc{-0.018}{0.382} & \hlc{0.072}{0.382} & \hlc{0.091}{0.382} & \hlc{0.081}{0.382} & \hlc{-0.065}{0.382} \\
 & Positive only & \hlc{0.199}{0.496} & \hlc{-0.045}{0.496} & \hlc{-0.071}{0.496} & \hlc{-0.150}{0.496} & \hlc{-0.040}{0.496} & \hlc{-0.039}{0.785} & \hlc{0.008}{0.785} & \hlc{0.023}{0.785} & \hlc{-0.066}{0.785} & \hlc{0.140}{0.785} & \hlc{-0.005}{0.382} & \hlc{0.126}{0.382} & \hlc{0.126}{0.382} & \hlc{0.056}{0.382} & \hlc{0.061}{0.382} \\
 & Fair input & \hlc{-0.010}{0.496} & \hlc{-0.001}{0.496} & \hlc{-0.099}{0.496} & \hlc{-0.101}{0.496} & \hlc{0.234}{0.496} & \hlc{-0.058}{0.785} & \hlc{0.013}{0.785} & \hlc{-0.039}{0.785} & \hlc{-0.072}{0.785} & \hlc{0.429}{0.785} & \hlc{0.015}{0.382} & \hlc{0.101}{0.382} & \hlc{0.069}{0.382} & \hlc{0.055}{0.382} & \hlc{0.191}{0.382} \\
 & Mixed input & \hlc{0.110}{0.496} & \hlc{-0.031}{0.496} & \hlc{-0.232}{0.496} & \hlc{-0.223}{0.496} & \hlc{0.112}{0.496} & \hlc{-0.032}{0.785} & \hlc{0.024}{0.785} & \hlc{0.044}{0.785} & \hlc{0.020}{0.785} & \hlc{0.412}{0.785} & \hlc{0.019}{0.382} & \hlc{0.106}{0.382} & \hlc{0.051}{0.382} & \hlc{0.142}{0.382} & \hlc{0.212}{0.382} \\
 & Biased output & \hlc{0.171}{0.496} & \hlc{-0.028}{0.496} & \hlc{-0.163}{0.496} & \hlc{-0.097}{0.496} & \hlc{0.190}{0.496} & \hlc{-0.067}{0.785} & \hlc{0.004}{0.785} & \hlc{-0.026}{0.785} & \hlc{0.006}{0.785} & \hlc{0.437}{0.785} & \hlc{0.004}{0.382} & \hlc{0.088}{0.382} & \hlc{0.111}{0.382} & \hlc{0.045}{0.382} & \hlc{0.186}{0.382} \\
 & Fair output & \hlc{0.076}{0.496} & \hlc{0.027}{0.496} & \hlc{-0.107}{0.496} & \hlc{-0.036}{0.496} & \hlc{0.206}{0.496} & \hlc{-0.009}{0.785} & \hlc{-0.003}{0.785} & \hlc{-0.002}{0.785} & \hlc{-0.063}{0.785} & \hlc{0.475}{0.785} & \hlc{0.013}{0.382} & \hlc{0.044}{0.382} & \hlc{0.086}{0.382} & \hlc{0.060}{0.382} & \hlc{0.187}{0.382} \\
 & Mixed output & \hlc{0.060}{0.496} & \hlc{-0.047}{0.496} & \hlc{-0.191}{0.496} & \hlc{-0.135}{0.496} & \hlc{0.120}{0.496} & \hlc{-0.087}{0.785} & \hlc{0.023}{0.785} & \hlc{0.159}{0.785} & \hlc{0.025}{0.785} & \hlc{0.438}{0.785} & \hlc{0.035}{0.382} & \hlc{0.103}{0.382} & \hlc{0.037}{0.382} & \hlc{0.219}{0.382} & \hlc{0.199}{0.382} \\
\bottomrule
\end{tabular}
\caption{Review summarisation}
\label{tab:review_various_input_fairness}
\end{subtable} 

\caption{SPD---Comparison of sparsity ratio, calibration sets, and model fairness using different datasets. The SPD value of the vanilla model is reported in brackets next to the model name. For political tweet summarisation, a positive vanilla SPD indicates the model is biased towards the right and negative indicates biased towards the left. For review summarisation a positive vanilla SPD indicates the model is biased towards positive views and negative SPD indicates biased towards negative views. We find that models are inherently biased towards left-leaning or positive opinions, including more left-leaning opinions when summarising political tweets and more positive opinions when summarising reviews. We report fairness improvement by calculating the absolute difference between the Statistical Parity Difference (SPD) of the vanilla model and that of the pruned model. A model demonstrating a positive impact on fairness should have an absolute difference ranging from 0 to its vanilla SPD, with values closer to the vanilla SPD indicating better improvement (values between 0 and vanilla SPD are highlighted, indicating that the pruned model is less biased than the original model). Darker colours indicate greater improvement in fairness.
}
\label{tab:various_input_fairness}
\end{table*}

\begin{table*}[htbp]
\definecolor{veryLightGreen}{rgb}{0.85, 0.95, 0.85}
\definecolor{lightGreen}{rgb}{0.70, 0.85, 0.70}
\definecolor{mediumGreen}{rgb}{0.40, 0.70, 0.40}
\definecolor{mediumDarkGreen}{rgb}{0.20, 0.60, 0.20}
\definecolor{darkGreen}{rgb}{0.07, 0.53, 0.03}

\newcommand{\hlc}[2]{%
  \ifdim#1pt>0pt
    \cellcolor{%
      \ifnum\pdfstrcmp{\fpeval{#1/#2}}{\fpeval{0.2}}<0 veryLightGreen\else
      \ifnum\pdfstrcmp{\fpeval{#1/#2}}{\fpeval{0.4}}<0 lightGreen\else
      \ifnum\pdfstrcmp{\fpeval{#1/#2}}{\fpeval{0.6}}<0 mediumGreen\else
      \ifnum\pdfstrcmp{\fpeval{#1/#2}}{\fpeval{0.8}}<0 mediumDarkGreen\else
      darkGreen\fi\fi\fi\fi
    }%
  \fi
  #1%
}
\centering
\tiny
\setlength{\tabcolsep}{4pt} % Reduce space between columns
\begin{subtable}{\textwidth}
\begin{tabular}{c c r r r r r r r r r r r r r r r}
\toprule
\multicolumn{2}{c}{} & \multicolumn{5}{c}{Llama3-8B (0.496)} & \multicolumn{5}{c}{Gemma-2B (0.785)} & \multicolumn{5}{c}{TinyLlama (0.382)} \\
\cmidrule(lr){3-7} \cmidrule(lr){8-12} \cmidrule(lr){13-17}
Sparsity & Calibration & \multicolumn{1}{c}{Sparse} & \multicolumn{1}{c}{} & \multicolumn{1}{c}{GBLM} & \multicolumn{1}{c}{GBLM} & \multicolumn{1}{c}{} & \multicolumn{1}{c}{Sparse} & \multicolumn{1}{c}{} & \multicolumn{1}{c}{GBLM} & \multicolumn{1}{c}{GBLM} & \multicolumn{1}{c}{} & \multicolumn{1}{c}{Sparse} & \multicolumn{1}{c}{} & \multicolumn{1}{c}{GBLM} & \multicolumn{1}{c}{GBLM} & \multicolumn{1}{c}{} \\
Ratio & Sets & \multicolumn{1}{c}{GPT} & \multicolumn{1}{c}{Wanda} & \multicolumn{1}{c}{Pruner} & \multicolumn{1}{c}{Gradient} & \multicolumn{1}{c}{HGLA} & \multicolumn{1}{c}{GPT} & \multicolumn{1}{c}{Wanda} & \multicolumn{1}{c}{Pruner} & \multicolumn{1}{c}{Gradient} & \multicolumn{1}{c}{HGLA} & \multicolumn{1}{c}{GPT} & \multicolumn{1}{c}{Wanda} & \multicolumn{1}{c}{Pruner} & \multicolumn{1}{c}{Gradient} & \multicolumn{1}{c}{HGLA} \\
\hline
\multirow{7}{*}{10\%} & Left only & \hlc{-0.007}{0.011} & \hlc{0.001}{0.011} & \hlc{-0.009}{0.011} & \hlc{0.002}{0.011} & \hlc{-0.003}{0.011} & \hlc{0.007}{0.013} & \hlc{0.006}{0.013} & \hlc{-0.013}{0.013} & \hlc{0.001}{0.013} & \hlc{0.005}{0.013} & \hlc{-0.008}{0.015} & \hlc{-0.012}{0.015} & \hlc{-0.003}{0.015} & \hlc{-0.008}{0.015} & \hlc{-0.012}{0.015} \\
 & Right only & \hlc{-0.008}{0.011} & \hlc{-0.005}{0.011} & \hlc{-0.009}{0.011} & \hlc{-0.007}{0.011} & \hlc{-0.006}{0.011} & \hlc{0.009}{0.013} & \hlc{0.005}{0.013} & \hlc{-0.013}{0.013} & \hlc{0.000}{0.013} & \hlc{0.005}{0.013} & \hlc{-0.014}{0.015} & \hlc{-0.012}{0.015} & \hlc{-0.008}{0.015} & \hlc{-0.003}{0.015} & \hlc{-0.007}{0.015} \\
 & Fair input & \hlc{-0.004}{0.011} & \hlc{0.003}{0.011} & \hlc{0.000}{0.011} & \hlc{-0.006}{0.011} & \hlc{-0.003}{0.011} & \hlc{0.007}{0.013} & \hlc{0.004}{0.013} & \hlc{0.013}{0.013} & \hlc{0.008}{0.013} & \hlc{0.004}{0.013} & \hlc{-0.014}{0.015} & \hlc{-0.011}{0.015} & \hlc{-0.011}{0.015} & \hlc{-0.006}{0.015} & \hlc{-0.010}{0.015} \\
 & Mixed input & \hlc{-0.002}{0.011} & \hlc{0.000}{0.011} & \hlc{0.000}{0.011} & \hlc{-0.003}{0.011} & \hlc{-0.006}{0.011} & \hlc{0.007}{0.013} & \hlc{0.009}{0.013} & \hlc{0.005}{0.013} & \hlc{0.011}{0.013} & \hlc{0.003}{0.013} & \hlc{-0.010}{0.015} & \hlc{-0.010}{0.015} & \hlc{-0.008}{0.015} & \hlc{-0.009}{0.015} & \hlc{-0.010}{0.015} \\
 & Biased output & \hlc{-0.006}{0.011} & \hlc{-0.004}{0.011} & \hlc{-0.006}{0.011} & \hlc{-0.005}{0.011} & \hlc{-0.003}{0.011} & \hlc{0.006}{0.013} & \hlc{0.008}{0.013} & \hlc{0.009}{0.013} & \hlc{0.015}{0.013} & \hlc{0.015}{0.013} & \hlc{-0.007}{0.015} & \hlc{-0.010}{0.015} & \hlc{-0.011}{0.015} & \hlc{-0.008}{0.015} & \hlc{-0.012}{0.015} \\
 & Fair output & \hlc{-0.009}{0.011} & \hlc{-0.004}{0.011} & \hlc{-0.004}{0.011} & \hlc{0.000}{0.011} & \hlc{-0.002}{0.011} & \hlc{0.013}{0.013} & \hlc{0.003}{0.013} & \hlc{0.010}{0.013} & \hlc{0.002}{0.013} & \hlc{0.006}{0.013} & \hlc{-0.011}{0.015} & \hlc{-0.011}{0.015} & \hlc{-0.009}{0.015} & \hlc{-0.011}{0.015} & \hlc{-0.012}{0.015} \\
 & Mixed output & \hlc{-0.008}{0.011} & \hlc{0.001}{0.011} & \hlc{-0.006}{0.011} & \hlc{-0.004}{0.011} & \hlc{0.003}{0.011} & \hlc{0.010}{0.013} & \hlc{0.007}{0.013} & \hlc{0.007}{0.013} & \hlc{0.002}{0.013} & \hlc{0.006}{0.013} & \hlc{-0.009}{0.015} & \hlc{-0.010}{0.015} & \hlc{-0.010}{0.015} & \hlc{-0.009}{0.015} & \hlc{-0.011}{0.015} \\
\hline
\multirow{7}{*}{20\%} & Left only & \hlc{-0.005}{0.011} & \hlc{-0.004}{0.011} & \hlc{-0.004}{0.011} & \hlc{-0.007}{0.011} & \hlc{0.004}{0.011} & \hlc{0.011}{0.013} & \hlc{-0.002}{0.013} & \hlc{0.007}{0.013} & \hlc{-0.001}{0.013} & \hlc{0.015}{0.013} & \hlc{-0.007}{0.015} & \hlc{-0.012}{0.015} & \hlc{-0.007}{0.015} & \hlc{-0.012}{0.015} & \hlc{-0.003}{0.015} \\
 & Right only & \hlc{0.000}{0.011} & \hlc{-0.001}{0.011} & \hlc{-0.004}{0.011} & \hlc{-0.006}{0.011} & \hlc{-0.004}{0.011} & \hlc{0.013}{0.013} & \hlc{0.007}{0.013} & \hlc{0.001}{0.013} & \hlc{0.003}{0.013} & \hlc{0.010}{0.013} & \hlc{-0.012}{0.015} & \hlc{-0.013}{0.015} & \hlc{-0.010}{0.015} & \hlc{-0.006}{0.015} & \hlc{0.003}{0.015} \\
 & Fair input & \hlc{0.000}{0.011} & \hlc{-0.003}{0.011} & \hlc{-0.005}{0.011} & \hlc{-0.001}{0.011} & \hlc{-0.004}{0.011} & \hlc{0.014}{0.013} & \hlc{0.001}{0.013} & \hlc{0.007}{0.013} & \hlc{0.004}{0.013} & \hlc{0.021}{0.013} & \hlc{-0.006}{0.015} & \hlc{-0.011}{0.015} & \hlc{-0.010}{0.015} & \hlc{-0.012}{0.015} & \hlc{-0.002}{0.015} \\
 & Mixed input & \hlc{-0.006}{0.011} & \hlc{0.000}{0.011} & \hlc{0.000}{0.011} & \hlc{0.001}{0.011} & \hlc{-0.006}{0.011} & \hlc{0.012}{0.013} & \hlc{0.003}{0.013} & \hlc{-0.002}{0.013} & \hlc{0.003}{0.013} & \hlc{0.016}{0.013} & \hlc{-0.007}{0.015} & \hlc{-0.008}{0.015} & \hlc{-0.004}{0.015} & \hlc{-0.011}{0.015} & \hlc{-0.008}{0.015} \\
 & Biased output & \hlc{-0.002}{0.011} & \hlc{-0.006}{0.011} & \hlc{0.003}{0.011} & \hlc{0.002}{0.011} & \hlc{0.002}{0.011} & \hlc{0.017}{0.013} & \hlc{0.004}{0.013} & \hlc{0.006}{0.013} & \hlc{0.001}{0.013} & \hlc{0.005}{0.013} & \hlc{-0.009}{0.015} & \hlc{-0.013}{0.015} & \hlc{-0.011}{0.015} & \hlc{-0.013}{0.015} & \hlc{0.006}{0.015} \\
 & Fair output & \hlc{-0.001}{0.011} & \hlc{0.009}{0.011} & \hlc{0.000}{0.011} & \hlc{-0.008}{0.011} & \hlc{0.005}{0.011} & \hlc{0.013}{0.013} & \hlc{0.000}{0.013} & \hlc{0.009}{0.013} & \hlc{0.002}{0.013} & \hlc{0.009}{0.013} & \hlc{-0.010}{0.015} & \hlc{-0.009}{0.015} & \hlc{-0.012}{0.015} & \hlc{-0.010}{0.015} & \hlc{0.005}{0.015} \\
 & Mixed output & \hlc{-0.002}{0.011} & \hlc{0.001}{0.011} & \hlc{-0.003}{0.011} & \hlc{-0.003}{0.011} & \hlc{-0.006}{0.011} & \hlc{0.009}{0.013} & \hlc{0.009}{0.013} & \hlc{0.012}{0.013} & \hlc{0.002}{0.013} & \hlc{0.015}{0.013} & \hlc{-0.012}{0.015} & \hlc{-0.009}{0.015} & \hlc{-0.008}{0.015} & \hlc{-0.013}{0.015} & \hlc{0.001}{0.015} \\
\hline
\multirow{7}{*}{30\%} & Left only & \hlc{-0.009}{0.011} & \hlc{-0.006}{0.011} & \hlc{0.000}{0.011} & \hlc{-0.002}{0.011} & \hlc{-0.002}{0.011} & \hlc{0.003}{0.013} & \hlc{0.007}{0.013} & \hlc{-0.002}{0.013} & \hlc{0.003}{0.013} & \hlc{0.009}{0.013} & \hlc{0.003}{0.015} & \hlc{-0.011}{0.015} & \hlc{-0.005}{0.015} & \hlc{-0.007}{0.015} & \hlc{-0.007}{0.015} \\
& Right only & \hlc{-0.008}{0.011} & \hlc{-0.003}{0.011} & \hlc{-0.003}{0.011} & \hlc{0.000}{0.011} & \hlc{-0.005}{0.011} & \hlc{0.005}{0.013} & \hlc{0.007}{0.013} & \hlc{0.005}{0.013} & \hlc{0.004}{0.013} & \hlc{0.004}{0.013} & \hlc{-0.009}{0.015} & \hlc{-0.010}{0.015} & \hlc{0.006}{0.015} & \hlc{-0.003}{0.015} & \hlc{0.004}{0.015} \\
 & Fair input & \hlc{-0.006}{0.011} & \hlc{0.000}{0.011} & \hlc{0.001}{0.011} & \hlc{-0.002}{0.011} & \hlc{-0.006}{0.011} & \hlc{0.010}{0.013} & \hlc{0.006}{0.013} & \hlc{0.000}{0.013} & \hlc{0.004}{0.013} & \hlc{-0.006}{0.013} & \hlc{-0.005}{0.015} & \hlc{-0.008}{0.015} & \hlc{-0.012}{0.015} & \hlc{-0.005}{0.015} & \hlc{0.008}{0.015} \\
 & Mixed input & \hlc{-0.005}{0.011} & \hlc{-0.004}{0.011} & \hlc{-0.003}{0.011} & \hlc{-0.004}{0.011} & \hlc{-0.000}{0.011} & \hlc{0.011}{0.013} & \hlc{0.004}{0.013} & \hlc{0.007}{0.013} & \hlc{-0.003}{0.013} & \hlc{0.010}{0.013} & \hlc{-0.008}{0.015} & \hlc{-0.003}{0.015} & \hlc{-0.006}{0.015} & \hlc{-0.011}{0.015} & \hlc{0.002}{0.015} \\
 & Biased output & \hlc{-0.008}{0.011} & \hlc{0.001}{0.011} & \hlc{-0.002}{0.011} & \hlc{-0.001}{0.011} & \hlc{-0.002}{0.011} & \hlc{0.018}{0.013} & \hlc{0.011}{0.013} & \hlc{-0.005}{0.013} & \hlc{0.006}{0.013} & \hlc{0.009}{0.013} & \hlc{-0.011}{0.015} & \hlc{-0.012}{0.015} & \hlc{0.003}{0.015} & \hlc{-0.001}{0.015} & \hlc{0.002}{0.015} \\
 & Fair output & \hlc{-0.006}{0.011} & \hlc{-0.002}{0.011} & \hlc{-0.005}{0.011} & \hlc{0.000}{0.011} & \hlc{-0.001}{0.011} & \hlc{0.013}{0.013} & \hlc{0.008}{0.013} & \hlc{0.003}{0.013} & \hlc{-0.002}{0.013} & \hlc{0.018}{0.013} & \hlc{-0.006}{0.015} & \hlc{-0.003}{0.015} & \hlc{-0.006}{0.015} & \hlc{-0.004}{0.015} & \hlc{0.001}{0.015} \\
 & Mixed output & \hlc{-0.007}{0.011} & \hlc{-0.004}{0.011} & \hlc{-0.001}{0.011} & \hlc{-0.002}{0.011} & \hlc{-0.004}{0.011} & \hlc{0.002}{0.013} & \hlc{0.008}{0.013} & \hlc{-0.004}{0.013} & \hlc{-0.002}{0.013} & \hlc{0.010}{0.013} & \hlc{-0.010}{0.015} & \hlc{-0.010}{0.015} & \hlc{-0.009}{0.015} & \hlc{-0.007}{0.015} & \hlc{-0.005}{0.015} \\
\hline
\multirow{7}{*}{40\%} & Left only & \hlc{-0.005}{0.011} & \hlc{-0.001}{0.011} & \hlc{-0.001}{0.011} & \hlc{-0.007}{0.011} & \hlc{0.001}{0.011} & \hlc{0.013}{0.013} & \hlc{0.004}{0.013} & \hlc{0.013}{0.013} & \hlc{0.008}{0.013} & \hlc{0.017}{0.013} & \hlc{-0.006}{0.015} & \hlc{-0.009}{0.015} & \hlc{-0.007}{0.015} & \hlc{-0.007}{0.015} & \hlc{-0.004}{0.015} \\
 & Right only & \hlc{-0.007}{0.011} & \hlc{0.003}{0.011} & \hlc{0.000}{0.011} & \hlc{0.000}{0.011} & \hlc{0.003}{0.011} & \hlc{0.003}{0.013} & \hlc{-0.003}{0.013} & \hlc{0.005}{0.013} & \hlc{0.008}{0.013} & \hlc{0.015}{0.013} & \hlc{-0.011}{0.015} & \hlc{-0.003}{0.015} & \hlc{-0.007}{0.015} & \hlc{-0.010}{0.015} & \hlc{0.010}{0.015} \\
 & Fair input & \hlc{-0.006}{0.011} & \hlc{0.003}{0.011} & \hlc{-0.001}{0.011} & \hlc{-0.008}{0.011} & \hlc{-0.007}{0.011} & \hlc{0.020}{0.013} & \hlc{0.001}{0.013} & \hlc{-0.002}{0.013} & \hlc{0.006}{0.013} & \hlc{-0.010}{0.013} & \hlc{-0.008}{0.015} & \hlc{-0.005}{0.015} & \hlc{-0.004}{0.015} & \hlc{-0.006}{0.015} & \hlc{-0.006}{0.015} \\
 & Mixed input & \hlc{-0.005}{0.011} & \hlc{0.001}{0.011} & \hlc{-0.008}{0.011} & \hlc{0.004}{0.011} & \hlc{0.012}{0.011} & \hlc{0.006}{0.013} & \hlc{0.006}{0.013} & \hlc{0.016}{0.013} & \hlc{0.004}{0.013} & \hlc{0.007}{0.013} & \hlc{-0.011}{0.015} & \hlc{-0.006}{0.015} & \hlc{-0.006}{0.015} & \hlc{0.002}{0.015} & \hlc{-0.012}{0.015} \\
 & Biased output & \hlc{-0.007}{0.011} & \hlc{-0.001}{0.011} & \hlc{-0.002}{0.011} & \hlc{-0.004}{0.011} & \hlc{0.011}{0.011} & \hlc{0.018}{0.013} & \hlc{0.004}{0.013} & \hlc{0.013}{0.013} & \hlc{0.003}{0.013} & \hlc{0.015}{0.013} & \hlc{-0.004}{0.015} & \hlc{-0.009}{0.015} & \hlc{-0.005}{0.015} & \hlc{-0.008}{0.015} & \hlc{0.002}{0.015} \\
 & Fair output & \hlc{0.001}{0.011} & \hlc{-0.008}{0.011} & \hlc{-0.003}{0.011} & \hlc{-0.003}{0.011} & \hlc{-0.001}{0.011} & \hlc{0.020}{0.013} & \hlc{-0.004}{0.013} & \hlc{0.017}{0.013} & \hlc{0.005}{0.013} & \hlc{0.005}{0.013} & \hlc{-0.007}{0.015} & \hlc{-0.008}{0.015} & \hlc{-0.013}{0.015} & \hlc{-0.009}{0.015} & \hlc{0.004}{0.015} \\
 & Mixed output & \hlc{-0.002}{0.011} & \hlc{-0.001}{0.011} & \hlc{-0.005}{0.011} & \hlc{0.001}{0.011} & \hlc{-0.001}{0.011} & \hlc{0.010}{0.013} & \hlc{0.008}{0.013} & \hlc{0.000}{0.013} & \hlc{-0.006}{0.013} & \hlc{-0.004}{0.013} & \hlc{-0.011}{0.015} & \hlc{-0.009}{0.015} & \hlc{-0.003}{0.015} & \hlc{-0.008}{0.015} & \hlc{-0.007}{0.015} \\
\bottomrule
\end{tabular}
\caption{Political tweet summarisation}
\end{subtable}
\bigskip

\begin{subtable}{\textwidth}
\begin{tabular}{c c r r r r r r r r r r r r r r r}
\toprule
\multicolumn{2}{c}{} & \multicolumn{5}{c}{Llama3-8B (0.496)} & \multicolumn{5}{c}{Gemma-2B (0.785)} & \multicolumn{5}{c}{TinyLlama (0.382)} \\
\cmidrule(lr){3-7} \cmidrule(lr){8-12} \cmidrule(lr){13-17}
Sparsity & Calibration & \multicolumn{1}{c}{Sparse} & \multicolumn{1}{c}{} & \multicolumn{1}{c}{GBLM} & \multicolumn{1}{c}{GBLM} & \multicolumn{1}{c}{} & \multicolumn{1}{c}{Sparse} & \multicolumn{1}{c}{} & \multicolumn{1}{c}{GBLM} & \multicolumn{1}{c}{GBLM} & \multicolumn{1}{c}{} & \multicolumn{1}{c}{Sparse} & \multicolumn{1}{c}{} & \multicolumn{1}{c}{GBLM} & \multicolumn{1}{c}{GBLM} & \multicolumn{1}{c}{} \\
Ratio & Sets & \multicolumn{1}{c}{GPT} & \multicolumn{1}{c}{Wanda} & \multicolumn{1}{c}{Pruner} & \multicolumn{1}{c}{Gradient} & \multicolumn{1}{c}{HGLA} & \multicolumn{1}{c}{GPT} & \multicolumn{1}{c}{Wanda} & \multicolumn{1}{c}{Pruner} & \multicolumn{1}{c}{Gradient} & \multicolumn{1}{c}{HGLA} & \multicolumn{1}{c}{GPT} & \multicolumn{1}{c}{Wanda} & \multicolumn{1}{c}{Pruner} & \multicolumn{1}{c}{Gradient} & \multicolumn{1}{c}{HGLA} \\
\hline
\multirow{7}{*}{10\%} 
& Positive only & \hlc{-0.003}{0.005} & \hlc{-0.004}{0.005} & \hlc{-0.002}{0.005} & \hlc{-0.001}{0.005} & \hlc{-0.000}{0.005} & \hlc{0.000}{0.005} & \hlc{0.000}{0.005} & \hlc{-0.001}{0.005} & \hlc{-0.001}{0.005} & \hlc{-0.001}{0.005} & \hlc{0.000}{0.003} & \hlc{-0.001}{0.003} & \hlc{0.001}{0.003} & \hlc{0.002}{0.003} & \hlc{-0.001}{0.003}\\
& Negative only & \hlc{-0.003}{0.005} & \hlc{-0.003}{0.005} & \hlc{0.000}{0.005} & \hlc{-0.003}{0.005} & \hlc{-0.003}{0.005} & \hlc{0.000}{0.005} & \hlc{0.001}{0.005} & \hlc{-0.001}{0.005} & \hlc{-0.002}{0.005} & \hlc{0.001}{0.005} & \hlc{0.001}{0.003} & \hlc{-0.001}{0.003} & \hlc{0.000}{0.003} & \hlc{-0.001}{0.003} & \hlc{-0.002}{0.003}\\
& Fair input & \hlc{-0.003}{0.005} & \hlc{-0.004}{0.005} & \hlc{-0.003}{0.005} & \hlc{0.000}{0.005} & \hlc{-0.003}{0.005} & \hlc{0.001}{0.005} & \hlc{-0.002}{0.005} & \hlc{-0.001}{0.005} & \hlc{0.001}{0.005} & \hlc{-0.001}{0.005} & \hlc{0.001}{0.003} & \hlc{-0.001}{0.003} & \hlc{0.001}{0.003} & \hlc{0.000}{0.003} & \hlc{-0.001}{0.003}\\
& Mixed input & \hlc{-0.004}{0.005} & \hlc{-0.002}{0.005} & \hlc{-0.002}{0.005} & \hlc{-0.002}{0.005} & \hlc{-0.001}{0.005} & \hlc{0.000}{0.005} & \hlc{-0.002}{0.005} & \hlc{-0.002}{0.005} & \hlc{0.000}{0.005} & \hlc{0.001}{0.005} & \hlc{0.001}{0.003} & \hlc{-0.001}{0.003} & \hlc{0.000}{0.003} & \hlc{0.001}{0.003} & \hlc{-0.002}{0.003}\\
& Biased output & \hlc{-0.002}{0.005} & \hlc{0.003}{0.005} & \hlc{-0.001}{0.005} & \hlc{-0.002}{0.005} & \hlc{-0.003}{0.005} & \hlc{0.000}{0.005} & \hlc{-0.002}{0.005} & \hlc{-0.002}{0.005} & \hlc{-0.001}{0.005} & \hlc{-0.001}{0.005} & \hlc{-0.001}{0.003} & \hlc{-0.002}{0.003} & \hlc{0.002}{0.003} & \hlc{0.002}{0.003} & \hlc{-0.002}{0.003}\\
& Fair output & \hlc{-0.001}{0.005} & \hlc{-0.003}{0.005} & \hlc{-0.002}{0.005} & \hlc{-0.002}{0.005} & \hlc{-0.001}{0.005} & \hlc{0.001}{0.005} & \hlc{-0.002}{0.005} & \hlc{-0.002}{0.005} & \hlc{-0.002}{0.005} & \hlc{-0.002}{0.005} & \hlc{0.000}{0.003} & \hlc{-0.001}{0.003} & \hlc{0.000}{0.003} & \hlc{0.000}{0.003} & \hlc{-0.001}{0.003}\\
& Mixed output & \hlc{-0.001}{0.005} & \hlc{0.001}{0.005} & \hlc{-0.001}{0.005} & \hlc{-0.003}{0.005} & \hlc{-0.002}{0.005} & \hlc{0.001}{0.005} & \hlc{-0.001}{0.005} & \hlc{-0.001}{0.005} & \hlc{-0.001}{0.005} & \hlc{0.001}{0.005} & \hlc{0.000}{0.003} & \hlc{-0.001}{0.003} & \hlc{0.001}{0.003} & \hlc{0.004}{0.003} & \hlc{-0.002}{0.003}\\
\hline
\multirow{7}{*}{20\%} 
& Positive only & \hlc{0.001}{0.005} & \hlc{0.002}{0.005} & \hlc{-0.002}{0.005} & \hlc{-0.001}{0.005} & \hlc{-0.003}{0.005} & \hlc{0.000}{0.005} & \hlc{-0.002}{0.005} & \hlc{-0.002}{0.005} & \hlc{0.000}{0.005} & \hlc{0.002}{0.005} & \hlc{-0.001}{0.003} & \hlc{0.001}{0.003} & \hlc{0.000}{0.003} & \hlc{0.000}{0.003} & \hlc{-0.001}{0.003}\\
& Negative only & \hlc{-0.003}{0.005} & \hlc{0.001}{0.005} & \hlc{-0.001}{0.005} & \hlc{-0.001}{0.005} & \hlc{-0.003}{0.005} & \hlc{0.000}{0.005} & \hlc{0.000}{0.005} & \hlc{0.000}{0.005} & \hlc{0.000}{0.005} & \hlc{0.001}{0.005} & \hlc{-0.002}{0.003} & \hlc{0.001}{0.003} & \hlc{0.003}{0.003} & \hlc{-0.001}{0.003} & \hlc{-0.001}{0.003}\\
& Fair input & \hlc{-0.002}{0.005} & \hlc{0.001}{0.005} & \hlc{-0.003}{0.005} & \hlc{0.001}{0.005} & \hlc{-0.003}{0.005} & \hlc{0.000}{0.005} & \hlc{-0.001}{0.005} & \hlc{-0.001}{0.005} & \hlc{-0.002}{0.005} & \hlc{0.002}{0.005} & \hlc{-0.002}{0.003} & \hlc{0.000}{0.003} & \hlc{-0.001}{0.003} & \hlc{0.001}{0.003} & \hlc{-0.002}{0.003}\\
& Mixed input & \hlc{-0.002}{0.005} & \hlc{-0.003}{0.005} & \hlc{-0.002}{0.005} & \hlc{-0.003}{0.005} & \hlc{-0.003}{0.005} & \hlc{0.001}{0.005} & \hlc{-0.002}{0.005} & \hlc{-0.001}{0.005} & \hlc{0.000}{0.005} & \hlc{0.002}{0.005} & \hlc{-0.001}{0.003} & \hlc{-0.001}{0.003} & \hlc{0.001}{0.003} & \hlc{0.002}{0.003} & \hlc{-0.001}{0.003}\\
& Biased output & \hlc{-0.002}{0.005} & \hlc{0.000}{0.005} & \hlc{-0.002}{0.005} & \hlc{-0.001}{0.005} & \hlc{-0.001}{0.005} & \hlc{-0.001}{0.005} & \hlc{-0.001}{0.005} & \hlc{0.001}{0.005} & \hlc{-0.002}{0.005} & \hlc{0.001}{0.005} & \hlc{0.000}{0.003} & \hlc{0.000}{0.003} & \hlc{0.003}{0.003} & \hlc{0.001}{0.003} & \hlc{-0.001}{0.003}\\
& Fair output & \hlc{-0.002}{0.005} & \hlc{-0.003}{0.005} & \hlc{-0.001}{0.005} & \hlc{-0.002}{0.005} & \hlc{-0.003}{0.005} & \hlc{-0.002}{0.005} & \hlc{-0.001}{0.005} & \hlc{-0.001}{0.005} & \hlc{0.000}{0.005} & \hlc{0.001}{0.005} & \hlc{-0.001}{0.003} & \hlc{0.001}{0.003} & \hlc{-0.001}{0.003} & \hlc{0.002}{0.003} & \hlc{-0.002}{0.003}\\
& Mixed output & \hlc{0.003}{0.005} & \hlc{-0.002}{0.005} & \hlc{-0.002}{0.005} & \hlc{-0.002}{0.005} & \hlc{-0.002}{0.005} & \hlc{0.001}{0.005} & \hlc{-0.001}{0.005} & \hlc{0.000}{0.005} & \hlc{0.001}{0.005} & \hlc{-0.000}{0.005} & \hlc{0.000}{0.003} & \hlc{0.001}{0.003} & \hlc{0.000}{0.003} & \hlc{0.001}{0.003} & \hlc{0.000}{0.003}\\
\hline
\multirow{7}{*}{30\%}
& Positive only & \hlc{-0.002}{0.005} & \hlc{-0.002}{0.005} & \hlc{-0.002}{0.005} & \hlc{-0.003}{0.005} & \hlc{-0.003}{0.005} & \hlc{0.000}{0.005} & \hlc{-0.003}{0.005} & \hlc{0.003}{0.005} & \hlc{0.001}{0.005} & \hlc{-0.001}{0.005} & \hlc{0.002}{0.003} & \hlc{-0.001}{0.003} & \hlc{0.001}{0.003} & \hlc{-0.002}{0.003} & \hlc{-0.001}{0.003}\\
& Negative only & \hlc{-0.003}{0.005} & \hlc{-0.003}{0.005} & \hlc{-0.001}{0.005} & \hlc{-0.004}{0.005} & \hlc{0.001}{0.005} & \hlc{0.003}{0.005} & \hlc{0.000}{0.005} & \hlc{0.001}{0.005} & \hlc{-0.001}{0.005} & \hlc{-0.003}{0.005} & \hlc{-0.001}{0.003} & \hlc{-0.002}{0.003} & \hlc{-0.001}{0.003} & \hlc{0.001}{0.003} & \hlc{-0.001}{0.003}\\
& Fair input & \hlc{-0.003}{0.005} & \hlc{-0.002}{0.005} & \hlc{-0.003}{0.005} & \hlc{-0.002}{0.005} & \hlc{-0.003}{0.005} & \hlc{-0.001}{0.005} & \hlc{-0.003}{0.005} & \hlc{-0.001}{0.005} & \hlc{-0.001}{0.005} & \hlc{-0.002}{0.005} & \hlc{-0.001}{0.003} & \hlc{-0.001}{0.003} & \hlc{-0.001}{0.003} & \hlc{0.001}{0.003} & \hlc{-0.001}{0.003}\\
& Mixed input & \hlc{-0.003}{0.005} & \hlc{-0.002}{0.005} & \hlc{-0.003}{0.005} & \hlc{0.000}{0.005} & \hlc{0.002}{0.005} & \hlc{-0.003}{0.005} & \hlc{-0.003}{0.005} & \hlc{0.000}{0.005} & \hlc{-0.003}{0.005} & \hlc{0.003}{0.005} & \hlc{-0.002}{0.003} & \hlc{-0.001}{0.003} & \hlc{0.003}{0.003} & \hlc{-0.002}{0.003} & \hlc{-0.001}{0.003}\\
& Biased output & \hlc{-0.003}{0.005} & \hlc{-0.003}{0.005} & \hlc{-0.003}{0.005} & \hlc{-0.003}{0.005} & \hlc{-0.001}{0.005} & \hlc{0.000}{0.005} & \hlc{-0.002}{0.005} & \hlc{0.001}{0.005} & \hlc{-0.001}{0.005} & \hlc{0.004}{0.005} & \hlc{-0.002}{0.003} & \hlc{-0.001}{0.003} & \hlc{0.001}{0.003} & \hlc{-0.001}{0.003} & \hlc{0.000}{0.003}\\
& Fair output & \hlc{-0.003}{0.005} & \hlc{-0.003}{0.005} & \hlc{0.000}{0.005} & \hlc{-0.003}{0.005} & \hlc{-0.002}{0.005} & \hlc{0.003}{0.005} & \hlc{-0.002}{0.005} & \hlc{0.002}{0.005} & \hlc{-0.001}{0.005} & \hlc{0.000}{0.005} & \hlc{0.000}{0.003} & \hlc{0.000}{0.003} & \hlc{0.001}{0.003} & \hlc{-0.001}{0.003} & \hlc{-0.002}{0.003}\\
& Mixed output & \hlc{0.000}{0.005} & \hlc{-0.003}{0.005} & \hlc{-0.003}{0.005} & \hlc{0.001}{0.005} & \hlc{-0.002}{0.005} & \hlc{-0.001}{0.005} & \hlc{-0.001}{0.005} & \hlc{0.001}{0.005} & \hlc{0.000}{0.005} & \hlc{0.002}{0.005} & \hlc{0.001}{0.003} & \hlc{-0.002}{0.003} & \hlc{0.004}{0.003} & \hlc{0.003}{0.003} & \hlc{-0.000}{0.003}\\
\hline
\multirow{7}{*}{40\%}
& Positive only & \hlc{-0.001}{0.005} & \hlc{-0.003}{0.005} & \hlc{-0.002}{0.005} & \hlc{-0.003}{0.005} & \hlc{0.001}{0.005} & \hlc{0.000}{0.005} & \hlc{-0.002}{0.005} & \hlc{0.002}{0.005} & \hlc{0.000}{0.005} & \hlc{-0.003}{0.005} & \hlc{-0.002}{0.003} & \hlc{-0.002}{0.003} & \hlc{-0.002}{0.003} & \hlc{-0.001}{0.003} & \hlc{-0.001}{0.003}\\
& Negative only & \hlc{0.001}{0.005} & \hlc{-0.003}{0.005} & \hlc{-0.002}{0.005} & \hlc{0.000}{0.005} & \hlc{-0.001}{0.005} & \hlc{-0.002}{0.005} & \hlc{0.002}{0.005} & \hlc{0.005}{0.005} & \hlc{0.001}{0.005} & \hlc{0.001}{0.005} & \hlc{-0.001}{0.003} & \hlc{0.001}{0.003} & \hlc{0.003}{0.003} & \hlc{0.001}{0.003} & \hlc{0.000}{0.003}\\
& Fair input & \hlc{0.000}{0.005} & \hlc{-0.003}{0.005} & \hlc{0.003}{0.005} & \hlc{-0.001}{0.005} & \hlc{-0.003}{0.005} & \hlc{-0.001}{0.005} & \hlc{0.002}{0.005} & \hlc{-0.001}{0.005} & \hlc{0.005}{0.005} & \hlc{-0.001}{0.005} & \hlc{-0.001}{0.003} & \hlc{0.001}{0.003} & \hlc{0.005}{0.003} & \hlc{-0.002}{0.003} & \hlc{-0.001}{0.003}\\
& Mixed input & \hlc{-0.003}{0.005} & \hlc{-0.002}{0.005} & \hlc{0.000}{0.005} & \hlc{0.000}{0.005} & \hlc{-0.000}{0.005} & \hlc{0.001}{0.005} & \hlc{0.002}{0.005} & \hlc{0.001}{0.005} & \hlc{0.001}{0.005} & \hlc{-0.001}{0.005} & \hlc{-0.001}{0.003} & \hlc{-0.002}{0.003} & \hlc{0.000}{0.003} & \hlc{0.004}{0.003} & \hlc{-0.000}{0.003}\\
& Biased output & \hlc{-0.003}{0.005} & \hlc{-0.003}{0.005} & \hlc{-0.001}{0.005} & \hlc{-0.002}{0.005} & \hlc{-0.001}{0.005} & \hlc{0.000}{0.005} & \hlc{-0.002}{0.005} & \hlc{0.000}{0.005} & \hlc{0.002}{0.005} & \hlc{-0.001}{0.005} & \hlc{-0.001}{0.003} & \hlc{-0.002}{0.003} & \hlc{-0.001}{0.003} & \hlc{0.000}{0.003} & \hlc{-0.002}{0.003}\\
& Fair output & \hlc{-0.001}{0.005} & \hlc{-0.003}{0.005} & \hlc{-0.004}{0.005} & \hlc{-0.002}{0.005} & \hlc{-0.002}{0.005} & \hlc{-0.001}{0.005} & \hlc{0.001}{0.005} & \hlc{-0.001}{0.005} & \hlc{0.003}{0.005} & \hlc{0.001}{0.005} & \hlc{0.001}{0.003} & \hlc{-0.001}{0.003} & \hlc{-0.001}{0.003} & \hlc{0.000}{0.003} & \hlc{-0.000}{0.003}\\
& Mixed output & \hlc{0.000}{0.005} & \hlc{-0.003}{0.005} & \hlc{-0.002}{0.005} & \hlc{-0.001}{0.005} & \hlc{-0.003}{0.005} & \hlc{0.000}{0.005} & \hlc{0.002}{0.005} & \hlc{0.000}{0.005} & \hlc{0.000}{0.005} & \hlc{0.002}{0.005} & \hlc{-0.001}{0.003} & \hlc{0.002}{0.003} & \hlc{-0.001}{0.003} & \hlc{0.004}{0.003} & \hlc{-0.001}{0.003}\\

\bottomrule
\end{tabular}
\caption{Review summarisation}
\label{tab:review_various_input_fairness_sof}
\end{subtable}
\caption{Comparison of sparsity ratio, calibration sets, and model fairness using different datasets. The SOF value of the vanilla model is reported in brackets next to the model name.  We report fairness improvement by calculating the absolute difference between the SOF of the vanilla model and that of the pruned model. A model demonstrating a positive impact on fairness should have an absolute difference ranging from 0 to its vanilla SOF, with values closer to the vanilla SOF indicating better improvement (values between 0 and vanilla SOF are highlighted, indicating that the pruned model is less biased than the original model). Darker colours indicate greater improvement in fairness.}

\label{tab:various_input_fairness_sof}
\end{table*}

\begin{table*}[htbp]
\definecolor{veryLightGreen}{rgb}{0.85, 0.95, 0.85}
\definecolor{lightGreen}{rgb}{0.70, 0.85, 0.70}
\definecolor{mediumGreen}{rgb}{0.40, 0.70, 0.40}
\definecolor{mediumDarkGreen}{rgb}{0.20, 0.60, 0.20}
\definecolor{darkGreen}{rgb}{0.07, 0.53, 0.03}

\newcommand{\hlc}[2]{%
  \ifdim#1pt>0pt
    \cellcolor{%
      \ifnum\pdfstrcmp{\fpeval{#1/#2}}{\fpeval{0.2}}<0 veryLightGreen\else
      \ifnum\pdfstrcmp{\fpeval{#1/#2}}{\fpeval{0.4}}<0 lightGreen\else
      \ifnum\pdfstrcmp{\fpeval{#1/#2}}{\fpeval{0.6}}<0 mediumGreen\else
      \ifnum\pdfstrcmp{\fpeval{#1/#2}}{\fpeval{0.8}}<0 mediumDarkGreen\else
      darkGreen\fi\fi\fi\fi
    }%
  \fi
  #1%
}
\centering
\tiny
\setlength{\tabcolsep}{4pt} % Reduce space between columns
\begin{subtable}{\textwidth}
\begin{tabular}{c c r r r r r r r r r r r r r r r}
\toprule
\multicolumn{2}{c}{} & \multicolumn{5}{c}{Llama3-8B (0.496)} & \multicolumn{5}{c}{Gemma-2B (0.785)} & \multicolumn{5}{c}{TinyLlama (0.382)} \\
\cmidrule(lr){3-7} \cmidrule(lr){8-12} \cmidrule(lr){13-17}
Sparsity & Calibration & \multicolumn{1}{c}{Sparse} & \multicolumn{1}{c}{} & \multicolumn{1}{c}{GBLM} & \multicolumn{1}{c}{GBLM} & \multicolumn{1}{c}{} & \multicolumn{1}{c}{Sparse} & \multicolumn{1}{c}{} & \multicolumn{1}{c}{GBLM} & \multicolumn{1}{c}{GBLM} & \multicolumn{1}{c}{} & \multicolumn{1}{c}{Sparse} & \multicolumn{1}{c}{} & \multicolumn{1}{c}{GBLM} & \multicolumn{1}{c}{GBLM} & \multicolumn{1}{c}{} \\
Ratio & Sets & \multicolumn{1}{c}{GPT} & \multicolumn{1}{c}{Wanda} & \multicolumn{1}{c}{Pruner} & \multicolumn{1}{c}{Gradient} & \multicolumn{1}{c}{HGLA} & \multicolumn{1}{c}{GPT} & \multicolumn{1}{c}{Wanda} & \multicolumn{1}{c}{Pruner} & \multicolumn{1}{c}{Gradient} & \multicolumn{1}{c}{HGLA} & \multicolumn{1}{c}{GPT} & \multicolumn{1}{c}{Wanda} & \multicolumn{1}{c}{Pruner} & \multicolumn{1}{c}{Gradient} & \multicolumn{1}{c}{HGLA} \\
\hline
\multirow{7}{*}{10\%} & Left only & \hlc{-0.040}{0.240} & \hlc{-0.113}{0.240} & \hlc{-0.240}{0.240} & \hlc{-0.073}{0.240} & \hlc{-0.033}{0.240} & \hlc{-0.013}{0.393} & \hlc{0.000}{0.393} & \hlc{-0.393}{0.393} & \hlc{-0.040}{0.393} & \hlc{-0.007}{0.393} & \hlc{0.100}{0.253} & \hlc{0.027}{0.253} & \hlc{-0.093}{0.253} & \hlc{0.040}{0.253} & \hlc{0.073}{0.253} \\
 & Right only & \hlc{-0.087}{0.240} & \hlc{-0.100}{0.240} & \hlc{-0.227}{0.240} & \hlc{-0.060}{0.240} & \hlc{-0.107}{0.240} & \hlc{-0.047}{0.393} & \hlc{0.000}{0.393} & \hlc{-0.387}{0.393} & \hlc{0.007}{0.393} & \hlc{0.013}{0.393} & \hlc{0.053}{0.253} & \hlc{0.060}{0.253} & \hlc{-0.207}{0.253} & \hlc{0.033}{0.253} & \hlc{0.040}{0.253} \\
 & Fair input & \hlc{-0.060}{0.240} & \hlc{-0.053}{0.240} & \hlc{-0.053}{0.240} & \hlc{-0.067}{0.240} & \hlc{-0.047}{0.240} & \hlc{-0.047}{0.393} & \hlc{-0.020}{0.393} & \hlc{-0.047}{0.393} & \hlc{-0.013}{0.393} & \hlc{-0.020}{0.393} & \hlc{0.093}{0.253} & \hlc{0.047}{0.253} & \hlc{0.047}{0.253} & \hlc{0.047}{0.253} & \hlc{0.047}{0.253} \\
 & Mixed input & \hlc{-0.087}{0.240} & \hlc{-0.060}{0.240} & \hlc{-0.093}{0.240} & \hlc{-0.053}{0.240} & \hlc{-0.007}{0.240} & \hlc{-0.020}{0.393} & \hlc{-0.027}{0.393} & \hlc{-0.040}{0.393} & \hlc{-0.067}{0.393} & \hlc{-0.007}{0.393} & \hlc{0.053}{0.253} & \hlc{0.047}{0.253} & \hlc{0.020}{0.253} & \hlc{0.033}{0.253} & \hlc{0.067}{0.253} \\
 & Biased output & \hlc{-0.067}{0.240} & \hlc{0.013}{0.240} & \hlc{-0.080}{0.240} & \hlc{-0.040}{0.240} & \hlc{-0.047}{0.240} & \hlc{-0.067}{0.393} & \hlc{-0.007}{0.393} & \hlc{-0.040}{0.393} & \hlc{-0.053}{0.393} & \hlc{0.007}{0.393} & \hlc{0.020}{0.253} & \hlc{0.040}{0.253} & \hlc{0.060}{0.253} & \hlc{0.047}{0.253} & \hlc{0.060}{0.253} \\
 & Fair output & \hlc{-0.060}{0.240} & \hlc{-0.007}{0.240} & \hlc{-0.040}{0.240} & \hlc{-0.067}{0.240} & \hlc{-0.060}{0.240} & \hlc{-0.040}{0.393} & \hlc{-0.013}{0.393} & \hlc{-0.067}{0.393} & \hlc{-0.020}{0.393} & \hlc{0.020}{0.393} & \hlc{0.067}{0.253} & \hlc{0.040}{0.253} & \hlc{0.047}{0.253} & \hlc{0.060}{0.253} & \hlc{0.067}{0.253} \\
 & Mixed output & \hlc{-0.033}{0.240} & \hlc{0.020}{0.240} & \hlc{-0.093}{0.240} & \hlc{-0.060}{0.240} & \hlc{-0.053}{0.240} & \hlc{-0.067}{0.393} & \hlc{-0.013}{0.393} & \hlc{-0.027}{0.393} & \hlc{-0.020}{0.393} & \hlc{-0.013}{0.393} & \hlc{0.073}{0.253} & \hlc{0.033}{0.253} & \hlc{0.073}{0.253} & \hlc{0.013}{0.253} & \hlc{0.087}{0.253} \\
\hline
\multirow{7}{*}{20\%} & Left only & \hlc{-0.073}{0.240} & \hlc{-0.053}{0.240} & \hlc{-0.047}{0.240} & \hlc{-0.067}{0.240} & \hlc{-0.020}{0.240} & \hlc{-0.067}{0.393} & \hlc{0.007}{0.393} & \hlc{-0.027}{0.393} & \hlc{0.000}{0.393} & \hlc{0.007}{0.393} & \hlc{0.107}{0.253} & \hlc{0.013}{0.253} & \hlc{0.073}{0.253} & \hlc{0.053}{0.253} & \hlc{0.073}{0.253} \\
 & Right only & \hlc{-0.053}{0.240} & \hlc{-0.060}{0.240} & \hlc{-0.100}{0.240} & \hlc{-0.087}{0.240} & \hlc{-0.040}{0.240} & \hlc{-0.040}{0.393} & \hlc{0.020}{0.393} & \hlc{-0.060}{0.393} & \hlc{-0.020}{0.393} & \hlc{-0.027}{0.393} & \hlc{0.060}{0.253} & \hlc{0.040}{0.253} & \hlc{0.033}{0.253} & \hlc{0.053}{0.253} & \hlc{0.040}{0.253} \\
 & Fair input & \hlc{-0.027}{0.240} & \hlc{-0.080}{0.240} & \hlc{-0.007}{0.240} & \hlc{-0.080}{0.240} & \hlc{-0.040}{0.240} & \hlc{-0.093}{0.393} & \hlc{-0.013}{0.393} & \hlc{-0.013}{0.393} & \hlc{-0.027}{0.393} & \hlc{-0.033}{0.393} & \hlc{0.093}{0.253} & \hlc{0.013}{0.253} & \hlc{-0.007}{0.253} & \hlc{0.040}{0.253} & \hlc{0.040}{0.253} \\
 & Mixed input & \hlc{-0.060}{0.240} & \hlc{-0.040}{0.240} & \hlc{-0.087}{0.240} & \hlc{-0.040}{0.240} & \hlc{-0.027}{0.240} & \hlc{0.000}{0.393} & \hlc{-0.040}{0.393} & \hlc{-0.027}{0.393} & \hlc{-0.040}{0.393} & \hlc{-0.007}{0.393} & \hlc{0.060}{0.253} & \hlc{0.047}{0.253} & \hlc{-0.033}{0.253} & \hlc{0.027}{0.253} & \hlc{0.047}{0.253} \\
 & Biased output & \hlc{-0.067}{0.240} & \hlc{-0.053}{0.240} & \hlc{-0.033}{0.240} & \hlc{-0.020}{0.240} & \hlc{-0.047}{0.240} & \hlc{-0.087}{0.393} & \hlc{-0.013}{0.393} & \hlc{-0.033}{0.393} & \hlc{0.013}{0.393} & \hlc{0.027}{0.393} & \hlc{0.027}{0.253} & \hlc{0.027}{0.253} & \hlc{0.040}{0.253} & \hlc{0.080}{0.253} & \hlc{0.067}{0.253} \\
 & Fair output & \hlc{-0.033}{0.240} & \hlc{-0.053}{0.240} & \hlc{-0.047}{0.240} & \hlc{-0.053}{0.240} & \hlc{0.007}{0.240} & \hlc{-0.073}{0.393} & \hlc{0.013}{0.393} & \hlc{-0.033}{0.393} & \hlc{-0.020}{0.393} & \hlc{0.020}{0.393} & \hlc{0.080}{0.253} & \hlc{0.020}{0.253} & \hlc{0.073}{0.253} & \hlc{0.053}{0.253} & \hlc{0.080}{0.253} \\
 & Mixed output & \hlc{-0.007}{0.240} & \hlc{-0.020}{0.240} & \hlc{-0.100}{0.240} & \hlc{-0.080}{0.240} & \hlc{-0.047}{0.240} & \hlc{-0.080}{0.393} & \hlc{-0.027}{0.393} & \hlc{-0.013}{0.393} & \hlc{-0.033}{0.393} & \hlc{-0.007}{0.393} & \hlc{0.093}{0.253} & \hlc{-0.007}{0.253} & \hlc{0.047}{0.253} & \hlc{0.007}{0.253} & \hlc{0.053}{0.253} \\
\hline
\multirow{7}{*}{30\%} & Left only & \hlc{-0.120}{0.240} & \hlc{-0.007}{0.240} & \hlc{-0.067}{0.240} & \hlc{-0.033}{0.240} & \hlc{0.007}{0.240} & \hlc{-0.020}{0.393} & \hlc{-0.040}{0.393} & \hlc{-0.073}{0.393} & \hlc{-0.047}{0.393} & \hlc{-0.053}{0.393} & \hlc{0.053}{0.253} & \hlc{0.013}{0.253} & \hlc{0.027}{0.253} & \hlc{-0.007}{0.253} & \hlc{0.067}{0.253} \\
 & Right only & \hlc{-0.067}{0.240} & \hlc{-0.100}{0.240} & \hlc{-0.033}{0.240} & \hlc{-0.060}{0.240} & \hlc{-0.060}{0.240} & \hlc{-0.060}{0.393} & \hlc{0.000}{0.393} & \hlc{-0.020}{0.393} & \hlc{-0.067}{0.393} & \hlc{-0.053}{0.393} & \hlc{0.033}{0.253} & \hlc{0.073}{0.253} & \hlc{0.107}{0.253} & \hlc{0.040}{0.253} & \hlc{0.147}{0.253} \\
 & Fair input & \hlc{-0.060}{0.240} & \hlc{-0.060}{0.240} & \hlc{-0.020}{0.240} & \hlc{-0.027}{0.240} & \hlc{0.053}{0.240} & \hlc{-0.073}{0.393} & \hlc{-0.027}{0.393} & \hlc{0.040}{0.393} & \hlc{-0.053}{0.393} & \hlc{0.000}{0.393} & \hlc{0.027}{0.253} & \hlc{0.013}{0.253} & \hlc{0.020}{0.253} & \hlc{0.033}{0.253} & \hlc{0.107}{0.253} \\
 & Mixed input & \hlc{-0.087}{0.240} & \hlc{-0.087}{0.240} & \hlc{-0.093}{0.240} & \hlc{-0.067}{0.240} & \hlc{-0.033}{0.240} & \hlc{-0.060}{0.393} & \hlc{-0.073}{0.393} & \hlc{-0.053}{0.393} & \hlc{-0.087}{0.393} & \hlc{-0.100}{0.393} & \hlc{0.007}{0.253} & \hlc{0.020}{0.253} & \hlc{0.120}{0.253} & \hlc{0.033}{0.253} & \hlc{0.020}{0.253} \\
 & Biased output & \hlc{-0.060}{0.240} & \hlc{-0.020}{0.240} & \hlc{-0.053}{0.240} & \hlc{-0.040}{0.240} & \hlc{-0.007}{0.240} & \hlc{-0.067}{0.393} & \hlc{-0.040}{0.393} & \hlc{-0.013}{0.393} & \hlc{-0.013}{0.393} & \hlc{-0.047}{0.393} & \hlc{-0.020}{0.253} & \hlc{0.067}{0.253} & \hlc{0.107}{0.253} & \hlc{0.020}{0.253} & \hlc{0.080}{0.253} \\
& Fair output & \hlc{-0.073}{0.240} & \hlc{-0.067}{0.240} & \hlc{-0.067}{0.240} & \hlc{-0.080}{0.240} & \hlc{-0.060}{0.240} & \hlc{-0.027}{0.393} & \hlc{-0.093}{0.393} & \hlc{-0.053}{0.393} & \hlc{-0.060}{0.393} & \hlc{-0.040}{0.393} & \hlc{0.033}{0.253} & \hlc{0.087}{0.253} & \hlc{0.073}{0.253} & \hlc{0.047}{0.253} & \hlc{0.047}{0.253} \\
 & Mixed output & \hlc{-0.113}{0.240} & \hlc{-0.027}{0.240} & \hlc{0.020}{0.240} & \hlc{-0.073}{0.240} & \hlc{0.033}{0.240} & \hlc{-0.087}{0.393} & \hlc{-0.067}{0.393} & \hlc{-0.060}{0.393} & \hlc{-0.053}{0.393} & \hlc{-0.120}{0.393} & \hlc{0.013}{0.253} & \hlc{0.060}{0.253} & \hlc{0.027}{0.253} & \hlc{-0.013}{0.253} & \hlc{0.087}{0.253} \\
\hline
\multirow{7}{*}{40\%} & Left only & \hlc{-0.020}{0.240} & \hlc{-0.020}{0.240} & \hlc{-0.067}{0.240} & \hlc{-0.080}{0.240} & \hlc{0.100}{0.240} & \hlc{-0.087}{0.393} & \hlc{0.013}{0.393} & \hlc{0.033}{0.393} & \hlc{-0.040}{0.393} & \hlc{-0.007}{0.393} & \hlc{0.060}{0.253} & \hlc{0.020}{0.253} & \hlc{0.027}{0.253} & \hlc{-0.020}{0.253} & \hlc{0.087}{0.253} \\
 & Right only & \hlc{-0.047}{0.240} & \hlc{0.040}{0.240} & \hlc{0.053}{0.240} & \hlc{0.100}{0.240} & \hlc{0.100}{0.240} & \hlc{-0.060}{0.393} & \hlc{-0.027}{0.393} & \hlc{-0.080}{0.393} & \hlc{0.007}{0.393} & \hlc{0.047}{0.393} & \hlc{0.053}{0.253} & \hlc{0.053}{0.253} & \hlc{0.080}{0.253} & \hlc{0.053}{0.253} & \hlc{0.067}{0.253} \\
 & Fair input & \hlc{-0.007}{0.240} & \hlc{0.007}{0.240} & \hlc{0.020}{0.240} & \hlc{-0.047}{0.240} & \hlc{0.093}{0.240} & \hlc{-0.047}{0.393} & \hlc{0.007}{0.393} & \hlc{0.047}{0.393} & \hlc{0.033}{0.393} & \hlc{-0.033}{0.393} & \hlc{-0.007}{0.253} & \hlc{0.040}{0.253} & \hlc{0.007}{0.253} & \hlc{-0.007}{0.253} & \hlc{0.147}{0.253} \\
 & Mixed input & \hlc{-0.053}{0.240} & \hlc{0.000}{0.240} & \hlc{0.000}{0.240} & \hlc{-0.060}{0.240} & \hlc{0.080}{0.240} & \hlc{-0.047}{0.393} & \hlc{0.027}{0.393} & \hlc{-0.053}{0.393} & \hlc{-0.080}{0.393} & \hlc{-0.047}{0.393} & \hlc{0.047}{0.253} & \hlc{0.040}{0.253} & \hlc{0.120}{0.253} & \hlc{-0.007}{0.253} & \hlc{0.027}{0.253} \\
 & Biased output & \hlc{-0.040}{0.240} & \hlc{0.013}{0.240} & \hlc{-0.033}{0.240} & \hlc{-0.047}{0.240} & \hlc{0.100}{0.240} & \hlc{-0.040}{0.393} & \hlc{0.013}{0.393} & \hlc{-0.013}{0.393} & \hlc{-0.087}{0.393} & \hlc{0.027}{0.393} & \hlc{0.013}{0.253} & \hlc{-0.013}{0.253} & \hlc{0.020}{0.253} & \hlc{-0.047}{0.253} & \hlc{0.080}{0.253} \\
 & Fair output & \hlc{-0.047}{0.240} & \hlc{-0.013}{0.240} & \hlc{0.047}{0.240} & \hlc{0.053}{0.240} & \hlc{0.127}{0.240} & \hlc{0.000}{0.393} & \hlc{-0.027}{0.393} & \hlc{-0.020}{0.393} & \hlc{-0.007}{0.393} & \hlc{0.027}{0.393} & \hlc{0.053}{0.253} & \hlc{0.040}{0.253} & \hlc{0.053}{0.253} & \hlc{0.047}{0.253} & \hlc{0.040}{0.253} \\
 & Mixed output & \hlc{-0.073}{0.240} & \hlc{0.020}{0.240} & \hlc{-0.027}{0.240} & \hlc{-0.040}{0.240} & \hlc{0.053}{0.240} & \hlc{-0.093}{0.393} & \hlc{0.000}{0.393} & \hlc{-0.020}{0.393} & \hlc{-0.040}{0.393} & \hlc{0.013}{0.393} & \hlc{0.027}{0.253} & \hlc{0.020}{0.253} & \hlc{-0.020}{0.253} & \hlc{-0.047}{0.253} & \hlc{0.013}{0.253} \\
\bottomrule
\end{tabular}
\caption{Political tweet summarisation}
\end{subtable}

\bigskip

\begin{subtable}{\textwidth}
\begin{tabular}{c c r r r r r r r r r r r r r r r}
\toprule
\multicolumn{2}{c}{} & \multicolumn{5}{c}{Llama3-8B (0.496)} & \multicolumn{5}{c}{Gemma-2B (0.785)} & \multicolumn{5}{c}{TinyLlama (0.382)} \\
\cmidrule(lr){3-7} \cmidrule(lr){8-12} \cmidrule(lr){13-17}
Sparsity & Calibration & \multicolumn{1}{c}{Sparse} & \multicolumn{1}{c}{} & \multicolumn{1}{c}{GBLM} & \multicolumn{1}{c}{GBLM} & \multicolumn{1}{c}{} & \multicolumn{1}{c}{Sparse} & \multicolumn{1}{c}{} & \multicolumn{1}{c}{GBLM} & \multicolumn{1}{c}{GBLM} & \multicolumn{1}{c}{} & \multicolumn{1}{c}{Sparse} & \multicolumn{1}{c}{} & \multicolumn{1}{c}{GBLM} & \multicolumn{1}{c}{GBLM} & \multicolumn{1}{c}{} \\
Ratio & Sets & \multicolumn{1}{c}{GPT} & \multicolumn{1}{c}{Wanda} & \multicolumn{1}{c}{Pruner} & \multicolumn{1}{c}{Gradient} & \multicolumn{1}{c}{HGLA} & \multicolumn{1}{c}{GPT} & \multicolumn{1}{c}{Wanda} & \multicolumn{1}{c}{Pruner} & \multicolumn{1}{c}{Gradient} & \multicolumn{1}{c}{HGLA} & \multicolumn{1}{c}{GPT} & \multicolumn{1}{c}{Wanda} & \multicolumn{1}{c}{Pruner} & \multicolumn{1}{c}{Gradient} & \multicolumn{1}{c}{HGLA} \\
\hline
\multirow{7}{*}{10\%} 
& Positive only & \hlc{0.004}{0.387} & \hlc{0.000}{0.387} & \hlc{-0.243}{0.387} & \hlc{0.019}{0.387} & \hlc{-0.008}{0.387} & \hlc{-0.004}{0.297} & \hlc{0.021}{0.297} & \hlc{-0.152}{0.297} & \hlc{0.002}{0.297} & \hlc{0.019}{0.297} & \hlc{-0.024}{0.413} & \hlc{-0.010}{0.413} & \hlc{-0.006}{0.413} & \hlc{0.007}{0.413} & \hlc{0.018}{0.413}\\
& Negative only & \hlc{0.012}{0.387} & \hlc{0.003}{0.387} & \hlc{-0.221}{0.387} & \hlc{0.024}{0.387} & \hlc{0.010}{0.387} & \hlc{-0.012}{0.297} & \hlc{0.020}{0.297} & \hlc{-0.150}{0.297} & \hlc{0.010}{0.297} & \hlc{0.003}{0.297} & \hlc{-0.014}{0.413} & \hlc{0.003}{0.413} & \hlc{-0.006}{0.413} & \hlc{-0.013}{0.413} & \hlc{0.022}{0.413}\\
& Fair input & \hlc{0.029}{0.387} & \hlc{0.002}{0.387} & \hlc{-0.002}{0.387} & \hlc{0.017}{0.387} & \hlc{0.026}{0.387} & \hlc{-0.012}{0.297} & \hlc{-0.019}{0.297} & \hlc{-0.013}{0.297} & \hlc{-0.022}{0.297} & \hlc{0.009}{0.297} & \hlc{-0.032}{0.413} & \hlc{-0.017}{0.413} & \hlc{-0.010}{0.413} & \hlc{-0.013}{0.413} & \hlc{0.014}{0.413}\\
& Mixed input & \hlc{0.020}{0.387} & \hlc{0.027}{0.387} & \hlc{-0.007}{0.387} & \hlc{0.024}{0.387} & \hlc{-0.004}{0.387} & \hlc{-0.011}{0.297} & \hlc{-0.024}{0.297} & \hlc{-0.006}{0.297} & \hlc{-0.008}{0.297} & \hlc{0.010}{0.297} & \hlc{-0.010}{0.413} & \hlc{0.008}{0.413} & \hlc{-0.013}{0.413} & \hlc{-0.018}{0.413} & \hlc{0.013}{0.413}\\
& Biased output & \hlc{0.028}{0.387} & \hlc{0.010}{0.387} & \hlc{0.017}{0.387} & \hlc{0.007}{0.387} & \hlc{0.006}{0.387} & \hlc{-0.011}{0.297} & \hlc{-0.024}{0.297} & \hlc{-0.010}{0.297} & \hlc{-0.008}{0.297} & \hlc{0.001}{0.297} & \hlc{-0.022}{0.413} & \hlc{-0.007}{0.413} & \hlc{-0.016}{0.413} & \hlc{-0.011}{0.413} & \hlc{0.016}{0.413}\\
& Fair output & \hlc{0.027}{0.387} & \hlc{-0.008}{0.387} & \hlc{0.021}{0.387} & \hlc{0.011}{0.387} & \hlc{0.024}{0.387} & \hlc{-0.006}{0.297} & \hlc{-0.018}{0.297} & \hlc{-0.013}{0.297} & \hlc{-0.017}{0.297} & \hlc{0.000}{0.297} & \hlc{-0.023}{0.413} & \hlc{0.001}{0.413} & \hlc{0.011}{0.413} & \hlc{-0.018}{0.413} & \hlc{0.008}{0.413}\\
& Mixed output & \hlc{0.024}{0.387} & \hlc{0.040}{0.387} & \hlc{0.003}{0.387} & \hlc{0.007}{0.387} & \hlc{0.008}{0.387} & \hlc{-0.002}{0.297} & \hlc{-0.012}{0.297} & \hlc{0.018}{0.297} & \hlc{-0.011}{0.297} & \hlc{0.012}{0.297} & \hlc{-0.027}{0.413} & \hlc{-0.001}{0.413} & \hlc{-0.007}{0.413} & \hlc{-0.036}{0.413} & \hlc{0.020}{0.413}\\
\hline
\multirow{7}{*}{20\%} 
& Positive only & \hlc{0.059}{0.387} & \hlc{0.044}{0.387} & \hlc{0.006}{0.387} & \hlc{-0.004}{0.387} & \hlc{0.034}{0.387} & \hlc{-0.007}{0.297} & \hlc{-0.016}{0.297} & \hlc{-0.011}{0.297} & \hlc{0.009}{0.297} & \hlc{0.016}{0.297} & \hlc{0.038}{0.413} & \hlc{-0.021}{0.413} & \hlc{-0.023}{0.413} & \hlc{-0.003}{0.413} & \hlc{0.032}{0.413}\\
& Negative only & \hlc{0.031}{0.387} & \hlc{0.017}{0.387} & \hlc{-0.016}{0.387} & \hlc{0.012}{0.387} & \hlc{0.017}{0.387} & \hlc{-0.007}{0.297} & \hlc{-0.017}{0.297} & \hlc{0.000}{0.297} & \hlc{0.014}{0.297} & \hlc{0.007}{0.297} & \hlc{0.050}{0.413} & \hlc{-0.006}{0.413} & \hlc{-0.026}{0.413} & \hlc{-0.003}{0.413} & \hlc{-0.004}{0.413}\\
& Fair input & \hlc{0.041}{0.387} & \hlc{0.048}{0.387} & \hlc{0.019}{0.387} & \hlc{0.009}{0.387} & \hlc{-0.006}{0.387} & \hlc{-0.019}{0.297} & \hlc{-0.009}{0.297} & \hlc{-0.009}{0.297} & \hlc{-0.011}{0.297} & \hlc{-0.008}{0.297} & \hlc{0.050}{0.413} & \hlc{-0.001}{0.413} & \hlc{-0.022}{0.413} & \hlc{-0.011}{0.413} & \hlc{0.017}{0.413}\\
& Mixed input & \hlc{0.013}{0.387} & \hlc{0.053}{0.387} & \hlc{0.002}{0.387} & \hlc{-0.019}{0.387} & \hlc{0.013}{0.387} & \hlc{0.014}{0.297} & \hlc{-0.001}{0.297} & \hlc{-0.019}{0.297} & \hlc{-0.010}{0.297} & \hlc{0.017}{0.297} & \hlc{0.048}{0.413} & \hlc{-0.009}{0.413} & \hlc{0.001}{0.413} & \hlc{-0.007}{0.413} & \hlc{-0.008}{0.413}\\
& Biased output & \hlc{0.042}{0.387} & \hlc{0.018}{0.387} & \hlc{0.007}{0.387} & \hlc{0.002}{0.387} & \hlc{0.010}{0.387} & \hlc{-0.010}{0.297} & \hlc{-0.003}{0.297} & \hlc{-0.026}{0.297} & \hlc{-0.013}{0.297} & \hlc{0.002}{0.297} & \hlc{0.047}{0.413} & \hlc{-0.018}{0.413} & \hlc{-0.030}{0.413} & \hlc{-0.012}{0.413} & \hlc{0.019}{0.413}\\
& Fair output & \hlc{0.030}{0.387} & \hlc{0.018}{0.387} & \hlc{0.018}{0.387} & \hlc{-0.016}{0.387} & \hlc{0.014}{0.387} & \hlc{0.006}{0.297} & \hlc{-0.014}{0.297} & \hlc{-0.010}{0.297} & \hlc{-0.009}{0.297} & \hlc{0.009}{0.297} & \hlc{0.059}{0.413} & \hlc{-0.002}{0.413} & \hlc{0.014}{0.413} & \hlc{-0.002}{0.413} & \hlc{0.018}{0.413}\\
& Mixed output & \hlc{0.016}{0.387} & \hlc{0.029}{0.387} & \hlc{0.003}{0.387} & \hlc{-0.002}{0.387} & \hlc{0.027}{0.387} & \hlc{0.011}{0.297} & \hlc{-0.007}{0.297} & \hlc{0.013}{0.297} & \hlc{0.008}{0.297} & \hlc{0.031}{0.297} & \hlc{0.066}{0.413} & \hlc{-0.016}{0.413} & \hlc{-0.083}{0.413} & \hlc{-0.082}{0.413} & \hlc{-0.007}{0.413}\\
\hline
\multirow{7}{*}{30\%}
& Positive only & \hlc{0.013}{0.387} & \hlc{-0.039}{0.387} & \hlc{-0.036}{0.387} & \hlc{-0.050}{0.387} & \hlc{0.016}{0.387} & \hlc{-0.009}{0.297} & \hlc{-0.017}{0.297} & \hlc{-0.003}{0.297} & \hlc{-0.002}{0.297} & \hlc{0.004}{0.297} & \hlc{-0.004}{0.413} & \hlc{0.014}{0.413} & \hlc{0.024}{0.413} & \hlc{0.001}{0.413} & \hlc{0.039}{0.413}\\
& Negative only & \hlc{0.021}{0.387} & \hlc{-0.023}{0.387} & \hlc{-0.010}{0.387} & \hlc{-0.051}{0.387} & \hlc{0.031}{0.387} & \hlc{-0.009}{0.297} & \hlc{-0.007}{0.297} & \hlc{-0.007}{0.297} & \hlc{-0.026}{0.297} & \hlc{0.021}{0.297} & \hlc{0.007}{0.413} & \hlc{0.039}{0.413} & \hlc{-0.001}{0.413} & \hlc{-0.003}{0.413} & \hlc{0.058}{0.413}\\
& Fair input & \hlc{0.027}{0.387} & \hlc{-0.019}{0.387} & \hlc{-0.034}{0.387} & \hlc{-0.048}{0.387} & \hlc{0.023}{0.387} & \hlc{-0.016}{0.297} & \hlc{-0.019}{0.297} & \hlc{0.004}{0.297} & \hlc{-0.026}{0.297} & \hlc{0.011}{0.297} & \hlc{0.026}{0.413} & \hlc{0.030}{0.413} & \hlc{0.006}{0.413} & \hlc{-0.022}{0.413} & \hlc{0.067}{0.413}\\
& Mixed input & \hlc{0.020}{0.387} & \hlc{-0.011}{0.387} & \hlc{-0.006}{0.387} & \hlc{-0.057}{0.387} & \hlc{-0.010}{0.387} & \hlc{-0.001}{0.297} & \hlc{-0.012}{0.297} & \hlc{0.016}{0.297} & \hlc{0.000}{0.297} & \hlc{0.012}{0.297} & \hlc{0.057}{0.413} & \hlc{0.057}{0.413} & \hlc{-0.012}{0.413} & \hlc{-0.089}{0.413} & \hlc{0.040}{0.413}\\
& Biased output & \hlc{0.041}{0.387} & \hlc{-0.033}{0.387} & \hlc{-0.012}{0.387} & \hlc{-0.028}{0.387} & \hlc{0.010}{0.387} & \hlc{-0.011}{0.297} & \hlc{-0.024}{0.297} & \hlc{0.014}{0.297} & \hlc{-0.006}{0.297} & \hlc{0.013}{0.297} & \hlc{0.049}{0.413} & \hlc{0.019}{0.413} & \hlc{0.002}{0.413} & \hlc{0.008}{0.413} & \hlc{0.029}{0.413}\\
& Fair output & \hlc{0.043}{0.387} & \hlc{-0.023}{0.387} & \hlc{-0.046}{0.387} & \hlc{-0.068}{0.387} & \hlc{-0.016}{0.387} & \hlc{0.010}{0.297} & \hlc{-0.028}{0.297} & \hlc{-0.002}{0.297} & \hlc{-0.013}{0.297} & \hlc{0.037}{0.297} & \hlc{0.076}{0.413} & \hlc{0.037}{0.413} & \hlc{0.024}{0.413} & \hlc{-0.020}{0.413} & \hlc{0.051}{0.413}\\
& Mixed output & \hlc{-0.007}{0.387} & \hlc{-0.011}{0.387} & \hlc{-0.048}{0.387} & \hlc{-0.053}{0.387} & \hlc{0.006}{0.387} & \hlc{0.013}{0.297} & \hlc{-0.012}{0.297} & \hlc{0.006}{0.297} & \hlc{-0.030}{0.297} & \hlc{-0.001}{0.297} & \hlc{0.041}{0.413} & \hlc{0.048}{0.413} & \hlc{-0.104}{0.413} & \hlc{-0.122}{0.413} & \hlc{0.031}{0.413}\\
\hline
\multirow{7}{*}{40\%}
& Positive only & \hlc{0.042}{0.387} & \hlc{-0.019}{0.387} & \hlc{-0.054}{0.387} & \hlc{-0.068}{0.387} & \hlc{0.010}{0.387} & \hlc{-0.022}{0.297} & \hlc{0.008}{0.297} & \hlc{0.034}{0.297} & \hlc{-0.022}{0.297} & \hlc{0.051}{0.297} & \hlc{0.048}{0.413} & \hlc{0.010}{0.413} & \hlc{-0.050}{0.413} & \hlc{-0.009}{0.413} & \hlc{-0.013}{0.413}\\
& Negative only & \hlc{0.013}{0.387} & \hlc{-0.013}{0.387} & \hlc{-0.048}{0.387} & \hlc{-0.089}{0.387} & \hlc{-0.018}{0.387} & \hlc{0.034}{0.297} & \hlc{-0.016}{0.297} & \hlc{0.016}{0.297} & \hlc{0.007}{0.297} & \hlc{0.033}{0.297} & \hlc{0.063}{0.413} & \hlc{-0.018}{0.413} & \hlc{-0.046}{0.413} & \hlc{-0.066}{0.413} & \hlc{-0.013}{0.413}\\
& Fair input & \hlc{-0.023}{0.387} & \hlc{-0.032}{0.387} & \hlc{-0.059}{0.387} & \hlc{-0.042}{0.387} & \hlc{-0.000}{0.387} & \hlc{-0.010}{0.297} & \hlc{-0.018}{0.297} & \hlc{0.002}{0.297} & \hlc{-0.026}{0.297} & \hlc{0.020}{0.297} & \hlc{0.056}{0.413} & \hlc{0.010}{0.413} & \hlc{0.002}{0.413} & \hlc{-0.071}{0.413} & \hlc{-0.008}{0.413}\\
& Mixed input & \hlc{-0.019}{0.387} & \hlc{-0.050}{0.387} & \hlc{-0.058}{0.387} & \hlc{-0.097}{0.387} & \hlc{-0.056}{0.387} & \hlc{0.003}{0.297} & \hlc{0.006}{0.297} & \hlc{-0.004}{0.297} & \hlc{-0.016}{0.297} & \hlc{0.049}{0.297} & \hlc{0.080}{0.413} & \hlc{-0.009}{0.413} & \hlc{-0.093}{0.413} & \hlc{-0.107}{0.413} & \hlc{-0.028}{0.413}\\
& Biased output & \hlc{0.041}{0.387} & \hlc{-0.013}{0.387} & \hlc{-0.097}{0.387} & \hlc{-0.034}{0.387} & \hlc{0.002}{0.387} & \hlc{-0.010}{0.297} & \hlc{0.000}{0.297} & \hlc{-0.001}{0.297} & \hlc{-0.001}{0.297} & \hlc{0.008}{0.297} & \hlc{0.039}{0.413} & \hlc{-0.017}{0.413} & \hlc{-0.052}{0.413} & \hlc{-0.068}{0.413} & \hlc{-0.060}{0.413}\\
& Fair output & \hlc{0.028}{0.387} & \hlc{-0.016}{0.387} & \hlc{-0.057}{0.387} & \hlc{-0.061}{0.387} & \hlc{-0.014}{0.387} & \hlc{-0.017}{0.297} & \hlc{-0.024}{0.297} & \hlc{0.039}{0.297} & \hlc{-0.021}{0.297} & \hlc{0.030}{0.297} & \hlc{0.087}{0.413} & \hlc{0.018}{0.413} & \hlc{-0.021}{0.413} & \hlc{0.004}{0.413} & \hlc{-0.041}{0.413}\\
& Mixed output & \hlc{0.027}{0.387} & \hlc{-0.031}{0.387} & \hlc{-0.070}{0.387} & \hlc{-0.106}{0.387} & \hlc{-0.054}{0.387} & \hlc{0.017}{0.297} & \hlc{-0.028}{0.297} & \hlc{-0.006}{0.297} & \hlc{-0.012}{0.297} & \hlc{0.032}{0.297} & \hlc{0.062}{0.413} & \hlc{0.006}{0.413} & \hlc{-0.116}{0.413} & \hlc{-0.146}{0.413} & \hlc{-0.037}{0.413}\\
\bottomrule
\end{tabular}
\caption{Review summarisation}
\end{subtable}
\caption{Comparison of sparsity ratio, calibration sets, and model fairness using different datasets. The BUR value of the vanilla model is reported in brackets next to the model name.  We report fairness improvement by calculating the absolute difference between the BUR of the vanilla model and that of the pruned model. A model demonstrating a positive impact on fairness should have an absolute difference ranging from 0 to its vanilla BUR, with values closer to the vanilla BUR indicating better improvement (values between 0 and vanilla BUR are highlighted, indicating that the pruned model is less biased than the original model). Darker colours indicate greater improvement in fairness.}

\label{tab:various_input_fairness_bur}
\end{table*}

\begin{table*}[htbp]
\definecolor{veryLightGreen}{rgb}{0.85, 0.95, 0.85}
\definecolor{lightGreen}{rgb}{0.70, 0.85, 0.70}
\definecolor{mediumGreen}{rgb}{0.40, 0.70, 0.40}
\definecolor{mediumDarkGreen}{rgb}{0.20, 0.60, 0.20}
\definecolor{darkGreen}{rgb}{0.07, 0.53, 0.03}

\newcommand{\hlc}[2]{%
  \ifdim#1pt>0pt
    \cellcolor{%
      \ifnum\pdfstrcmp{\fpeval{#1/#2}}{\fpeval{0.2}}<0 veryLightGreen\else
      \ifnum\pdfstrcmp{\fpeval{#1/#2}}{\fpeval{0.4}}<0 lightGreen\else
      \ifnum\pdfstrcmp{\fpeval{#1/#2}}{\fpeval{0.6}}<0 mediumGreen\else
      \ifnum\pdfstrcmp{\fpeval{#1/#2}}{\fpeval{0.8}}<0 mediumDarkGreen\else
      darkGreen\fi\fi\fi\fi
    }%
  \fi
  #1%
}
\centering
\tiny
\setlength{\tabcolsep}{4pt} % Reduce space between columns
\begin{subtable}{\textwidth}
\begin{tabular}{c c r r r r r r r r r r r r r r r}
\toprule
\multicolumn{2}{c}{} & \multicolumn{5}{c}{Llama3-8B (0.496)} & \multicolumn{5}{c}{Gemma-2B (0.785)} & \multicolumn{5}{c}{TinyLlama (0.382)} \\
\cmidrule(lr){3-7} \cmidrule(lr){8-12} \cmidrule(lr){13-17}
Sparsity & Calibration & \multicolumn{1}{c}{Sparse} & \multicolumn{1}{c}{} & \multicolumn{1}{c}{GBLM} & \multicolumn{1}{c}{GBLM} & \multicolumn{1}{c}{} & \multicolumn{1}{c}{Sparse} & \multicolumn{1}{c}{} & \multicolumn{1}{c}{GBLM} & \multicolumn{1}{c}{GBLM} & \multicolumn{1}{c}{} & \multicolumn{1}{c}{Sparse} & \multicolumn{1}{c}{} & \multicolumn{1}{c}{GBLM} & \multicolumn{1}{c}{GBLM} & \multicolumn{1}{c}{} \\
Ratio & Sets & \multicolumn{1}{c}{GPT} & \multicolumn{1}{c}{Wanda} & \multicolumn{1}{c}{Pruner} & \multicolumn{1}{c}{Gradient} & \multicolumn{1}{c}{HGLA} & \multicolumn{1}{c}{GPT} & \multicolumn{1}{c}{Wanda} & \multicolumn{1}{c}{Pruner} & \multicolumn{1}{c}{Gradient} & \multicolumn{1}{c}{HGLA} & \multicolumn{1}{c}{GPT} & \multicolumn{1}{c}{Wanda} & \multicolumn{1}{c}{Pruner} & \multicolumn{1}{c}{Gradient} & \multicolumn{1}{c}{HGLA} \\
\hline
\multirow{7}{*}{10\%} & Left only & \hlc{-0.003}{0.036} & \hlc{-0.008}{0.036} & \hlc{-0.021}{0.036} & \hlc{-0.003}{0.036} & \hlc{-0.001}{0.036} & \hlc{0.002}{0.058} & \hlc{0.001}{0.058} & \hlc{-0.043}{0.058} & \hlc{-0.003}{0.058} & \hlc{0.001}{0.058} & \hlc{0.016}{0.037} & \hlc{0.009}{0.037} & \hlc{0.008}{0.037} & \hlc{0.014}{0.037} & \hlc{0.015}{0.037} \\
 & Right only & \hlc{-0.006}{0.036} & \hlc{-0.005}{0.036} & \hlc{-0.018}{0.036} & \hlc{0.001}{0.036} & \hlc{-0.006}{0.036} & \hlc{-0.003}{0.058} & \hlc{0.000}{0.058} & \hlc{-0.042}{0.058} & \hlc{0.000}{0.058} & \hlc{0.002}{0.058} & \hlc{0.014}{0.037} & \hlc{0.013}{0.037} & \hlc{-0.015}{0.037} & \hlc{0.011}{0.037} & \hlc{0.011}{0.037} \\
 & Fair input & \hlc{-0.003}{0.036} & \hlc{0.000}{0.036} & \hlc{-0.002}{0.036} & \hlc{-0.004}{0.036} & \hlc{0.000}{0.036} & \hlc{-0.003}{0.058} & \hlc{0.002}{0.058} & \hlc{-0.003}{0.058} & \hlc{0.001}{0.058} & \hlc{-0.000}{0.058} & \hlc{0.014}{0.037} & \hlc{0.011}{0.037} & \hlc{0.012}{0.037} & \hlc{0.011}{0.037} & \hlc{0.010}{0.037} \\
 & Mixed input & \hlc{-0.008}{0.036} & \hlc{-0.004}{0.036} & \hlc{-0.007}{0.036} & \hlc{-0.005}{0.036} & \hlc{0.005}{0.036} & \hlc{0.001}{0.058} & \hlc{0.001}{0.058} & \hlc{-0.004}{0.058} & \hlc{-0.004}{0.058} & \hlc{0.000}{0.058} & \hlc{0.012}{0.037} & \hlc{0.012}{0.037} & \hlc{0.008}{0.037} & \hlc{0.011}{0.037} & \hlc{0.014}{0.037} \\
 & Biased output & \hlc{-0.005}{0.036} & \hlc{0.006}{0.036} & \hlc{-0.001}{0.036} & \hlc{-0.006}{0.036} & \hlc{-0.002}{0.036} & \hlc{-0.004}{0.058} & \hlc{0.000}{0.058} & \hlc{-0.002}{0.058} & \hlc{-0.002}{0.058} & \hlc{-0.001}{0.058} & \hlc{0.009}{0.037} & \hlc{0.011}{0.037} & \hlc{0.013}{0.037} & \hlc{0.010}{0.037} & \hlc{0.012}{0.037} \\
 & Fair output & \hlc{-0.007}{0.036} & \hlc{-0.001}{0.036} & \hlc{-0.001}{0.036} & \hlc{-0.001}{0.036} & \hlc{-0.003}{0.036} & \hlc{-0.002}{0.058} & \hlc{0.001}{0.058} & \hlc{-0.005}{0.058} & \hlc{-0.002}{0.058} & \hlc{0.004}{0.058} & \hlc{0.015}{0.037} & \hlc{0.010}{0.037} & \hlc{0.012}{0.037} & \hlc{0.012}{0.037} & \hlc{0.013}{0.037} \\
 & Mixed output & \hlc{-0.003}{0.036} & \hlc{0.000}{0.036} & \hlc{-0.004}{0.036} & \hlc{-0.001}{0.036} & \hlc{-0.002}{0.036} & \hlc{-0.004}{0.058} & \hlc{0.000}{0.058} & \hlc{-0.001}{0.058} & \hlc{0.000}{0.058} & \hlc{0.003}{0.058} & \hlc{0.015}{0.037} & \hlc{0.010}{0.037} & \hlc{0.011}{0.037} & \hlc{0.006}{0.037} & \hlc{0.015}{0.037} \\
\hline
\multirow{7}{*}{20\%} & Left only & \hlc{-0.005}{0.036} & \hlc{-0.002}{0.036} & \hlc{-0.003}{0.036} & \hlc{-0.003}{0.036} & \hlc{0.002}{0.036} & \hlc{-0.006}{0.058} & \hlc{0.003}{0.058} & \hlc{-0.001}{0.058} & \hlc{0.000}{0.058} & \hlc{0.000}{0.058} & \hlc{0.015}{0.037} & \hlc{0.005}{0.037} & \hlc{0.013}{0.037} & \hlc{0.011}{0.037} & \hlc{0.015}{0.037} \\
 & Right only & \hlc{0.000}{0.036} & \hlc{-0.004}{0.036} & \hlc{-0.004}{0.036} & \hlc{-0.005}{0.036} & \hlc{0.000}{0.036} & \hlc{-0.004}{0.058} & \hlc{0.001}{0.058} & \hlc{-0.003}{0.058} & \hlc{-0.002}{0.058} & \hlc{-0.002}{0.058} & \hlc{0.009}{0.037} & \hlc{0.006}{0.037} & \hlc{0.011}{0.037} & \hlc{0.009}{0.037} & \hlc{0.010}{0.037} \\
 & Fair input & \hlc{0.000}{0.036} & \hlc{-0.002}{0.036} & \hlc{-0.001}{0.036} & \hlc{-0.006}{0.036} & \hlc{-0.000}{0.036} & \hlc{-0.007}{0.058} & \hlc{0.000}{0.058} & \hlc{0.000}{0.058} & \hlc{-0.006}{0.058} & \hlc{-0.002}{0.058} & \hlc{0.011}{0.037} & \hlc{0.008}{0.037} & \hlc{0.002}{0.037} & \hlc{0.007}{0.037} & \hlc{0.008}{0.037} \\
 & Mixed input & \hlc{-0.003}{0.036} & \hlc{0.000}{0.036} & \hlc{-0.005}{0.036} & \hlc{-0.001}{0.036} & \hlc{0.002}{0.036} & \hlc{0.001}{0.058} & \hlc{-0.002}{0.058} & \hlc{-0.003}{0.058} & \hlc{0.000}{0.058} & \hlc{-0.002}{0.058} & \hlc{0.012}{0.037} & \hlc{0.015}{0.037} & \hlc{0.004}{0.037} & \hlc{0.005}{0.037} & \hlc{0.009}{0.037} \\
 & Biased output & \hlc{-0.003}{0.036} & \hlc{0.003}{0.036} & \hlc{0.000}{0.036} & \hlc{0.000}{0.036} & \hlc{0.001}{0.036} & \hlc{-0.007}{0.058} & \hlc{0.000}{0.058} & \hlc{-0.004}{0.058} & \hlc{0.004}{0.058} & \hlc{0.002}{0.058} & \hlc{0.007}{0.037} & \hlc{0.008}{0.037} & \hlc{0.008}{0.037} & \hlc{0.016}{0.037} & \hlc{0.015}{0.037} \\
 & Fair output & \hlc{0.001}{0.036} & \hlc{0.001}{0.036} & \hlc{-0.002}{0.036} & \hlc{-0.002}{0.036} & \hlc{0.005}{0.036} & \hlc{-0.008}{0.058} & \hlc{0.006}{0.058} & \hlc{-0.003}{0.058} & \hlc{-0.003}{0.058} & \hlc{0.000}{0.058} & \hlc{0.015}{0.037} & \hlc{0.006}{0.037} & \hlc{0.010}{0.037} & \hlc{0.009}{0.037} & \hlc{0.013}{0.037} \\
 & Mixed output & \hlc{0.006}{0.036} & \hlc{0.000}{0.036} & \hlc{-0.010}{0.036} & \hlc{-0.004}{0.036} & \hlc{-0.000}{0.036} & \hlc{-0.009}{0.058} & \hlc{0.001}{0.058} & \hlc{0.003}{0.058} & \hlc{0.000}{0.058} & \hlc{-0.003}{0.058} & \hlc{0.013}{0.037} & \hlc{0.003}{0.037} & \hlc{0.011}{0.037} & \hlc{0.005}{0.037} & \hlc{0.012}{0.037} \\
\hline
\multirow{7}{*}{30\%} & Left only & \hlc{-0.007}{0.036} & \hlc{0.001}{0.036} & \hlc{-0.002}{0.036} & \hlc{0.000}{0.036} & \hlc{0.002}{0.036} & \hlc{-0.004}{0.058} & \hlc{-0.004}{0.058} & \hlc{-0.008}{0.058} & \hlc{-0.004}{0.058} & \hlc{-0.008}{0.058} & \hlc{0.015}{0.037} & \hlc{0.006}{0.037} & \hlc{0.008}{0.037} & \hlc{0.003}{0.037} & \hlc{0.010}{0.037} \\
 & Right only & \hlc{-0.007}{0.036} & \hlc{-0.005}{0.036} & \hlc{-0.002}{0.036} & \hlc{0.000}{0.036} & \hlc{-0.001}{0.036} & \hlc{-0.009}{0.058} & \hlc{0.002}{0.058} & \hlc{-0.003}{0.058} & \hlc{-0.008}{0.058} & \hlc{-0.013}{0.058} & \hlc{0.007}{0.037} & \hlc{0.008}{0.037} & \hlc{0.016}{0.037} & \hlc{0.012}{0.037} & \hlc{0.024}{0.037} \\
 & Fair input & \hlc{-0.004}{0.036} & \hlc{-0.001}{0.036} & \hlc{0.005}{0.036} & \hlc{0.002}{0.036} & \hlc{0.009}{0.036} & \hlc{-0.007}{0.058} & \hlc{-0.001}{0.058} & \hlc{0.012}{0.058} & \hlc{-0.004}{0.058} & \hlc{0.001}{0.058} & \hlc{0.008}{0.037} & \hlc{0.003}{0.037} & \hlc{0.005}{0.037} & \hlc{0.008}{0.037} & \hlc{0.019}{0.037} \\
 & Mixed input & \hlc{-0.004}{0.036} & \hlc{-0.008}{0.036} & \hlc{-0.006}{0.036} & \hlc{0.002}{0.036} & \hlc{0.002}{0.036} & \hlc{-0.009}{0.058} & \hlc{-0.007}{0.058} & \hlc{-0.008}{0.058} & \hlc{-0.007}{0.058} & \hlc{-0.017}{0.058} & \hlc{0.004}{0.037} & \hlc{0.008}{0.037} & \hlc{0.017}{0.037} & \hlc{0.007}{0.037} & \hlc{0.010}{0.037} \\
 & Biased output & \hlc{-0.005}{0.036} & \hlc{-0.001}{0.036} & \hlc{-0.001}{0.036} & \hlc{-0.001}{0.036} & \hlc{0.005}{0.036} & \hlc{-0.008}{0.058} & \hlc{-0.004}{0.058} & \hlc{-0.003}{0.058} & \hlc{0.000}{0.058} & \hlc{-0.011}{0.058} & \hlc{0.002}{0.037} & \hlc{0.011}{0.037} & \hlc{0.017}{0.037} & \hlc{0.006}{0.037} & \hlc{0.015}{0.037} \\
 & Fair output & \hlc{-0.005}{0.036} & \hlc{-0.007}{0.036} & \hlc{-0.003}{0.036} & \hlc{-0.002}{0.036} & \hlc{-0.004}{0.036} & \hlc{-0.004}{0.058} & \hlc{-0.010}{0.058} & \hlc{-0.006}{0.058} & \hlc{-0.008}{0.058} & \hlc{-0.010}{0.058} & \hlc{0.009}{0.037} & \hlc{0.013}{0.037} & \hlc{0.011}{0.037} & \hlc{0.009}{0.037} & \hlc{0.013}{0.037} \\
 & Mixed output & \hlc{-0.005}{0.036} & \hlc{-0.001}{0.036} & \hlc{0.004}{0.036} & \hlc{-0.003}{0.036} & \hlc{0.005}{0.036} & \hlc{-0.009}{0.058} & \hlc{-0.006}{0.058} & \hlc{-0.004}{0.058} & \hlc{-0.005}{0.058} & \hlc{-0.016}{0.058} & \hlc{0.007}{0.037} & \hlc{0.010}{0.037} & \hlc{0.010}{0.037} & \hlc{0.003}{0.037} & \hlc{0.018}{0.037} \\
\hline
\multirow{7}{*}{40\%} & Left only & \hlc{-0.002}{0.036} & \hlc{0.003}{0.036} & \hlc{0.000}{0.036} & \hlc{-0.006}{0.036} & \hlc{0.019}{0.036} & \hlc{-0.007}{0.058} & \hlc{0.005}{0.058} & \hlc{0.004}{0.058} & \hlc{-0.001}{0.058} & \hlc{0.002}{0.058} & \hlc{0.012}{0.037} & \hlc{0.007}{0.037} & \hlc{0.009}{0.037} & \hlc{0.005}{0.037} & \hlc{0.010}{0.037} \\
 & Right only & \hlc{-0.004}{0.036} & \hlc{0.008}{0.036} & \hlc{0.009}{0.036} & \hlc{0.017}{0.036} & \hlc{0.015}{0.036} & \hlc{-0.002}{0.058} & \hlc{-0.006}{0.058} & \hlc{-0.005}{0.058} & \hlc{0.000}{0.058} & \hlc{0.004}{0.058} & \hlc{0.014}{0.037} & \hlc{0.015}{0.037} & \hlc{0.013}{0.037} & \hlc{0.011}{0.037} & \hlc{0.011}{0.037} \\
 & Fair input & \hlc{0.007}{0.036} & \hlc{0.004}{0.036} & \hlc{0.010}{0.036} & \hlc{0.010}{0.036} & \hlc{0.017}{0.036} & \hlc{-0.005}{0.058} & \hlc{0.000}{0.058} & \hlc{0.008}{0.058} & \hlc{0.007}{0.058} & \hlc{0.001}{0.058} & \hlc{0.008}{0.037} & \hlc{0.011}{0.037} & \hlc{0.010}{0.037} & \hlc{0.005}{0.037} & \hlc{0.029}{0.037} \\
 & Mixed input & \hlc{-0.006}{0.036} & \hlc{0.007}{0.036} & \hlc{0.000}{0.036} & \hlc{0.004}{0.036} & \hlc{0.016}{0.036} & \hlc{-0.005}{0.058} & \hlc{0.004}{0.058} & \hlc{-0.007}{0.058} & \hlc{-0.011}{0.058} & \hlc{-0.005}{0.058} & \hlc{0.010}{0.037} & \hlc{0.011}{0.037} & \hlc{0.025}{0.037} & \hlc{0.008}{0.037} & \hlc{0.012}{0.037} \\
 & Biased output & \hlc{-0.001}{0.036} & \hlc{0.005}{0.036} & \hlc{-0.002}{0.036} & \hlc{-0.001}{0.036} & \hlc{0.018}{0.036} & \hlc{-0.004}{0.058} & \hlc{0.004}{0.058} & \hlc{0.000}{0.058} & \hlc{-0.006}{0.058} & \hlc{-0.001}{0.058} & \hlc{0.008}{0.037} & \hlc{0.005}{0.037} & \hlc{0.011}{0.037} & \hlc{0.000}{0.037} & \hlc{0.014}{0.037} \\
 & Fair output & \hlc{-0.002}{0.036} & \hlc{0.010}{0.036} & \hlc{0.008}{0.036} & \hlc{0.009}{0.036} & \hlc{0.023}{0.036} & \hlc{0.001}{0.058} & \hlc{-0.003}{0.058} & \hlc{-0.002}{0.058} & \hlc{-0.003}{0.058} & \hlc{0.005}{0.058} & \hlc{0.012}{0.037} & \hlc{0.013}{0.037} & \hlc{0.008}{0.037} & \hlc{0.009}{0.037} & \hlc{0.012}{0.037} \\
 & Mixed output & \hlc{-0.004}{0.036} & \hlc{0.007}{0.036} & \hlc{0.009}{0.036} & \hlc{0.005}{0.036} & \hlc{0.011}{0.036} & \hlc{-0.007}{0.058} & \hlc{0.002}{0.058} & \hlc{0.002}{0.058} & \hlc{-0.001}{0.058} & \hlc{-0.002}{0.058} & \hlc{0.010}{0.037} & \hlc{0.009}{0.037} & \hlc{0.008}{0.037} & \hlc{0.003}{0.037} & \hlc{0.011}{0.037} \\
\bottomrule
\end{tabular}
\caption{Political tweet summarisation}
\end{subtable}

\bigskip

\begin{subtable}{\textwidth}
\begin{tabular}{c c r r r r r r r r r r r r r r r}
\toprule
\multicolumn{2}{c}{} & \multicolumn{5}{c}{Llama3-8B (0.496)} & \multicolumn{5}{c}{Gemma-2B (0.785)} & \multicolumn{5}{c}{TinyLlama (0.382)} \\
\cmidrule(lr){3-7} \cmidrule(lr){8-12} \cmidrule(lr){13-17}
Sparsity & Calibration & \multicolumn{1}{c}{Sparse} & \multicolumn{1}{c}{} & \multicolumn{1}{c}{GBLM} & \multicolumn{1}{c}{GBLM} & \multicolumn{1}{c}{} & \multicolumn{1}{c}{Sparse} & \multicolumn{1}{c}{} & \multicolumn{1}{c}{GBLM} & \multicolumn{1}{c}{GBLM} & \multicolumn{1}{c}{} & \multicolumn{1}{c}{Sparse} & \multicolumn{1}{c}{} & \multicolumn{1}{c}{GBLM} & \multicolumn{1}{c}{GBLM} & \multicolumn{1}{c}{} \\
Ratio & Sets & \multicolumn{1}{c}{GPT} & \multicolumn{1}{c}{Wanda} & \multicolumn{1}{c}{Pruner} & \multicolumn{1}{c}{Gradient} & \multicolumn{1}{c}{HGLA} & \multicolumn{1}{c}{GPT} & \multicolumn{1}{c}{Wanda} & \multicolumn{1}{c}{Pruner} & \multicolumn{1}{c}{Gradient} & \multicolumn{1}{c}{HGLA} & \multicolumn{1}{c}{GPT} & \multicolumn{1}{c}{Wanda} & \multicolumn{1}{c}{Pruner} & \multicolumn{1}{c}{Gradient} & \multicolumn{1}{c}{HGLA} \\
\hline
\multirow{7}{*}{10\%} 
& Positive only & \hlc{0.001}{0.062} & \hlc{-0.001}{0.062} & \hlc{-0.025}{0.062} & \hlc{0.000}{0.062} & \hlc{-0.003}{0.062} & \hlc{0.000}{0.048} & \hlc{0.002}{0.048} & \hlc{-0.013}{0.048} & \hlc{0.000}{0.048} & \hlc{0.001}{0.048} & \hlc{-0.003}{0.059} & \hlc{-0.001}{0.059} & \hlc{-0.002}{0.059} & \hlc{-0.002}{0.059} & \hlc{0.003}{0.059}\\
& Negative only & \hlc{0.001}{0.062} & \hlc{-0.001}{0.062} & \hlc{-0.023}{0.062} & \hlc{0.000}{0.062} & \hlc{-0.001}{0.062} & \hlc{-0.001}{0.048} & \hlc{0.001}{0.048} & \hlc{-0.012}{0.048} & \hlc{0.001}{0.048} & \hlc{0.000}{0.048} & \hlc{-0.002}{0.059} & \hlc{0.000}{0.059} & \hlc{-0.002}{0.059} & \hlc{-0.003}{0.059} & \hlc{0.003}{0.059}\\
& Fair input & \hlc{0.003}{0.062} & \hlc{-0.002}{0.062} & \hlc{-0.001}{0.062} & \hlc{-0.002}{0.062} & \hlc{-0.002}{0.062} & \hlc{-0.002}{0.048} & \hlc{-0.001}{0.048} & \hlc{-0.002}{0.048} & \hlc{-0.002}{0.048} & \hlc{-0.000}{0.048} & \hlc{-0.003}{0.059} & \hlc{-0.002}{0.059} & \hlc{-0.003}{0.059} & \hlc{-0.002}{0.059} & \hlc{0.003}{0.059}\\
& Mixed input & \hlc{0.002}{0.062} & \hlc{0.002}{0.062} & \hlc{-0.001}{0.062} & \hlc{0.001}{0.062} & \hlc{-0.003}{0.062} & \hlc{0.000}{0.048} & \hlc{-0.002}{0.048} & \hlc{0.000}{0.048} & \hlc{-0.002}{0.048} & \hlc{0.000}{0.048} & \hlc{-0.002}{0.059} & \hlc{0.001}{0.059} & \hlc{-0.003}{0.059} & \hlc{-0.002}{0.059} & \hlc{0.003}{0.059}\\
& Biased output & \hlc{0.001}{0.062} & \hlc{-0.003}{0.062} & \hlc{-0.001}{0.062} & \hlc{-0.001}{0.062} & \hlc{-0.001}{0.062} & \hlc{-0.001}{0.048} & \hlc{-0.001}{0.048} & \hlc{-0.001}{0.048} & \hlc{-0.001}{0.048} & \hlc{0.000}{0.048} & \hlc{-0.003}{0.059} & \hlc{-0.001}{0.059} & \hlc{-0.003}{0.059} & \hlc{-0.002}{0.059} & \hlc{0.004}{0.059}\\
& Fair output & \hlc{0.001}{0.062} & \hlc{-0.002}{0.062} & \hlc{0.001}{0.062} & \hlc{0.001}{0.062} & \hlc{0.001}{0.062} & \hlc{-0.002}{0.048} & \hlc{-0.001}{0.048} & \hlc{-0.002}{0.048} & \hlc{-0.001}{0.048} & \hlc{-0.001}{0.048} & \hlc{-0.003}{0.059} & \hlc{0.000}{0.059} & \hlc{-0.001}{0.059} & \hlc{-0.003}{0.059} & \hlc{0.002}{0.059}\\
& Mixed output & \hlc{0.002}{0.062} & \hlc{0.002}{0.062} & \hlc{0.000}{0.062} & \hlc{0.000}{0.062} & \hlc{-0.001}{0.062} & \hlc{-0.001}{0.048} & \hlc{-0.001}{0.048} & \hlc{0.001}{0.048} & \hlc{-0.001}{0.048} & \hlc{0.001}{0.048} & \hlc{-0.003}{0.059} & \hlc{0.000}{0.059} & \hlc{-0.002}{0.059} & \hlc{-0.005}{0.059} & \hlc{0.003}{0.059}\\
\hline
\multirow{7}{*}{20\%} 
& Positive only & \hlc{0.002}{0.062} & \hlc{0.001}{0.062} & \hlc{-0.001}{0.062} & \hlc{-0.002}{0.062} & \hlc{0.004}{0.062} & \hlc{-0.001}{0.048} & \hlc{-0.001}{0.048} & \hlc{-0.001}{0.048} & \hlc{-0.001}{0.048} & \hlc{0.000}{0.048} & \hlc{0.005}{0.059} & \hlc{-0.002}{0.059} & \hlc{-0.002}{0.059} & \hlc{-0.001}{0.059} & \hlc{0.002}{0.059}\\
& Negative only & \hlc{0.001}{0.062} & \hlc{-0.001}{0.062} & \hlc{-0.001}{0.062} & \hlc{-0.001}{0.062} & \hlc{0.000}{0.062} & \hlc{0.000}{0.048} & \hlc{-0.001}{0.048} & \hlc{0.000}{0.048} & \hlc{0.001}{0.048} & \hlc{0.000}{0.048} & \hlc{0.005}{0.059} & \hlc{-0.002}{0.059} & \hlc{-0.003}{0.059} & \hlc{0.001}{0.059} & \hlc{0.000}{0.059}\\
& Fair input & \hlc{0.001}{0.062} & \hlc{-0.001}{0.062} & \hlc{0.001}{0.062} & \hlc{-0.003}{0.062} & \hlc{-0.001}{0.062} & \hlc{-0.001}{0.048} & \hlc{0.000}{0.048} & \hlc{-0.001}{0.048} & \hlc{-0.002}{0.048} & \hlc{-0.000}{0.048} & \hlc{0.006}{0.059} & \hlc{0.000}{0.059} & \hlc{-0.003}{0.059} & \hlc{-0.001}{0.059} & \hlc{0.002}{0.059}\\
& Mixed input & \hlc{0.002}{0.062} & \hlc{0.003}{0.062} & \hlc{0.000}{0.062} & \hlc{-0.004}{0.062} & \hlc{-0.000}{0.062} & \hlc{0.001}{0.048} & \hlc{0.000}{0.048} & \hlc{-0.002}{0.048} & \hlc{-0.001}{0.048} & \hlc{0.002}{0.048} & \hlc{0.006}{0.059} & \hlc{0.000}{0.059} & \hlc{-0.002}{0.059} & \hlc{-0.002}{0.059} & \hlc{-0.001}{0.059}\\
& Biased output & \hlc{0.003}{0.062} & \hlc{0.001}{0.062} & \hlc{0.000}{0.062} & \hlc{-0.003}{0.062} & \hlc{-0.001}{0.062} & \hlc{0.000}{0.048} & \hlc{0.001}{0.048} & \hlc{-0.001}{0.048} & \hlc{-0.001}{0.048} & \hlc{-0.001}{0.048} & \hlc{0.005}{0.059} & \hlc{-0.001}{0.059} & \hlc{-0.002}{0.059} & \hlc{-0.001}{0.059} & \hlc{0.002}{0.059}\\
& Fair output & \hlc{0.002}{0.062} & \hlc{0.001}{0.062} & \hlc{0.000}{0.062} & \hlc{-0.002}{0.062} & \hlc{-0.000}{0.062} & \hlc{0.000}{0.048} & \hlc{0.000}{0.048} & \hlc{-0.001}{0.048} & \hlc{0.000}{0.048} & \hlc{0.001}{0.048} & \hlc{0.007}{0.059} & \hlc{0.000}{0.059} & \hlc{0.002}{0.059} & \hlc{-0.001}{0.059} & \hlc{0.001}{0.059}\\
& Mixed output & \hlc{0.001}{0.062} & \hlc{0.000}{0.062} & \hlc{0.000}{0.062} & \hlc{0.000}{0.062} & \hlc{0.004}{0.062} & \hlc{0.001}{0.048} & \hlc{0.000}{0.048} & \hlc{0.002}{0.048} & \hlc{0.000}{0.048} & \hlc{0.003}{0.048} & \hlc{0.007}{0.059} & \hlc{0.000}{0.059} & \hlc{-0.009}{0.059} & \hlc{-0.008}{0.059} & \hlc{-0.001}{0.059}\\
\hline
\multirow{7}{*}{30\%}
& Positive only & \hlc{0.000}{0.062} & \hlc{-0.004}{0.062} & \hlc{-0.002}{0.062} & \hlc{-0.005}{0.062} & \hlc{0.000}{0.062} & \hlc{0.000}{0.048} & \hlc{0.000}{0.048} & \hlc{0.000}{0.048} & \hlc{0.001}{0.048} & \hlc{0.005}{0.048} & \hlc{-0.001}{0.059} & \hlc{0.002}{0.059} & \hlc{0.003}{0.059} & \hlc{-0.001}{0.059} & \hlc{0.005}{0.059}\\
& Negative only & \hlc{0.000}{0.062} & \hlc{-0.004}{0.062} & \hlc{0.000}{0.062} & \hlc{-0.006}{0.062} & \hlc{0.001}{0.062} & \hlc{0.000}{0.048} & \hlc{-0.001}{0.048} & \hlc{0.001}{0.048} & \hlc{-0.002}{0.048} & \hlc{0.005}{0.048} & \hlc{0.000}{0.059} & \hlc{0.003}{0.059} & \hlc{0.000}{0.059} & \hlc{0.000}{0.059} & \hlc{0.007}{0.059}\\
& Fair input & \hlc{0.000}{0.062} & \hlc{-0.005}{0.062} & \hlc{-0.001}{0.062} & \hlc{-0.006}{0.062} & \hlc{0.001}{0.062} & \hlc{-0.001}{0.048} & \hlc{-0.001}{0.048} & \hlc{0.002}{0.048} & \hlc{-0.001}{0.048} & \hlc{0.005}{0.048} & \hlc{0.002}{0.059} & \hlc{0.003}{0.059} & \hlc{-0.001}{0.059} & \hlc{-0.004}{0.059} & \hlc{0.008}{0.059}\\
& Mixed input & \hlc{0.002}{0.062} & \hlc{-0.003}{0.062} & \hlc{0.001}{0.062} & \hlc{-0.006}{0.062} & \hlc{-0.001}{0.062} & \hlc{0.000}{0.048} & \hlc{-0.001}{0.048} & \hlc{0.005}{0.048} & \hlc{0.002}{0.048} & \hlc{0.007}{0.048} & \hlc{0.006}{0.059} & \hlc{0.006}{0.059} & \hlc{0.000}{0.059} & \hlc{-0.009}{0.059} & \hlc{0.004}{0.059}\\
& Biased output & \hlc{0.002}{0.062} & \hlc{-0.006}{0.062} & \hlc{0.001}{0.062} & \hlc{-0.003}{0.062} & \hlc{-0.000}{0.062} & \hlc{-0.002}{0.048} & \hlc{-0.001}{0.048} & \hlc{0.004}{0.048} & \hlc{0.000}{0.048} & \hlc{0.005}{0.048} & \hlc{0.005}{0.059} & \hlc{0.001}{0.059} & \hlc{-0.001}{0.059} & \hlc{0.001}{0.059} & \hlc{0.004}{0.059}\\
& Fair output & \hlc{0.004}{0.062} & \hlc{-0.006}{0.062} & \hlc{-0.003}{0.062} & \hlc{-0.007}{0.062} & \hlc{-0.003}{0.062} & \hlc{0.002}{0.048} & \hlc{-0.002}{0.048} & \hlc{0.004}{0.048} & \hlc{0.000}{0.048} & \hlc{0.006}{0.048} & \hlc{0.008}{0.059} & \hlc{0.004}{0.059} & \hlc{0.003}{0.059} & \hlc{-0.002}{0.059} & \hlc{0.005}{0.059}\\
& Mixed output & \hlc{0.000}{0.062} & \hlc{-0.004}{0.062} & \hlc{-0.005}{0.062} & \hlc{-0.004}{0.062} & \hlc{-0.001}{0.062} & \hlc{0.000}{0.048} & \hlc{-0.001}{0.048} & \hlc{0.002}{0.048} & \hlc{-0.003}{0.048} & \hlc{0.003}{0.048} & \hlc{0.003}{0.059} & \hlc{0.004}{0.059} & \hlc{-0.009}{0.059} & \hlc{-0.011}{0.059} & \hlc{0.003}{0.059}\\
\hline
\multirow{7}{*}{40\%}
& Positive only & \hlc{0.005}{0.062} & \hlc{-0.004}{0.062} & \hlc{-0.006}{0.062} & \hlc{-0.005}{0.062} & \hlc{0.003}{0.062} & \hlc{0.000}{0.048} & \hlc{0.005}{0.048} & \hlc{0.006}{0.048} & \hlc{0.001}{0.048} & \hlc{0.012}{0.048} & \hlc{0.004}{0.059} & \hlc{0.000}{0.059} & \hlc{-0.004}{0.059} & \hlc{0.001}{0.059} & \hlc{-0.001}{0.059}\\
& Negative only & \hlc{0.001}{0.062} & \hlc{-0.002}{0.062} & \hlc{-0.005}{0.062} & \hlc{-0.010}{0.062} & \hlc{0.001}{0.062} & \hlc{0.003}{0.048} & \hlc{0.004}{0.048} & \hlc{0.005}{0.048} & \hlc{0.003}{0.048} & \hlc{0.007}{0.048} & \hlc{0.007}{0.059} & \hlc{-0.002}{0.059} & \hlc{-0.005}{0.059} & \hlc{-0.006}{0.059} & \hlc{0.000}{0.059}\\
& Fair input & \hlc{-0.003}{0.062} & \hlc{-0.005}{0.062} & \hlc{-0.002}{0.062} & \hlc{-0.003}{0.062} & \hlc{0.004}{0.062} & \hlc{-0.001}{0.048} & \hlc{0.001}{0.048} & \hlc{0.004}{0.048} & \hlc{0.000}{0.048} & \hlc{0.010}{0.048} & \hlc{0.006}{0.059} & \hlc{0.001}{0.059} & \hlc{0.000}{0.059} & \hlc{-0.005}{0.059} & \hlc{0.000}{0.059}\\
& Mixed input & \hlc{-0.001}{0.062} & \hlc{-0.005}{0.062} & \hlc{0.000}{0.062} & \hlc{-0.004}{0.062} & \hlc{-0.002}{0.062} & \hlc{0.000}{0.048} & \hlc{0.005}{0.048} & \hlc{0.006}{0.048} & \hlc{0.002}{0.048} & \hlc{0.014}{0.048} & \hlc{0.008}{0.059} & \hlc{0.001}{0.059} & \hlc{-0.005}{0.059} & \hlc{-0.008}{0.059} & \hlc{-0.001}{0.059}\\
& Biased output & \hlc{0.003}{0.062} & \hlc{-0.003}{0.062} & \hlc{-0.009}{0.062} & \hlc{-0.005}{0.062} & \hlc{0.003}{0.062} & \hlc{0.001}{0.048} & \hlc{0.002}{0.048} & \hlc{0.001}{0.048} & \hlc{0.003}{0.048} & \hlc{0.009}{0.048} & \hlc{0.005}{0.059} & \hlc{-0.002}{0.059} & \hlc{-0.005}{0.059} & \hlc{-0.008}{0.059} & \hlc{-0.003}{0.059}\\
& Fair output & \hlc{0.004}{0.062} & \hlc{-0.002}{0.062} & \hlc{-0.004}{0.062} & \hlc{-0.006}{0.062} & \hlc{0.002}{0.062} & \hlc{-0.001}{0.048} & \hlc{0.000}{0.048} & \hlc{0.005}{0.048} & \hlc{0.002}{0.048} & \hlc{0.011}{0.048} & \hlc{0.009}{0.059} & \hlc{0.000}{0.059} & \hlc{-0.001}{0.059} & \hlc{0.001}{0.059} & \hlc{-0.004}{0.059}\\
& Mixed output & \hlc{0.004}{0.062} & \hlc{-0.004}{0.062} & \hlc{-0.006}{0.062} & \hlc{-0.006}{0.062} & \hlc{-0.000}{0.062} & \hlc{0.001}{0.048} & \hlc{0.001}{0.048} & \hlc{0.001}{0.048} & \hlc{0.002}{0.048} & \hlc{0.014}{0.048} & \hlc{0.007}{0.059} & \hlc{0.000}{0.059} & \hlc{-0.010}{0.059} & \hlc{-0.011}{0.059} & \hlc{-0.000}{0.059}\\
\bottomrule
\end{tabular}
\caption{Review summarisation}
\end{subtable}
\caption{Comparison of sparsity ratio, calibration sets, and model fairness using different datasets. The UER value of the vanilla model is reported in brackets next to the model name.  We report fairness improvement by calculating the absolute difference between the UER of the vanilla model and that of the pruned model. A model demonstrating a positive impact on fairness should have an absolute difference ranging from 0 to its vanilla UER, with values closer to the vanilla UER indicating better improvement (values between 0 and vanilla UER are highlighted, indicating that the pruned model is less biased than the original model). Darker colours indicate greater improvement in fairness.}

\label{tab:various_input_fairness_uer}
\end{table*}

\subsection{Model Performance Pruning by High Gradient and Low Activation}
\label{sec:model_performance_hgla}
Model performance using single-sided input and high gradient and low activation pruning are reported in Table~\ref{tab:hgla_political} and Table~\ref{tab:hgla_review} for summarising political tweets and reviews respectively.

\begin{table*}[htbp]
\centering
\small
\setlength{\tabcolsep}{4pt}
\begin{tabular}{@{}c l *{6}{c}@{}}
\toprule
\multirow{2}{*}{\begin{tabular}[c]{@{}c@{}}Sparsity\\Ratio\end{tabular}} & \multirow{2}{*}{\begin{tabular}[c]{@{}l@{}}Calibration\\Sets\end{tabular}} & \multicolumn{2}{c}{Llama3-8B} & \multicolumn{2}{c}{Gemma-2B} & \multicolumn{2}{c}{TinyLlama} \\
\cmidrule(lr){3-4} \cmidrule(lr){5-6} \cmidrule(l){7-8}
 &  & ROUGE1/2/L & BERTScore & ROUGE1/2/L & BERTScore & ROUGE1/2/L & BERTScore \\
\midrule
\multirow{2}{*}{10\%} & Left only & 0.216/0.063/0.160 & 0.832 & 0.303/0.106/0.218 & 0.866 & 0.265/0.094/0.189 & 0.850 \\
& Right only & 0.239/0.076/0.168 & 0.836 & 0.298/0.096/0.219 & 0.865 & 0.274/0.097/0.191 & 0.851 \\
\midrule
\multirow{2}{*}{20\%} & Left only & 0.229/0.066/0.161 & 0.837 & 0.310/0.099/0.225 & 0.864 & 0.258/0.078/0.180 & 0.844 \\
& Right only & 0.208/0.064/0.164 & 0.835 & 0.292/0.095/0.213 & 0.862 & 0.251/0.079/0.178 & 0.846 \\
\midrule
\multirow{2}{*}{30\%} & Left only & 0.230/0.068/0.161 & 0.832 & 0.257/0.076/0.185 & 0.854 & 0.259/0.086/0.185 & 0.849 \\
& Right only & 0.242/0.070/0.175 & 0.837 & 0.269/0.085/0.197 & 0.852 & 0.235/0.075/0.172 & 0.846 \\
\midrule
\multirow{2}{*}{40\%} & Left only & 0.181/0.059/0.145 & 0.819 & 0.207/0.067/0.162 & 0.834 & 0.250/0.079/0.181 & 0.844 \\
& Right only & 0.212/0.060/0.163 & 0.825 & 0.208/0.059/0.164 & 0.836 & 0.268/0.093/0.199 & 0.848 \\
\bottomrule
\end{tabular}
\caption{ROUGE and BERTScore metrics for different models and calibration sets across various sparsity ratios for summarising political tweets using high gradient low activation pruning.}
\label{tab:hgla_political}
\end{table*}

\begin{table*}[htbp]
\centering
\small
\setlength{\tabcolsep}{4pt}
\begin{tabular}{@{}c l *{6}{c}@{}}
\toprule
\multirow{2}{*}{\begin{tabular}[c]{@{}c@{}}Sparsity\\Ratio\end{tabular}} & \multirow{2}{*}{\begin{tabular}[c]{@{}l@{}}Calibration\\Sets\end{tabular}} & \multicolumn{2}{c}{Llama3-8B} & \multicolumn{2}{c}{Gemma-2B} & \multicolumn{2}{c}{TinyLlama} \\
\cmidrule(lr){3-4} \cmidrule(lr){5-6} \cmidrule(l){7-8}
 &  & ROUGE1/2/L & BERTScore & ROUGE1/2/L & BERTScore & ROUGE1/2/L & BERTScore \\
\midrule
\multirow{2}{*}{10\%} & Negative only & 0.195/0.029/0.119 & 0.826 & 0.271/0.043/0.180 & 0.863 & 0.246/0.043/0.153 & 0.850 \\
& Positive only & 0.199/0.028/0.122 & 0.831 & 0.273/0.044/0.182 & 0.862 & 0.247/0.045/0.153 & 0.850 \\
\midrule
\multirow{2}{*}{20\%} & Negative only & 0.189/0.027/0.116 & 0.828 & 0.264/0.044/0.177 & 0.861 & 0.242/0.043/0.150 & 0.853 \\
& Positive only & 0.185/0.026/0.115 & 0.829 & 0.261/0.043/0.178 & 0.859 & 0.249/0.046/0.157 & 0.856 \\
\midrule
\multirow{2}{*}{30\%} & Negative only & 0.175/0.026/0.109 & 0.820 & 0.201/0.027/0.147 & 0.838 & 0.248/0.041/0.150 & 0.851 \\
& Positive only & 0.186/0.026/0.113 & 0.825 & 0.214/0.029/0.157 & 0.842 & 0.245/0.040/0.150 & 0.851 \\
\midrule
\multirow{2}{*}{40\%} & Negative only & 0.178/0.023/0.118 & 0.827 & 0.117/0.012/0.093 & 0.810 & 0.236/0.040/0.154 & 0.849 \\
& Positive only & 0.156/0.020/0.114 & 0.816 & 0.155/0.016/0.126 & 0.820 & 0.229/0.038/0.152 & 0.847 \\
\bottomrule
\end{tabular}
\caption{ROUGE and BERTScore metrics for different models and calibration sets across various sparsity ratios for summarising reviews using high gradient low activation pruning.}
\label{tab:hgla_review}
\end{table*}

\subsection{Comparative Impact Analysis: Pruning Methods vs. Calibration Sets}
\label{sec:var_pruning_calibration}
Standard deviations across methods for each evaluation metric are presented in Tables~\ref{tab:pruning_methods} and~\ref{tab:calibration_sets}. The analysis reveals a consistent pattern across all four fairness metrics: pruning methods exhibit a greater impact on fairness outcomes than calibration sets. For each pruning method (rows in Table~\ref{tab:pruning_methods}), the variation across different calibration sets remains relatively small. Conversely, for each calibration set (rows in Table~\ref{tab:calibration_sets}), the variation across different pruning methods is substantially larger. This pattern holds consistently across both datasets and all fairness metrics. The systematic nature of these differences confirms that when pruning language models, pruning method has a greater impact on fairness compared to calibration set, while calibration set selection, though still relevant, has a more modest and secondary impact. These quantitative findings provide statistical support for our visual observations in Figure~\ref{fig:combined_by_method_calibration}, where calibration set trend lines show close proximity and frequent intersections, while pruning method differences remain pronounced and consistent across experimental conditions.

\begin{table*}[h]
\centering
\small
\begin{tabular}{l|cc|cc|cc|cc}
\hline
\multirow{2}{*}{\textbf{Pruning Method}} & \multicolumn{2}{c|}{\textbf{SPD}} & \multicolumn{2}{c|}{\textbf{SOF}} & \multicolumn{2}{c|}{\textbf{BUR}} & \multicolumn{2}{c}{\textbf{UER}} \\
& Political & Reviews & Political & Reviews & Political & Reviews & Political & Reviews \\
\hline
Sparse GPT & 0.1065 & 0.0476 & 0.0029 & 0.0016 & 0.0273 & 0.0185 & 0.0035 & 0.0017 \\
Wanda & 0.1475 & 0.0601 & 0.0035 & 0.0019 & 0.0411 & 0.0279 & 0.0048 & 0.0025 \\
GBLM Pruner & 0.1543 & 0.0952 & 0.0030 & 0.0014 & 0.0657 & 0.0635 & 0.0070 & 0.0064 \\
GBLM Gradient & 0.1472 & 0.0845 & 0.0033 & 0.0013 & 0.0398 & 0.0389 & 0.0053 & 0.0027 \\
HGLA & 0.0934 & 0.1158 & 0.0049 & 0.0015 & 0.0634 & 0.0217 & 0.0079 & 0.0021 \\
\hline
\textbf{Average} & 0.1298 & 0.0806 & 0.0035 & 0.0015 & 0.0475 & 0.0341 & 0.0057 & 0.0031 \\
\hline
\end{tabular}
\caption{Fairness Standard Deviation Across Pruning Methods: Each row shows how a specific pruning method performs with different \textbf{calibration sets}, with averages revealing the overall impact of pruning algorithm choice on fairness metrics.}

\label{tab:pruning_methods}
\end{table*}

\begin{table*}[h]
\centering
\small
\begin{tabular}{l|cc|cc|cc|cc}
\hline
\multirow{2}{*}{\textbf{Calibration Set}} & \multicolumn{2}{c|}{\textbf{SPD}} & \multicolumn{2}{c|}{\textbf{SOF}} & \multicolumn{2}{c|}{\textbf{BUR}} & \multicolumn{2}{c}{\textbf{UER}} \\
& Political & Reviews & Political & Reviews & Political & Reviews & Political & Reviews \\
\hline
Left only / Negative only & 0.1161 & 0.0845 & 0.0037 & 0.0016 & 0.0643 & 0.0574 & 0.0071 & 0.0056 \\
Right only / Positive only & 0.1339 & 0.0813 & 0.0036 & 0.0016 & 0.0760 & 0.0640 & 0.0080 & 0.0062 \\
Fair input & 0.1585 & 0.1239 & 0.0033 & 0.0019 & 0.0446 & 0.0309 & 0.0059 & 0.0026 \\
Mixed input & 0.1898 & 0.1393 & 0.0045 & 0.0015 & 0.0426 & 0.0366 & 0.0060 & 0.0026 \\
Biased output & 0.1030 & 0.1177 & 0.0044 & 0.0015 & 0.0395 & 0.0322 & 0.0053 & 0.0032 \\
Fair output & 0.1681 & 0.1119 & 0.0043 & 0.0010 & 0.0526 & 0.0333 & 0.0071 & 0.0032 \\
Mixed output & 0.1693 & 0.1292 & 0.0029 & 0.0016 & 0.0472 & 0.0386 & 0.0054 & 0.0030 \\
\hline
\textbf{Average} & 0.1484 & 0.1125 & 0.0038 & 0.0015 & 0.0524 & 0.0419 & 0.0064 & 0.0038 \\
\hline
\end{tabular}
\caption{Fairness Standard Deviation Across Calibration Sets: Each row shows how a specific calibration set performs with different \textbf{pruning methods}, with averages revealing the overall impact of calibration data choice on fairness metrics.}
\label{tab:calibration_sets}
\end{table*}

\subsection{Human Evaluation Detail}
\label{sec:human_evaluation_detail}
\begin{figure*}[htbp]
    \centering
    \includegraphics[width=\textwidth, keepaspectratio]{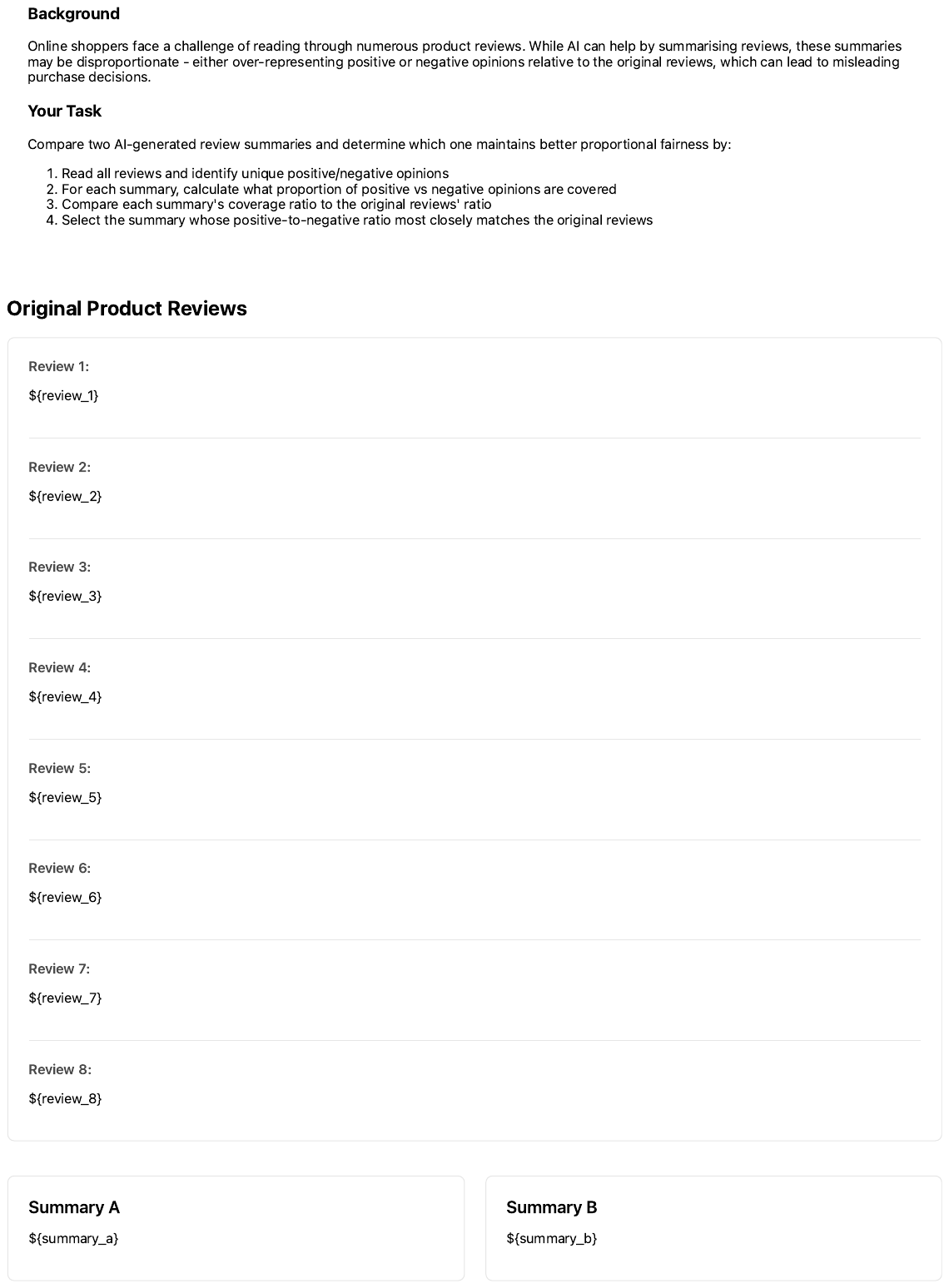}
    \caption{Annotation interface}
    \label{fig:annotation_interface}
\end{figure*}

% The human evaluation of fairness in summarisation presents significant challenges, primarily due to the extensive time required to read complete document sets. To address this limitation, we conducted experiments using a review dataset comprising eight concise reviews per document set. This dataset selection was advantageous as it enables untrained evaluators to more readily distinguish between positive and negative sentiments compared to identifying political leanings (e.g., left-leaning versus right-leaning orientations). 
% We implement the human evaluation protocol for summarisation fairness developed by~\citet{shandilya2020fairness}, which requires annotators to analyse all input documents and identify both positive and negative opinions.
% In our evaluation protocol, annotators first identified all distinct positive and negative opinions within each input document set. Subsequently, they assessed which of two generated summaries more accurately represented the original opinions' distribution. 
The evaluation interface design is illustrated in Figure~\ref{fig:annotation_interface}. The annotators need to meet the following criterias: HIT Approval Rate above 98\%, over 10,000 approved HITs, from English-speaking countries, and successful completion of quality check questions during annotation. We paid annotators \$10 USD per hour to meet ethical compensation standards and ensure quality participation. The final annotation achieved a Fleiss' Kappa of 0.555, indicating substantial inter-rater reliability.

\subsection{Biased and Fair Summaries}
\begin{figure*}[htbp]
    \centering
    \includegraphics[width=\textwidth, keepaspectratio]{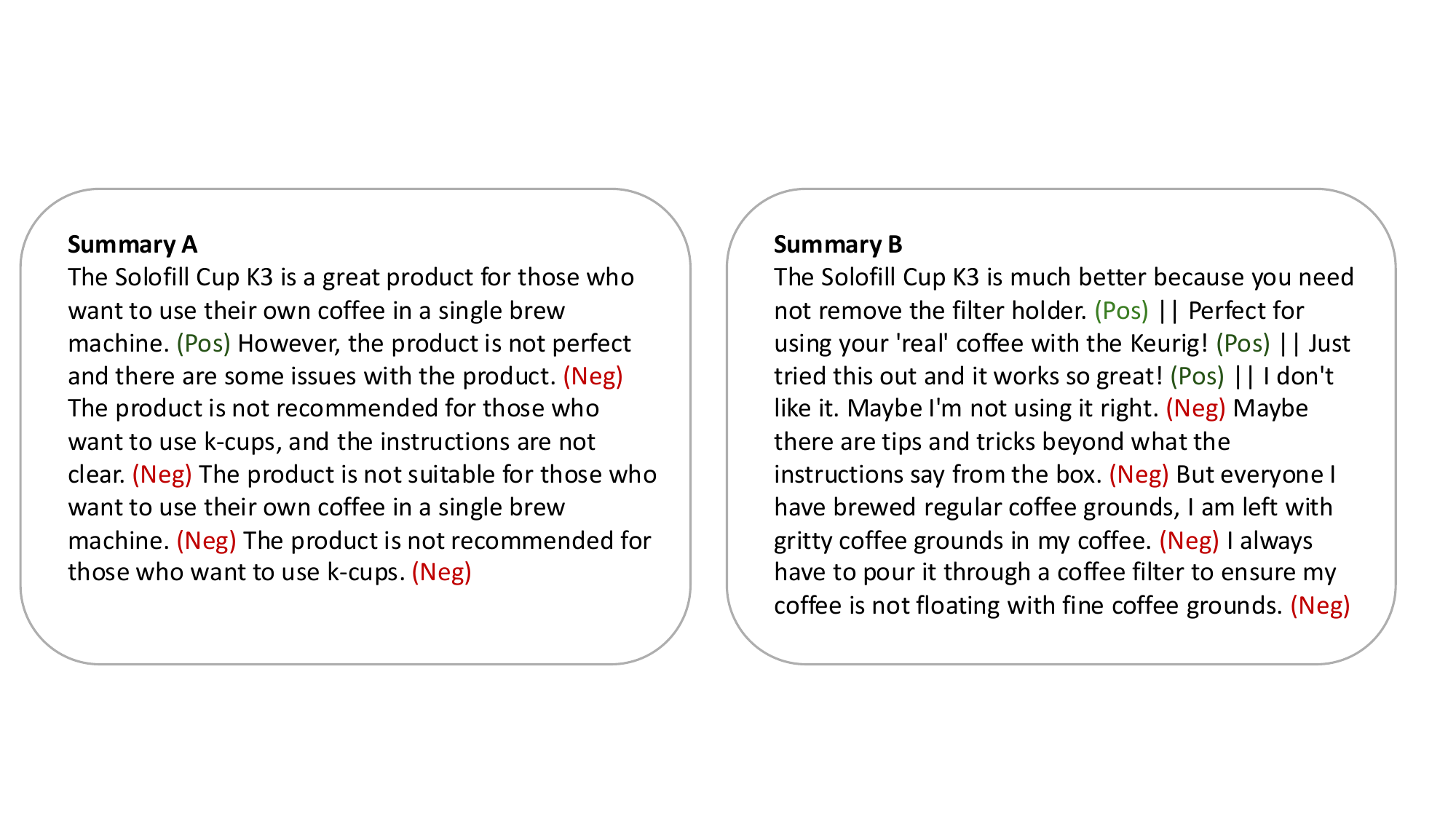}
    \caption{Example generated summaries}
    \label{fig:examples}
\end{figure*}

To illustrate the contrast between biased and more balanced output, we present examples of generated summaries in Figure~\ref{fig:examples}. In our analysis, sentences expressing positive sentiments are highlighted in green (Pos), while those expressing negative sentiments are highlighted in red (Neg). When provided with balanced input data, Summary A demonstrates bias by incorporating one positive sentiment and four negative sentiments. In contrast, Summary B achieves greater balance in sentiment distribution, containing three positive sentiments and four negative sentiments, thus providing a more balanced representation of the source documents.

\subsection{Raw Second-Order SPD}
The Second Ordered Statistical Parity Difference (SPD) metric we utilised monitors both the direction and magnitude of potential bias, enabling a comprehensive assessment of model fairness across pruning operations. This metric quantifies directional shifts in opinion fairness, with negative values indicating a left-leaning shift and positive values showing a right-leaning shift after pruning. The comprehensive analysis across various models, pruning techniques and calibration sets reveals distinct patterns in compression effects on opinion fairness. Detailed results are reported in Table~\ref{tab:raw_second_spd}.

\begin{table*}[htbp]

\centering
\tiny
\setlength{\tabcolsep}{4pt} % Reduce space between columns

\begin{subtable}{\textwidth}
\begin{tabular}{c c r r r r r r r r r r r r r r r}
\toprule
\multicolumn{2}{c}{} & \multicolumn{5}{c}{Llama3-8B} & \multicolumn{5}{c}{Gemma-2B} & \multicolumn{5}{c}{TinyLlama} \\
\cmidrule(lr){3-7} \cmidrule(lr){8-12} \cmidrule(lr){13-17}
Sparsity & Calibration & \multicolumn{1}{c}{Sparse} & \multicolumn{1}{c}{} & \multicolumn{1}{c}{GBLM} & \multicolumn{1}{c}{GBLM} & \multicolumn{1}{c}{} & \multicolumn{1}{c}{Sparse} & \multicolumn{1}{c}{} & \multicolumn{1}{c}{GBLM} & \multicolumn{1}{c}{GBLM} & \multicolumn{1}{c}{} & \multicolumn{1}{c}{Sparse} & \multicolumn{1}{c}{} & \multicolumn{1}{c}{GBLM} & \multicolumn{1}{c}{GBLM} & \multicolumn{1}{c}{} \\
Ratio & Sets & \multicolumn{1}{c}{GPT} & \multicolumn{1}{c}{Wanda} & \multicolumn{1}{c}{Pruner} & \multicolumn{1}{c}{Gradient} & \multicolumn{1}{c}{HGLA} & \multicolumn{1}{c}{GPT} & \multicolumn{1}{c}{Wanda} & \multicolumn{1}{c}{Pruner} & \multicolumn{1}{c}{Gradient} & \multicolumn{1}{c}{HGLA} & \multicolumn{1}{c}{GPT} & \multicolumn{1}{c}{Wanda} & \multicolumn{1}{c}{Pruner} & \multicolumn{1}{c}{Gradient} & \multicolumn{1}{c}{HGLA} \\
\hline
\multirow{7}{*}{10\%} & Left only & -0.134 & -0.160 & -0.183 & -0.183 & -0.209 & -0.165 & -0.142 & -0.189 & -0.157 & -0.152 & -0.008 & -0.041 & 0.002 & -0.062 & -0.011 \\
 & Right only & -0.152 & -0.146 & -0.123 & -0.199 & -0.124 & -0.176 & -0.165 & -0.137 & -0.183 & -0.158 & -0.035 & -0.050 & 0.009 & -0.070 & -0.025 \\
 & Fair input & -0.134 & -0.159 & -0.227 & -0.188 & -0.136 & -0.142 & -0.175 & -0.212 & -0.114 & -0.124 & -0.076 & -0.021 & -0.033 & -0.024 & -0.027 \\
 & Mixed input & -0.095 & -0.164 & -0.112 & -0.165 & -0.131 & -0.188 & -0.153 & -0.109 & -0.179 & -0.140 & -0.072 & -0.043 & 0.013 & -0.059 & -0.050 \\
 & Biased output & -0.132 & -0.130 & -0.157 & -0.078 & -0.206 & -0.122 & -0.175 & -0.128 & -0.174 & -0.125 & -0.071 & -0.042 & -0.052 & -0.082 & -0.025 \\
 & Fair output & -0.108 & -0.154 & -0.208 & -0.166 & -0.237 & -0.162 & -0.173 & -0.138 & -0.127 & -0.118 & -0.027 & -0.028 & -0.027 & -0.021 & -0.056 \\
 & Mixed output & -0.131 & -0.142 & -0.149 & -0.133 & -0.217 & -0.173 & -0.152 & -0.140 & -0.179 & -0.166 & -0.054 & -0.047 & -0.055 & -0.061 & -0.065 \\
\hline
\multirow{7}{*}{20\%} & Left only & -0.149 & -0.155 & -0.168 & -0.162 & -0.161 & -0.153 & -0.132 & -0.138 & -0.167 & -0.164 & -0.016 & -0.015 & -0.005 & -0.057 & 0.063 \\
 & Right only & -0.136 & -0.168 & -0.132 & -0.171 & -0.095 & -0.113 & -0.114 & -0.128 & -0.130 & -0.112 & -0.040 & -0.055 & -0.045 & -0.064 & 0.005 \\
 & Fair input & -0.142 & -0.148 & -0.171 & -0.159 & -0.137 & -0.156 & -0.160 & -0.147 & -0.137 & -0.119 & -0.064 & -0.039 & -0.057 & -0.050 & 0.082 \\
 & Mixed input & -0.139 & -0.149 & -0.181 & -0.152 & -0.187 & -0.144 & -0.171 & -0.150 & -0.153 & -0.156 & -0.055 & -0.033 & -0.059 & -0.059 & -0.008 \\
 & Biased output & -0.151 & -0.129 & -0.176 & -0.156 & -0.145 & -0.154 & -0.159 & -0.153 & -0.134 & -0.122 & -0.059 & -0.046 & -0.045 & -0.042 & 0.017 \\
 & Fair output & -0.142 & -0.154 & -0.182 & -0.170 & -0.150 & -0.148 & -0.132 & -0.131 & -0.144 & -0.159 & -0.016 & -0.028 & -0.039 & -0.055 & 0.072 \\
 & Mixed output & -0.124 & -0.170 & -0.127 & -0.160 & -0.127 & -0.141 & -0.162 & -0.127 & -0.160 & -0.152 & -0.054 & -0.055 & -0.034 & -0.060 & 0.017 \\
\hline
\multirow{7}{*}{30\%} & Left only & -0.151 & -0.154 & -0.180 & -0.157 & -0.072 & -0.155 & -0.152 & -0.173 & -0.167 & -0.238 & -0.021 & -0.032 & -0.031 & -0.062 & -0.026 \\
 & Right only & -0.165 & -0.158 & -0.148 & -0.152 & -0.034 & -0.143 & -0.127 & -0.131 & -0.143 & -0.161 & -0.061 & -0.053 & -0.021 & -0.050 & 0.100 \\
 & Fair input & -0.143 & -0.164 & -0.133 & -0.127 & -0.020 & -0.142 & -0.158 & -0.120 & -0.154 & -0.101 & -0.066 & -0.051 & -0.066 & -0.050 & 0.164 \\
 & Mixed input & -0.150 & -0.162 & -0.146 & -0.128 & -0.087 & -0.143 & -0.165 & -0.162 & -0.144 & -0.213 & -0.070 & -0.045 & -0.015 & -0.057 & -0.017 \\
 & Biased output & -0.142 & -0.156 & -0.161 & -0.157 & -0.052 & -0.143 & -0.170 & -0.143 & -0.161 & -0.141 & -0.079 & -0.032 & -0.014 & -0.063 & 0.112 \\
 & Fair output & -0.151 & -0.161 & -0.150 & -0.162 & -0.077 & -0.146 & -0.158 & -0.159 & -0.130 & -0.195 & -0.053 & -0.025 & -0.035 & -0.047 & 0.079 \\
 & Mixed output & -0.149 & -0.166 & -0.138 & -0.178 & -0.096 & -0.143 & -0.164 & -0.136 & -0.145 & -0.194 & -0.064 & -0.038 & -0.059 & -0.071 & 0.064 \\
\hline
\multirow{7}{*}{40\%} & Left only & -0.154 & -0.154 & -0.167 & -0.170 & -0.047 & -0.166 & -0.151 & -0.130 & -0.157 & -0.212 & -0.035 & -0.040 & -0.049 & -0.058 & -0.024 \\
 & Right only & -0.146 & -0.132 & -0.132 & -0.135 & -0.008 & -0.154 & -0.176 & -0.143 & -0.156 & -0.079 & -0.015 & -0.024 & -0.033 & -0.042 & 0.173 \\
 & Fair input & -0.134 & -0.152 & -0.129 & -0.134 & -0.178 & -0.158 & -0.159 & -0.137 & -0.139 & -0.097 & -0.066 & -0.038 & -0.056 & -0.061 & 0.046 \\
 & Mixed input & -0.149 & -0.137 & -0.152 & -0.140 & 0.006 & -0.161 & -0.146 & -0.160 & -0.158 & -0.169 & -0.050 & -0.033 & -0.011 & -0.050 & 0.164 \\
 & Biased output & -0.155 & -0.137 & -0.164 & -0.161 & -0.086 & -0.167 & -0.161 & -0.145 & -0.165 & -0.164 & -0.057 & -0.051 & -0.038 & -0.074 & 0.224 \\
 & Fair output & -0.147 & -0.130 & -0.137 & -0.138 & -0.086 & -0.155 & -0.182 & -0.157 & -0.170 & -0.091 & -0.035 & -0.023 & -0.055 & -0.044 & 0.103 \\
 & Mixed output & -0.149 & -0.147 & -0.131 & -0.140 & -0.041 & -0.159 & -0.151 & -0.128 & -0.158 & -0.110 & -0.053 & -0.046 & -0.057 & -0.065 & 0.124 \\
\bottomrule
\end{tabular}
\caption{Political tweet summarisation}
\end{subtable}

\bigskip

\begin{subtable}{\textwidth}
\begin{tabular}{c c r r r r r r r r r r r r r r r}
\toprule
\multicolumn{2}{c}{} & \multicolumn{5}{c}{Llama3-8B} & \multicolumn{5}{c}{Gemma-2B} & \multicolumn{5}{c}{TinyLlama} \\
\cmidrule(lr){3-7} \cmidrule(lr){8-12} \cmidrule(lr){13-17}
Sparsity & Calibration & \multicolumn{1}{c}{Sparse} & \multicolumn{1}{c}{} & \multicolumn{1}{c}{GBLM} & \multicolumn{1}{c}{GBLM} & \multicolumn{1}{c}{} & \multicolumn{1}{c}{Sparse} & \multicolumn{1}{c}{} & \multicolumn{1}{c}{GBLM} & \multicolumn{1}{c}{GBLM} & \multicolumn{1}{c}{} & \multicolumn{1}{c}{Sparse} & \multicolumn{1}{c}{} & \multicolumn{1}{c}{GBLM} & \multicolumn{1}{c}{GBLM} & \multicolumn{1}{c}{} \\
Ratio & Sets & \multicolumn{1}{c}{GPT} & \multicolumn{1}{c}{Wanda} & \multicolumn{1}{c}{Pruner} & \multicolumn{1}{c}{Gradient} & \multicolumn{1}{c}{HGLA} & \multicolumn{1}{c}{GPT} & \multicolumn{1}{c}{Wanda} & \multicolumn{1}{c}{Pruner} & \multicolumn{1}{c}{Gradient} & \multicolumn{1}{c}{HGLA} & \multicolumn{1}{c}{GPT} & \multicolumn{1}{c}{Wanda} & \multicolumn{1}{c}{Pruner} & \multicolumn{1}{c}{Gradient} & \multicolumn{1}{c}{HGLA} \\
\hline
\multirow{7}{*}{10\%} & Positive only & -0.236 & -0.190 & -0.208 & -0.220 & -0.238 & -0.397 & -0.393 & -0.401 & -0.393 & -0.399 & -0.187 & -0.145 & -0.179 & -0.177 & -0.176\\
 & Negative only & -0.211 & -0.249 & -0.266 & -0.271 & -0.206 & -0.398 & -0.416 & -0.397 & -0.395 & -0.399 & -0.172 & -0.184 & -0.179 & -0.182 & -0.177\\
 & Fair input & -0.220 & -0.257 & -0.296 & -0.282 & -0.198 & -0.404 & -0.421 & -0.400 & -0.426 & -0.406 & -0.200 & -0.162 & -0.187 & -0.182 & -0.177 \\
 & Mixed input & -0.195 & -0.243 & -0.261 & -0.261 & -0.213 & -0.403 & -0.417 & -0.411 & -0.412 & -0.391 & -0.192 & -0.171 & -0.175 & -0.182 & -0.178 \\
 & Biased output & -0.209 & -0.248 & -0.253 & -0.246 & -0.212 & -0.399 & -0.391 & -0.399 & -0.391 & -0.406 & -0.201 & -0.159 & -0.191 & -0.187 & -0.171 \\
 & Fair output & -0.192 & -0.253 & -0.272 & -0.278 & -0.207 & -0.402 & -0.416 & -0.401 & -0.411 & -0.392 & -0.171 & -0.156 & -0.178 & -0.173 & -0.179 \\
 & Mixed output & -0.183 & -0.238 & -0.240 & -0.222 & -0.205 & -0.399 & -0.402 & -0.385 & -0.392 & -0.395 & -0.190 & -0.164 & -0.175 & -0.191 & -0.180 \\
\hline
\multirow{7}{*}{20\%} & Positive only & -0.196 & -0.239 & -0.231 & -0.237 & -0.200 & -0.408 & -0.392 & -0.414 & -0.415 & -0.406 & -0.165 & -0.150 & -0.190 & -0.183 & -0.153\\
 & Negative only & -0.219 & -0.248 & -0.262 & -0.246 & -0.176 & -0.409 & -0.408 & -0.398 & -0.402 & -0.409 & -0.172 & -0.173 & -0.160 & -0.176 & -0.151\\
 & Fair input & -0.186 & -0.205 & -0.218 & -0.222 & -0.197 & -0.401 & -0.383 & -0.396 & -0.375 & -0.418 & -0.176 & -0.173 & -0.161 & -0.159 & -0.138 \\
 & Mixed input & -0.203 & -0.255 & -0.246 & -0.242 & -0.190 & -0.407 & -0.397 & -0.403 & -0.411 & -0.397 & -0.170 & -0.171 & -0.155 & -0.176 & -0.190 \\
 & Biased output & -0.202 & -0.225 & -0.231 & -0.228 & -0.212 & -0.395 & -0.387 & -0.392 & -0.406 & -0.398 & -0.186 & -0.162 & -0.160 & -0.175 & -0.163 \\
 & Fair output & -0.174 & -0.248 & -0.244 & -0.240 & -0.227 & -0.404 & -0.392 & -0.407 & -0.401 & -0.400 & -0.181 & -0.177 & -0.165 & -0.179 & -0.151 \\
 & Mixed output & -0.206 & -0.215 & -0.207 & -0.265 & -0.208 & -0.399 & -0.402 & -0.391 & -0.401 & -0.398 & -0.178 & -0.154 & -0.171 & -0.162 & -0.176 \\
\hline
\multirow{7}{*}{30\%} & Positive only & -0.193 & -0.224 & -0.214 & -0.206 & -0.210 & -0.404 & -0.393 & -0.402 & -0.393 & -0.408 & -0.182 & -0.158 & -0.166 & -0.170 & -0.159\\
 & Negative only & -0.183 & -0.210 & -0.187 & -0.213 & -0.191 & -0.410 & -0.405 & -0.418 & -0.400 & -0.394 & -0.162 & -0.164 & -0.154 & -0.158 & -0.141\\
 & Fair input & -0.210 & -0.252 & -0.253 & -0.268 & -0.190 & -0.416 & -0.404 & -0.410 & -0.407 & -0.411 & -0.182 & -0.151 & -0.171 & -0.163 & -0.136 \\
 & Mixed input & -0.199 & -0.229 & -0.231 & -0.228 & -0.218 & -0.410 & -0.392 & -0.405 & -0.415 & -0.427 & -0.196 & -0.175 & -0.180 & -0.191 & -0.151 \\
 & Biased output & -0.185 & -0.250 & -0.237 & -0.248 & -0.234 & -0.418 & -0.405 & -0.411 & -0.434 & -0.378 & -0.181 & -0.156 & -0.167 & -0.183 & -0.165 \\
 & Fair output & -0.178 & -0.231 & -0.230 & -0.252 & -0.241 & -0.412 & -0.408 & -0.417 & -0.424 & -0.402 & -0.171 & -0.159 & -0.180 & -0.176 & -0.137 \\
 & Mixed output & -0.182 & -0.227 & -0.236 & -0.251 & -0.224 & -0.408 & -0.401 & -0.404 & -0.402 & -0.431 & -0.196 & -0.153 & -0.165 & -0.162 & -0.173 \\
\hline
\multirow{7}{*}{40\%} & Positive only & -0.200 & -0.236 & -0.222 & -0.220 & -0.268 & -0.417 & -0.378 & -0.385 & -0.380 & -0.323 & -0.193 & -0.160 & -0.171 & -0.182 & -0.161\\
 & Negative only & -0.178 & -0.248 & -0.262 & -0.263 & -0.304 & -0.406 & -0.378 & -0.383 & -0.359 & -0.283 & -0.173 & -0.160 & -0.168 & -0.162 & -0.224\\
 & Fair input & -0.196 & -0.261 & -0.269 & -0.280 & -0.262 & -0.419 & -0.392 & -0.391 & -0.385 & -0.356 & -0.176 & -0.165 & -0.159 & -0.166 & -0.192 \\
 & Mixed input & -0.199 & -0.289 & -0.289 & -0.304 & -0.385 & -0.413 & -0.379 & -0.371 & -0.382 & -0.372 & -0.185 & -0.155 & -0.160 & -0.143 & -0.171 \\
 & Biased output & -0.182 & -0.262 & -0.282 & -0.284 & -0.307 & -0.421 & -0.378 & -0.361 & -0.358 & -0.348 & -0.178 & -0.144 & -0.169 & -0.136 & -0.196 \\
 & Fair output & -0.206 & -0.270 & -0.322 & -0.302 & -0.290 & -0.424 & -0.396 & -0.371 & -0.379 & -0.310 & -0.186 & -0.165 & -0.207 & -0.120 & -0.196 \\
 & Mixed output & -0.218 & -0.272 & -0.344 & -0.315 & -0.377 & -0.436 & -0.381 & -0.313 & -0.380 & -0.347 & -0.174 & -0.140 & -0.173 & -0.082 & -0.183 \\
\bottomrule
\end{tabular}
\caption{Review summarisation}
\end{subtable}
\caption{
Raw Second-Order SPD
}
\label{tab:raw_second_spd}
\end{table*}

\subsection{Comparative Analysis of HGLA and Wanda}
We evaluated the statistical significance of differences between HGLA and Wanda by conducting paired Wilcoxon signed-rank tests across all experimental conditions. The values in Tables~\ref{tab:political_tweet}, \ref{tab:review_summarisation}, \ref{tab:dataset3}, and~\ref{tab:dataset4}
 represent the absolute differences using HGLA. Results marked with an asterisk (*) denote statistically significant different from Wanda (p < 0.05). Our analysis demonstrates that HGLA produces significant modifications across various models, pruning scenarios, and metrics, with effectiveness generally increasing at higher pruning ratios (0.3-0.4). These findings provide statistical evidence that HGLA's selective information pruning approach can reliably alter model behavior, with implications for mitigating bias and enhancing fairness in language models.

\begin{table*}[htbp]
\centering
\small
\setlength{\tabcolsep}{4pt} % Reduce space between columns

\begin{tabular}{|c|ccc|ccc|ccc|}
\hline
\multirow{2}{*}{Ratio} & \multicolumn{3}{c|}{Llama3-8B} & \multicolumn{3}{c|}{Gemma-2B} & \multicolumn{3}{c|}{TinyLlama} \\
\cline{2-10}
 & SPD & SOF & UER & SPD & SOF & UER & SPD & SOF & UER \\
\hline
0.1 & -0.060 & -0.006 & 0.006 & -0.046 & 0.005 & 0.002 & 0.038 & -0.007 & 0.011 \\
0.2 & -0.004 & -0.004 & 0.000 & 0.045 & 0.010 & -0.002 & 0.077 & 0.003 & 0.010 \\
0.3 & 0.119 & -0.005 & -0.001 & -0.052 & 0.004* & -0.013 & -0.111* & 0.004* & 0.024 \\
0.4 & 0.170* & 0.003 & 0.015 & 0.112 & 0.015 & 0.004 & -0.258* & 0.010 & 0.011 \\
\hline
\end{tabular}
\caption{Political tweet summarisation---Right Information Only Pruned (HGLA vs Wanda)}
\label{tab:political_tweet}

\bigskip

\begin{tabular}{|c|ccc|ccc|ccc|}
\hline
\multirow{2}{*}{Ratio} & \multicolumn{3}{c|}{Llama3-8B} & \multicolumn{3}{c|}{Gemma-2B} & \multicolumn{3}{c|}{TinyLlama} \\
\cline{2-10}
 & SPD & SOF & UER & SPD & SOF & UER & SPD & SOF & UER \\
\hline
0.1 & -0.232 & -0.003* & -0.001 & -0.035 & 0.005 & 0.001 & 0.065 & -0.012* & 0.015* \\
0.2 & -0.134 & 0.004 & 0.002 & -0.059* & 0.015 & 0.000 & -0.037* & -0.003* & 0.015 \\
0.3 & 0.043 & -0.002 & 0.002 & -0.205* & 0.009 & -0.008 & 0.037 & -0.007 & 0.010 \\
0.4 & 0.092* & 0.001* & 0.019 & -0.153 & 0.017 & 0.002 & 0.040 & -0.004 & 0.010 \\
\hline
\end{tabular}
\caption{Political tweet summarisation---Left Information Only Pruned (HGLA vs Wanda)}
\label{tab:review_summarisation}
\bigskip

\begin{tabular}{|c|ccc|ccc|ccc|}
\hline
\multirow{2}{*}{Ratio} & \multicolumn{3}{c|}{Llama3-8B} & \multicolumn{3}{c|}{Gemma-2B} & \multicolumn{3}{c|}{TinyLlama} \\
\cline{2-10}
 & SPD & SOF & UER & SPD & SOF & UER & SPD & SOF & UER \\
\hline
0.1 & 0.020* & 0.000 & -0.003 & -0.013 & -0.001 & 0.001 & 0.031 & -0.001* & 0.003* \\
0.2 & 0.097 & -0.003 & 0.004 & -0.028* & 0.002 & 0.000 & 0.077 & -0.001* & 0.002 \\
0.3 & 0.077* & -0.003 & 0.000 & -0.032 & -0.001* & 0.005 & 0.064 & -0.001 & 0.005 \\
0.4 & -0.040 & 0.001* & 0.003* & 0.140* & -0.003* & 0.012 & 0.061 & -0.001 & -0.001 \\
\hline
\end{tabular}
\caption{Review summarisation---Positive Information Only Pruned (HGLA vs Wanda)}
\label{tab:dataset3}

\bigskip

\begin{tabular}{|c|ccc|ccc|ccc|}
\hline
\multirow{2}{*}{Ratio} & \multicolumn{3}{c|}{Llama3-8B} & \multicolumn{3}{c|}{Gemma-2B} & \multicolumn{3}{c|}{TinyLlama} \\
\cline{2-10}
 & SPD & SOF & UER & SPD & SOF & UER & SPD & SOF & UER \\
\hline
0.1 & 0.085 & -0.003 & -0.001 & -0.013 & 0.001 & 0.000 & 0.029 & -0.002* & 0.003* \\
0.2 & 0.145 & -0.003 & 0.000 & -0.034* & 0.001 & 0.000 & 0.080 & -0.001 & 0.000 \\
0.3 & 0.114* & 0.001* & 0.001* & -0.004 & -0.003* & 0.005* & 0.101 & -0.001 & 0.007 \\
0.4 & -0.112 & -0.001 & 0.001 & 0.219* & -0.001* & 0.007* & -0.065* & 0.000 & 0.000 \\
\hline
\end{tabular}
\caption{Review summarisation---Negative Information Only Pruned (HGLA vs Wanda)}
\label{tab:dataset4}
\end{table*}

\end{document}